\documentclass[letterpaper]{article}

\usepackage{ActionForcing}

\usepackage[hyphens]{url}
\usepackage{graphicx}
\usepackage[numbers,sort&compress]{natbib}

\usepackage{caption}
\usepackage{amsmath}
\usepackage{amssymb}

\usepackage{algorithm}
\usepackage{algorithmic}

\usepackage{subcaption}

\usepackage{newfloat}
\usepackage{listings}
\DeclareCaptionStyle{ruled}{
    labelfont=normalfont,
    labelsep=colon,
    strut=off
}
\floatstyle{ruled}
\newfloat{listing}{tb}{lst}{}
\floatname{listing}{Listing}

\usepackage{booktabs}

\usepackage{xcolor}

\makeatletter
\providecommand{\reviewoldlabel}[2]{%
    \begingroup
    \def\@currentlabel{#2}%
    \label{#1}%
    \endgroup
}
\makeatother

\title{
Action Forcing: Training World Models on Unsupervised Video by Recovering
Underlying Egomotion Bases
}

\author{
\textbf{Ashish Sundar}\textsuperscript{1,2} \qquad
\textbf{Tiankuo Hou}\textsuperscript{1} \qquad
\textbf{Zhong Fan}\textsuperscript{1}\\[0.35em]
\textbf{Chunbo Luo}\textsuperscript{1} \qquad
\textbf{Xiaoyang Wang}\textsuperscript{1}
}

\affiliations{
\textsuperscript{1}University of Exeter, Exeter, United Kingdom\\
\textsuperscript{2}Environmental Intelligence CDT
}

\begin{document}

\maketitle

\begin{abstract}
We are living through a golden age of videos: cameras are ubiquitous,
storage is inexpensive, and vast amounts of footage are recorded every day.
Yet, the synchronised action annotations needed to train controllable world
models remain elusive. Existing approaches make use of instrumented
platforms with calibrated sensors, costly manual
annotation, or latent-action models which lack grounding. We instead turn ordinary unlabelled video into
action-supervised training data by recovering (without training) a data-derived egomotion basis. We track pixel displacements across frames and exploit the recurring coherent
structure induced by egomotion to obtain grounded control signals directly. Using a method as simple as principal components analysis perform this, we find that the leading components
provide signed, scalable, and composable throttle--yaw controls, although the
method can recover only motion axes represented in the data. To prevent a high-capacity video DiT from
exploiting pixel-level supervision, an online latent critic distils a frozen
decoder--tracker--PCA (Principal Components Analysis) teacher without backpropagating through the decoder or
tracker. Finally we critique the use of video generation metrics to evaluate WMs and introduce an example of an alternative, reference-free evaluation method. We measure \textit{controllability}, \textit{plausibility}, \textit{conjuring} (creating objects out of thin air) and \textit{geometric integrity}, revealing failures that conventional video metrics miss.
We show that most baselines follow familiar action directions but struggle to reverse or remain stationary. Our model handles both while retaining compositional control and generation quality. Despite backwards actions being less than $1\%$ of our training data, we find that the model learns to reverse, scale its response linearly, and compose throttle with steering, all simply by learning through a grounded action space. Cross model action forcing shows that most evaluated systems either fail to reverse and perhaps surprisingly, struggle to stay still. Our model does both consistently.
Ultimately, our results show that unlabelled real world video can be used to train WMs, removing the shackles of action labels from WM research and training.
\end{abstract}


\section{Introduction}

Large video diffusion transformers have recently found much success in learning powerful visual and temporal priors, but converting them into world models remains challenging without good action supervision. To get supervision, new data either needs to be painstakingly collected \cite{navwormod, cameracontrol2, cosmos2, uniwm, xworld, gaia2}, or training is relegated to the domain of videogames \cite{gamengen2024, diamond, olafworld, mineworld, gamefactory, oasis2}: synthesised worlds that can only try to mimic the complexity and stochasticity of the real world, but are not suitable for real world applications. While some \cite{genie, gamengen2024, olafworld, adaworld} use latent action models to bridge the gap, these often prove to lack any meaningful grounding (cross context action identifiability). Tellingly, even high performing systems retreat from messy data when they can: ~\citet{relic} abandons real world walking and drone data for unreal engine generated data, because real world data is dominated by forward motion and motion entangled across different dimensions.

We train directly on this difficult regime. FrodoBots
video~\cite{frodobots} contains low-speed rover journeys through real cities,
with pedestrians, uneven ground, occlusion, weather, lens contamination, and
unstructured interactions. Its recorded commands are significantly affected by network latency and do not reliably describe realized motion; we find that the labels actively destabilise training. We therefore treat the videos as unlabelled and recover control inputs from the motion rendered in the pixels.

Our approach rests on one primary idea - the pixel movements within frame follow the same broad stroke patterns for the same egomotion directions. When principal components of these pixel vectors are taken over a large corpus, larger egomotion patterns emerge clearly. We use this as an action space as it ends up being linearly scalable, invertible, and composable across multiple directions. We use the two highest variance component weights as the egomotion input and thus avoid conditioning on dense flow or raw tracks
directly~\cite{flovd,motionprompting,onlyflow}. This means we do not need to construct a full target motion
field at inference time.

We instantiate the method by adapting Wan2.1 on a coherent $88$ hour subset composed of 11 cities from the larger $2000$ hour FrodoBots video dataset. Despite finding only approximately $55$ minutes (1\% of the total footage) of sustained
reverse motion in the full corpus, the resulting model learns backward
control and combines it well with steering. Representative qualitative rollouts are provided in
Appendix~\ref{app:ex}.

Our contributions:
\begin{enumerate}

\item We recover an interpretable, signed, and composable egomotion action
    space directly from unlabelled real-world video using point tracking and
    principal components analysis (PCA), without recorded actions or a learned latent-action model.

\item We introduce an online latent egomotion critic that distils pixel-space motion into differentiable latent supervision, grounding action following without backpropagating through the video decoder or tracker.

\item We show that standard video evaluation metrics are inadequate for
world models, where action forced rollouts have no paired reference to judge against. We introduce an example of a reference-free action-forcing evaluation that measures controllability, geometric integrity, and plausibility.

\end{enumerate}

\section{Related Work}

\paragraph{World models from pretrained video generators.}
Recent work adapts pretrained video DiTs into action-conditioned rollout
models using camera poses, keyboard--mouse inputs, game controls, or curated
robot actions~\cite{minwm,worldcam,astra,matrixgame2,hyworldplay,yume,
cosmos,dws,vid2world}. We compare directly against minWM, WorldCam, Astra,
Matrix-Game 2.0, HY-WorldPlay, and Yume. These systems demonstrate strong
controllability when actions are available, but rely on labelled,
instrumented, or simulated control signals. We instead recover a grounded
egomotion interface directly from unlabelled real-world video.

\paragraph{Actions recovered from video.}
Latent-action models learn controls through inverse dynamics without recorded
labels~\cite{genie,lapo,lapa,adaworld}, but their codes can be opaque and
inconsistent across contexts~\cite{olafworld}. Motion-conditioned generators
instead recover optical flow or dense point tracks from video
\cite{motionprompting,flovd,onlyflow}, but these representations are
high-dimensional and require a full motion field at inference time. We
compress tracked motion with PCA into a signed, low-dimensional action space
that supports direct commands, reversal, and composition.

\paragraph{VLM-based evaluation of generated video.}
Recent metrics use multimodal models or learned evaluators to assess generated
content \cite{viescore,vqascore,videophy,videoscore,videobench}. We instead use
narrow reference-conditioned probes for specific world-model failures and
validate each instrument against human annotations.

\section{Method}
\subsection{Preliminaries}
We use conditional flow matching with the linear path
$x_t=(1-t)x_0+t x_1$ between noise $x_0\sim\mathcal{N}(0,I)$ and a clean latent
video chunk $x_1$. The model regresses the path velocity:
\begin{equation}
\mathcal{L}_{\mathrm{flow}}
=
\mathbb{E}\!\left[
\left\|v_\theta(x_t,t,c)-(x_1-x_0)\right\|_2^2
\right].
\end{equation}

\subsection{Action-Conditioned Flow}
\label{sec:planar}

A pretrained video DiT defines a context-conditioned velocity field
$v_\theta(x_t,t,c)$ whose integration produces a plausible continuation of
the preceding frames $c$. For brevity, let $z_t=(x_t,t,c)$ denote the current
latent state, flow time, and context. We expose a low-dimensional egomotion
interface to redirect the flow by augmenting this field with an action $a$:
\begin{equation}
    d x_t = v_\theta(z_t,a)\,dt.
    \label{eq:action_conditioned_ode}
\end{equation}
The context $z_t$ primarily determines the scene contents and plausible scene evolution while the action $a$ directs scene motion, affecting both generation and scene evolution. Because egomotion influences generation in this way, we condition the velocity network itself rather than applying a
fixed transformation to the completed sample.

At a common latent state, we isolate and ground the action-induced change
relative to a no-op command. Here and below, $\mathbf{0}$ denotes the semantic
no-egomotion command:
\begin{equation}
    \Delta v_\theta(z_t,a)
    =
    v_\theta(z_t,a)
    -
    v_\theta(z_t,\mathbf{0}).
    \label{eq:action_flow_difference}
\end{equation}
The action representation determines the geometry that the conditional model
is asked to learn. Rover egomotion is continuous and low dimensional: zero
denotes no commanded motion, scalar changes represent command strength,
different control axes may act simultaneously, and changing the sign of an
axis requests the opposite motion. We therefore represent these actions in a
centred affine coordinate system in which
\[
\{0,\qquad
\lambda a,\qquad
a_1+a_2,\qquad
-a\}
\]
denote respectively no egomotion, scaled motion, compound motion, and motion
in the opposite direction. We use this structure as an inductive bias. A generic
categorical or learned latent action representation can in principle learn
the same relationships, but the action space does not provide an intrinsic notion of neighbourhood, magnitude, addition, or inverse. Under limited data, those relationships must then be inferred separately.

The visual realization of a command nevertheless remains scene dependent.
Depth, perspective, pose, visibility, occlusion, and scene geometry determine
how the same egomotion changes the video. We therefore do not assume a fixed
action-to-warp mapping. Instead, we hypothesize that, at a fixed $z_t$, the
action-induced change in the velocity field is locally control-affine:
\begin{equation}
    \Delta v_\theta(z_t,a)
    =
    G_\theta(z_t)a
    +
    \epsilon_\theta(z_t,a),
    \label{eq:implicit_control_affine}
\end{equation}
where $G_\theta$ is an implicit state-, time-, and scene-dependent action
operator, and $\epsilon_\theta$ captures action-dependent departures from
this affine approximation. The nonlinear realization of motion is contained
in $G_\theta$; the approximation concerns only how the low-dimensional
command selects and combines these scene-dependent responses.

When $\epsilon_\theta$ is small over the operational action range, the model
approximately transfers the geometry of the command space to its velocity
field:
\begin{equation}
    \begin{aligned}
        \Delta v_\theta(z_t,\lambda a_1+\gamma a_2)
        &\approx
        \lambda\Delta v_\theta(z_t,a_1)
        +
        \gamma\Delta v_\theta(z_t,a_2),\\
        \Delta v_\theta(z_t,-a)
        &\approx
        -\Delta v_\theta(z_t,a),\\
        v_\theta(z_t,0)
        &\approx
        v_\theta(z_t).
    \end{aligned}
    \label{eq:control_affine_predictions}
\end{equation}
The first relation captures smooth scaling and composition: command
magnitudes and combinations reuse common scene-dependent action directions
rather than defining separate conditional behaviours. The second captures
sign reversal: opposite egomotions occupy opposite sides of a shared axis.
This in theory permits abundant motion on one side of an axis (forward motion, perhaps) to inform its shared
response direction, while the scarcer opposite examples (backwards motion here) calibrate and test the negative branch. The
zero command provides the common reference from which every action-induced
change is measured. Actions are therefore represented as redirections of one
context-conditioned flow rather than as independent flow maps. Section~\ref{sec:pca}
constructs the centred action coordinates, while our response, composition,
and counterfactual evaluations test whether the learned generator transfers
their geometry to the flow.

\subsection{Data Derived Action Space}
\label{sec:pca}

Section~\ref{sec:planar} motivates a centred command space with consistent
notions of scale, composition, and sign. We recover such coordinates directly
from unlabeled video motion, while leaving the scene-dependent visual
realization of each command to the model.

\begin{figure*}[t]
    \centering
    \includegraphics[width=\textwidth]
    {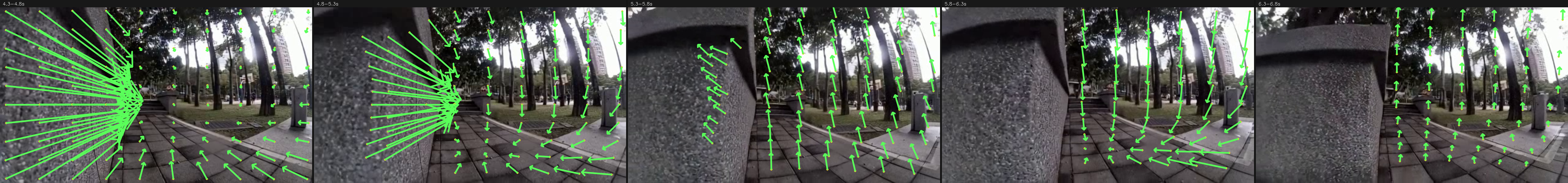}

    \includegraphics[width=\textwidth]
    {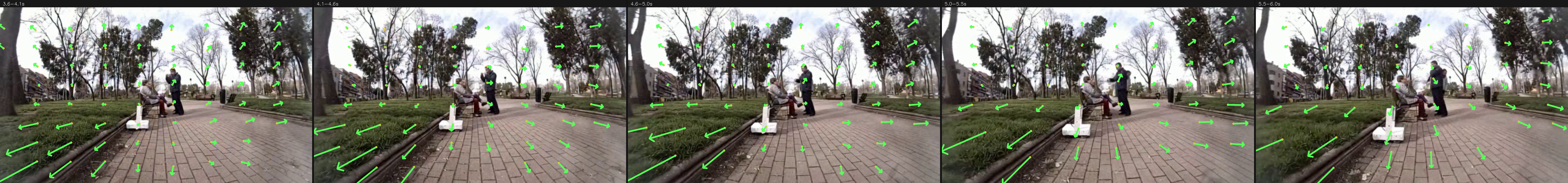}
    \caption{
        Tracked displacement fields from two real-world FrodoBots-2K sequences.
        Columns show successive $0.5$\,s chunks, and arrows show net grid-point
        displacement. Coherent egomotion patterns persist despite local motion
        and occlusion. Source video frames are from the FrodoBots-2K dataset,
        licensed under CC BY-SA 4.0; tracking overlays are ours.
    }
    \label{fig:tracked_motion_examples}
\end{figure*}

In order to get fundamental movement signals, we use a pretrained off-the-shelf pixel tracking model \cite{cotracker} on a fixed $10\times10$ grid of starting pixels indexed by $j=1,\dots,100$. For every $0.5$s chunk, represented by 3 latent frames spanning 12 pixel frames, the tracker follows and returns the net displacement $\delta = (\Delta x, \Delta y$) for each grid point across the chunk.
\begin{equation}
    m =
    \begin{bmatrix}
        \delta_1^\top &
        \delta_2^\top &
        \cdots &
        \delta_{100}^\top
    \end{bmatrix}^{\top}
    \in\mathbb{R}^{200}.
    \label{eq:movement_vector}
\end{equation}
The chunk is long enough for coherent displacement to exceed tracker noise,
but short enough that one approximately constant egomotion describes it and
successive commands may be composed autoregressively. Examples are shown in Figure~\ref{fig:tracked_motion_examples}.

Let $\xi_{s,n}\in\mathbb{R}^2$ denote the unobserved planar egomotion of
chunk $n$ in scene $s$. Over a short interval, its stacked image-plane
displacement can be locally represented as
\begin{equation}
    m_{s,n}
    =
    J(c_{s,n})\,\xi_{s,n}
    +
    \rho_{s,n}
    +
    \varepsilon_{s,n},
    \label{eq:tracked_motion_model}
\end{equation}
where $J(c_{s,n})$ maps the rover egomotion into scene-dependent image
motion, $\rho_{s,n}$ contains motion independent of the rover, and
$\varepsilon_{s,n}$ contains tracking error and higher-order effects. The operator $J(c)$ accounts for depth, perspective, camera pose,
foreshortening, visibility, and occlusion, and may vary substantially between
scenes. Linearity is assumed only with respect to the short-horizon
egomotion $\xi$ at a fixed scene; image formation itself is not assumed to
be globally linear.

Rover egomotion produces coherent global motion patterns that recur
through the dataset. Independent pedestrians, vegetation, and tracker
errors are more localized and vary across scenes. Pooling observations
therefore emphasizes recurring egomotion structure while dispersing
scene-specific residual motion.

Let $\bar m=\mathbb{E}[m]$ denote the training-set mean. We compute
\begin{equation}
    \begin{aligned}
        C
        &=
        \mathbb{E}
        \left[
            (m-\bar m)(m-\bar m)^\top
        \right],\\
        C u_k
        &=
        \lambda_k u_k.
    \end{aligned}
    \label{eq:motion_pca}
\end{equation}

Here, $\lambda_k$ is the variance captured by the $k$th principal component
$u_k$. Although the scene-dependent operators $J(c)$ need not share an identical
column space, the rover has only two commanded planar degrees of freedom and
their global flow patterns recur across scenes. We therefore expect their
contribution to concentrate in a small leading eigenspace when it dominates
independent object motion and tracker noise. This is a data-dependent
assumption rather than a property guaranteed by PCA; we verify it through
the explained variance and coherent flow structure of the recovered
components.

Let $U=[u_1,\ldots,u_d]$ contain the retained PCA directions. We project each
tracked-motion vector into this basis as
\begin{equation}
    p = U^\top(m-\bar m).
    \label{eq:pca_projection}
\end{equation}
The resulting scores are scaled using their training-set variation and passed
through a $\tanh$ function to obtain bounded actions $a\in(-1,1)^d$. This
preserves the centred, signed ordering of each PCA direction while reducing
the influence of extreme motions. The resulting action space is locally
linear near its centre and smoothly saturates toward its boundaries.

The PCA origin represents the mean tracked-motion field $\bar m$,
not physical stationarity. Encoding an all-zero displacement field
through the same pipeline gives
\begin{equation}
    p_{\mathrm{null}}=-U^\top\bar m,
    \qquad
    a_{\mathrm{null}}=(-0.0234,-0.0013).
    \label{eq:null_action}
\end{equation}
Thus, the physical no-egomotion command is slightly offset from the numerical
PCA origin. We use this correction only when explicitly requesting zero camera
motion; all active training and evaluation commands retain their original
encoded values. Throughout the geometric discussion, $\mathbf{0}$ denotes this
semantic no-egomotion command.

The raw scores form affine coordinates in the retained subspace; after
squashing, sign and ordering are preserved while scaling and composition are
local because of saturation. On FrodoBots, the leading component captures
forward--backward motion and explains approximately $60\%$ of tracked-motion
variance, while the second captures yaw and explains approximately $20\%$.
Visualisations of the recovered basis are provided in
Appendix~\ref{app:pca_components}.

\subsection{Latent Egomotion Critic without Decoder Gradients}
\label{sec:critic}

Flow matching alone does not ensure that generated video realises the requested motion, so we supervise rendered egomotion with an online critic; Section~\ref{sec:controllability} shows that removing the critic severely impairs command following. Let
$y\in\mathbb{R}^{8}$ denote the retained PCA motion code and
$a=y_{1:2}\in\mathbb{R}^{2}$ its throttle and yaw components. Only $a$ is
supplied to the DiT; the critic receives $y$ and predicts all eight components
to provide richer training supervision.

Direct supervision would require decoding, tracking, projecting, and
differentiating through the full pixel pipeline, greatly increasing memory and
potentially rewarding local pixel displacement rather than coherent
scene-level egomotion. We therefore use the decoder and tracker as a frozen,
no-gradient visual teacher.

At each training step, the current velocity prediction is converted into a
one step clean latent estimate $\hat{x}_{\mathrm{clean}}$ in pixel space. We apply the visual teacher to this estimate at every sampled flow timestep, without
completing the generation trajectory or any time based filtering.

Let $\mathcal{T}_8$ denote the frozen pixel-space readout consisting of
latent decoding, point tracking, displacement assembly, PCA projection,
scaling, and squashing from Section~\ref{sec:pca}. The motion rendered by the
current sample is measured as
\begin{equation}
    y^{\mathrm{teach}}
    =
    \mathcal{T}_8
    \left(
        \mathrm{sg}(\hat{x}_{\mathrm{clean}})
    \right)
    \in\mathbb{R}^{8},
    \label{eq:teacher_readout}
\end{equation}
where $\mathrm{sg}(\cdot)$ denotes stop-gradient. The supervisory label is
therefore obtained through a full pixel-space motion measurement, but this
pathway is forward-only: neither the decoder nor the tracker transmits
gradients to the generator.

A lightweight critic $D_\phi$ learns to predict the same motion code directly
from the same generated latent (in latent space, $\tau$ denotes the sampled flow timestep):
\begin{equation}
    \hat y_\phi
    =
    D_\phi
    \left(
        \mathrm{sg}(\hat{x}_{\mathrm{clean}}),\tau,y
    \right).
    \label{eq:critic_prediction}
\end{equation}
The critic also receives the sampled flow timestep because the quality and
statistics of $\hat{x}_{\mathrm{clean}}$ vary with respect to diffusion time.
\begin{equation}
    \mathcal{L}_{\mathrm{crit}}
    =
    \left\|
        W_c
        \left(
            \hat y_\phi-y^{\mathrm{teach}}
        \right)
    \right\|_2^2.
    \label{eq:critic_loss}
\end{equation}
The first two dimensions receive full weight because they correspond to the
commanded throttle and yaw axes. The remaining six receive reduced weight,
providing a richer description of rendered motion without expanding the
two-dimensional control interface. Because the critic
is trained online on samples from the generator before it is asked to supervise those same latents, its approximation is fresh and continuously on manifold.
\begin{equation}
    \mathcal{L}_{\mathrm{act}}
    =
    \left\|
        W_g
        \left(
            D_{\bar\phi}(\hat{x}_{\mathrm{clean}},\tau,y)
            -
            y
        \right)
    \right\|_2^2.
    \label{eq:latent_action_loss}
\end{equation}
Here, $D_{\bar\phi}$ denotes the current critic with its parameters held fixed
during the generator update and $y$ is the PCA code of the real training chunk from which
$a=y_{1:2}$ was obtained. The first two dimensions enforce the requested
throttle and yaw, while the remaining six are lower-weight auxiliary targets
for secondary camera and scene motion. Our pca4 and pca2 ablations show
that reducing this auxiliary supervision degrades geometry and fine detail.

The complete generator objective is
\begin{equation}
    \mathcal{L}_{\mathrm{gen}}
    =
    \mathcal{L}_{\mathrm{flow}}
    +
    \lambda_{\mathrm{act}}
    \mathcal{L}_{\mathrm{act}},
    \label{eq:generator_objective}
\end{equation}
Thus, the frozen teacher measures rendered motion in pixel space and the online
critic transfers this signal to the latent generator without decoder or tracker
gradients. Both are used only during training and add no inference-time cost.

We train with teacher-forced chunks and generate autoregressively,
allowing actions to change between chunks. Implementation details
appear in Appendix~\ref{app:implementation}.

\section{Evaluation}
\label{sec:eval}

\paragraph{Setup}
We train on ${\sim}88$ hours of FrodoBots video comprising $2{,}577$ rides and ${\sim}64$K windows across eleven cities. Genuine reverse motion is exceptionally rare: the full ${\sim}2{,}000$-hour corpus contains only $632$ sustained reverse windows (${\sim}55$ minutes), which we oversample $6\times$. Training dynamics are analysed on four fixed training rides using both recorded commands (GT) and 180 degree rotated counterfactual commands (FLIP).

Each training example contains seven temporal latent chunks and is decomposed
into seven teacher-forced next-chunk predictions, providing supervision at each
autoregressive transition. We train all variants for $5{,}000$ optimiser steps.
Global batch sizes of $16$, $32$, and $64$ correspond to approximately $1.25$,
$2.5$, and $5$ dataset-equivalent epochs, respectively. Unless otherwise
stated, \texttt{Default} uses a global batch size of $32$ and eight PCA
components for critic supervision; only the leading two components are exposed as throttle and yaw input controls.

For cross-model evaluation, we use $32$ low-egomotion contexts from held-out
rides, each continued under eight constant compass commands
$\uparrow$ (forward),$\downarrow$ (back),$\leftarrow$ (left),$\rightarrow$ (right),$\nearrow$ (forward right),$\searrow$ (back right),$\swarrow$ (back left),$\nwarrow$ (forward left), giving
$32\times8=256$ directional rollouts per model. Every model is rolled out over
a common six-second horizon at its native frame rate and through its native
autoregressive interface. We additionally generate one no-op continuation from
each context, giving $32$ stationary rollouts per model; these are evaluated
separately from the directional fleet.

\paragraph{Comparisons.}
We compare against state-of-the-art WMs including minWM~\cite{minwm}, WorldCam~\cite{worldcam},
Astra~\cite{astra}, Matrix-Game 2.0~\cite{matrixgame2},
HY-World 1.5 WorldPlay~\cite{hyworldplay}, and
Yume-1.5~\cite{yume}. Each compass command is translated into the model's native pose, keyboard–mouse, or text interface and held constant throughout the rollout. Because these interfaces differ in units and permitted command
strengths, we compare realized direction, forward--backward asymmetry, and
translation--steering composition rather than absolute control gain.
Full benchmark construction and baseline control mappings are provided in
Appendix~\ref{app:evaluation_protocol}, with model-specific interfaces detailed
in Section~\ref{app:baseline_controls}. We ablate conditioning pathways, critic supervision, supervision
width, and batch size. \texttt{No AdaLN} retains only action tokens;
\texttt{No Act Tok} retains only AdaLN action conditioning.
\texttt{No Critic} retains both pathways but trains the generator with
$\mathcal{L}_{\mathrm{flow}}$ alone. \texttt{pca4}/\texttt{pca2}
reduce critic supervision from eight to four/two PCA components,
leaving the two-dimensional control input unchanged.
\texttt{Batch16}/\texttt{Batch64} use global batch sizes of $16$/$64$
instead of the default $32$.

\subsection{Controllability}
\label{sec:controllability}

We measure every model with the same frozen CoTracker-PCA readout, which
projects fixed-grid displacements into signed throttle-yaw coordinates for
forward, backward, steering, and compound commands, allowing them to be evaluated within one representation. Existing image-based camera estimators cannot provide
this comparison reliably: they generally assume a rigid projective transform
and become unreliable when generated scenes deform or when several motion axes
must be measured simultaneously.

Directional control is evaluated with two complementary tests. CoTracker--PCA measures whether realised motion follows the commanded direction, while ORB feature matching with RANSAC homography verification detects near-stasis. Matching uses $3000$ keypoints at $640\times352$ resolution and a $5$-pixel homography tolerance; a rollout is treated as near-static when more than $500$ verified correspondences remain between its first generated frame and its six-second frame.

The same readout also supplies our training labels, but the resulting advantage
is strictly bounded: evaluation is performed using a fixed,
non-parametric pipeline applied unchanged to every model. Training on this space
does not guarantee that a model will learn it.
Moreover, the readout requires only unlabelled video and remains available equally
to other world-model systems as either a supervision signal or an evaluation
instrument.

\begin{figure}[t]
\centering
\includegraphics[
    width=0.8\columnwidth
]{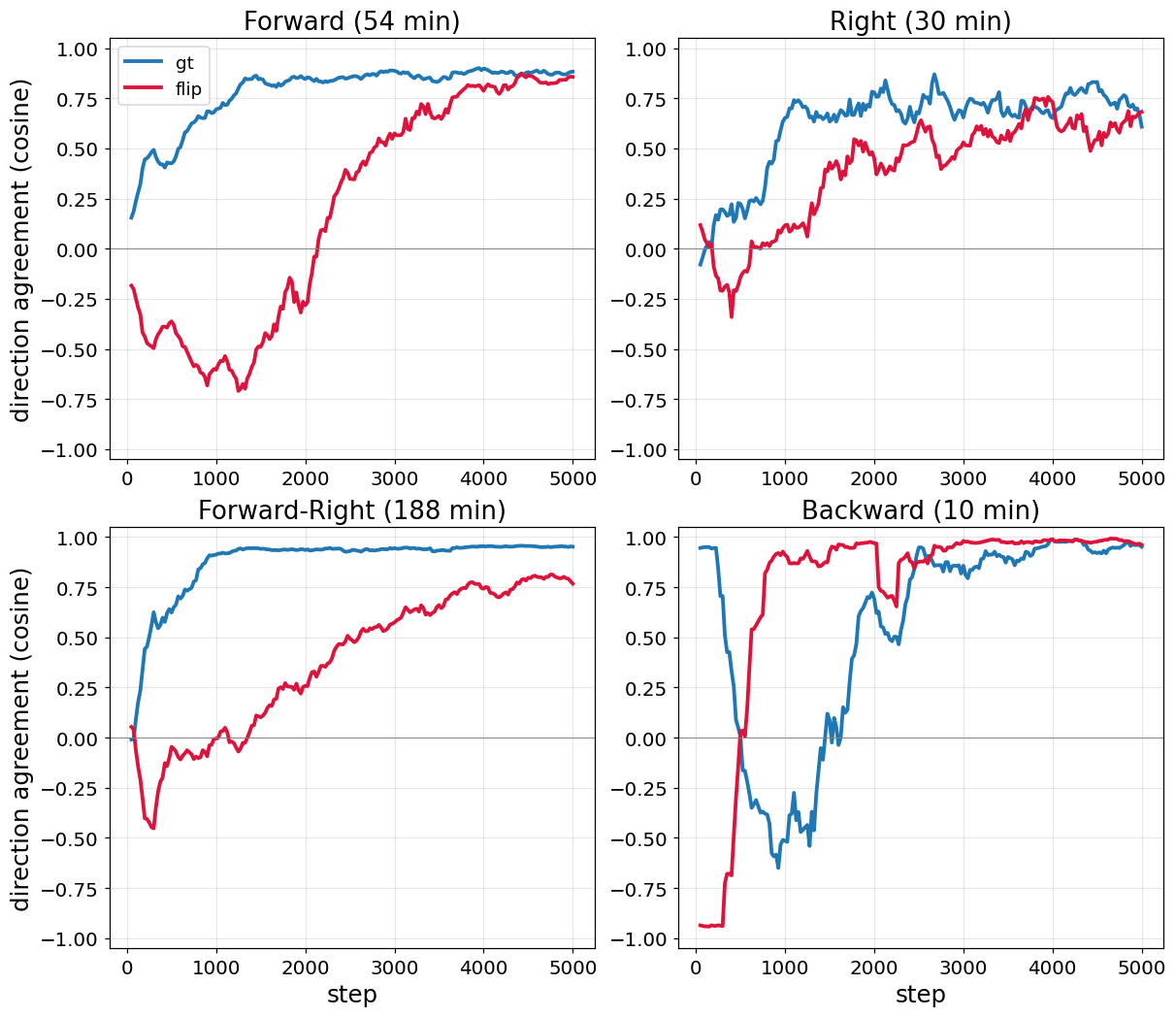}
\caption{
    Training-time cosine agreement for recorded (GT) and sign-reversed (FLIP)
    commands. Positive FLIP agreement indicates genuine command response;
    panel titles give source-direction duration in minutes. The full
    eight-direction result appears in Fig.~\ref{fig:following_8pca}.
}
\label{fig:following_dir}
\end{figure}

\paragraph{GT continuation can masquerade as control.}
Figure~\ref{fig:following_dir} compares recorded commands (GT) with
counterfactual commands obtained by reversing both action axes (FLIP). GT agreement is high early because the pretrained model continues motion from
the context. Backward motion is the exception: although reverse errors are penalised by the
squared loss, reverse examples are so rare that the aggregate objective can
improve while this branch remains underfit
(Fig.~\ref{fig:following_8pca}). FLIP instead tests whether the command
redirects that continuation:
steering emerges quickly, whereas forward--backward control appears later.
The diagonal responses show that throttle and steering are composed rather
than learned as unrelated categories. Extended training curves, conditioning ablations, settled responses, and
additional cross-model analyses appear in
Appendix~\ref{app:extended_control}.

\begin{figure}[t]
    \centering
    \begin{subfigure}[t]{\columnwidth}
        \centering
        \includegraphics[
            width=0.7\linewidth
        ]{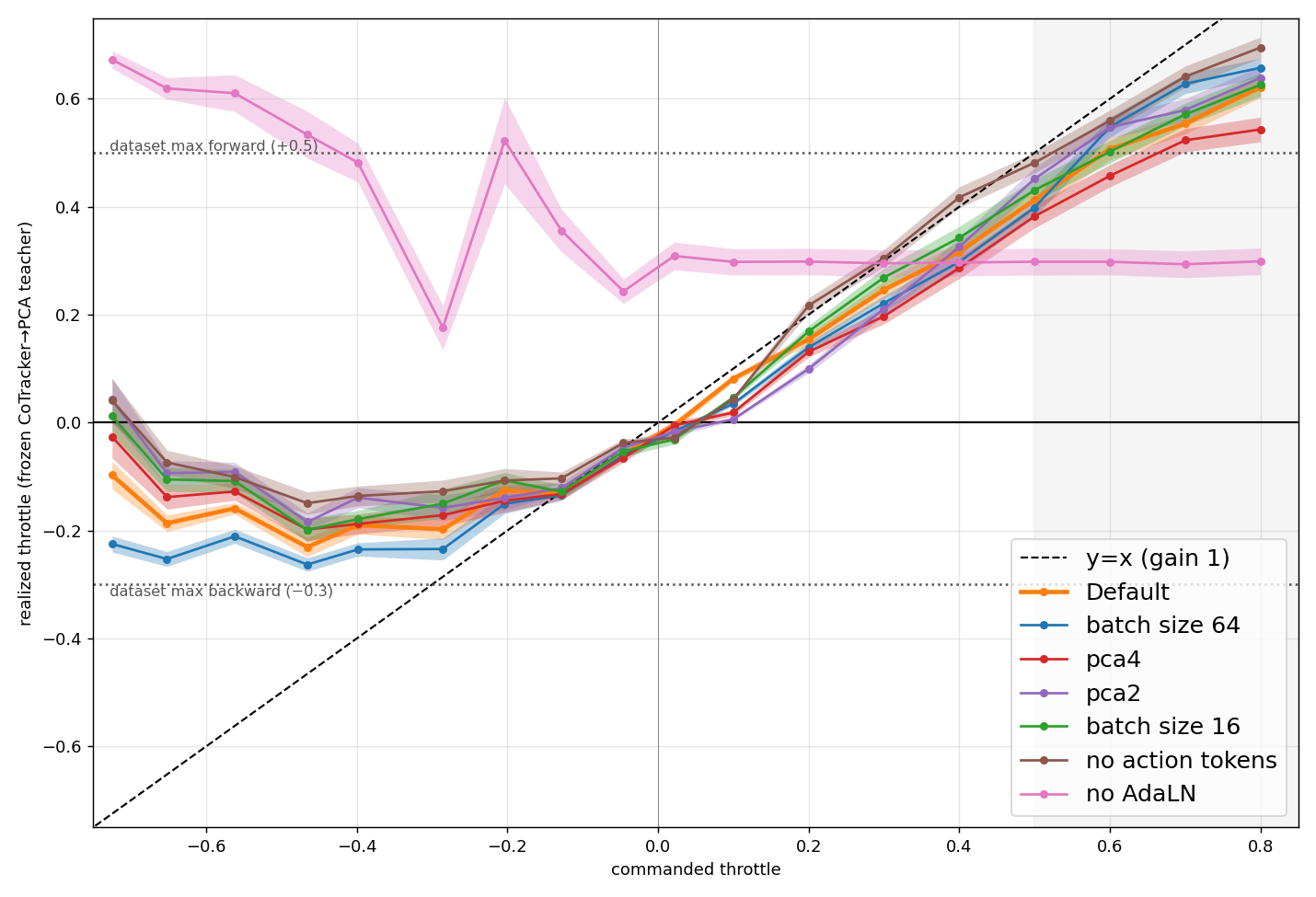}
        \caption{Forward and backward response.}
        \label{fig:response_forward}
    \end{subfigure}

    \begin{subfigure}[t]{\columnwidth}
        \centering
        \includegraphics[
            width=0.7\linewidth
        ]{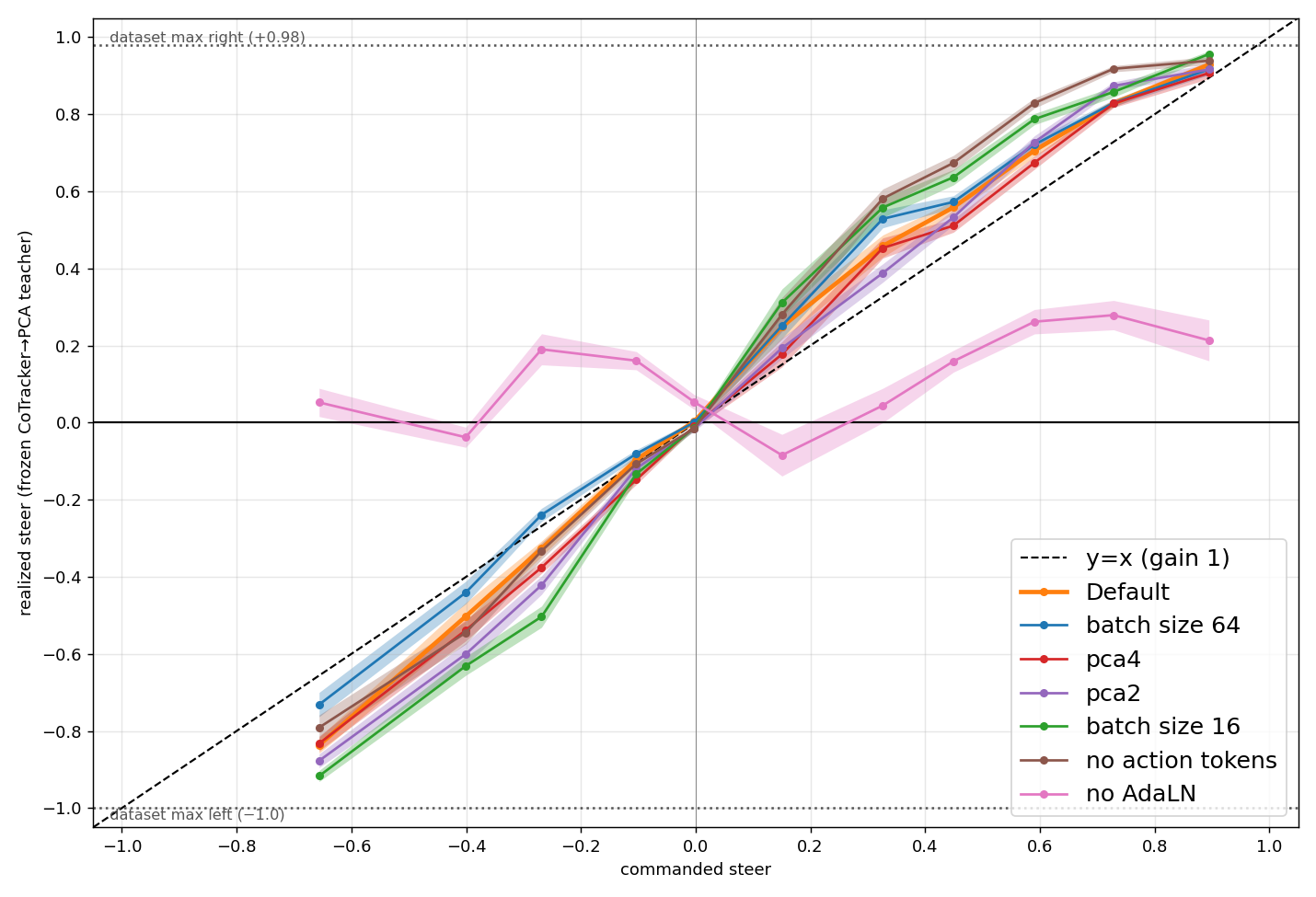}
        \caption{Steering response.}
        \label{fig:response_steer}
    \end{subfigure}

    \caption{
        Settled throttle and steering responses of the ablation variants, measured
        by the CoTracker--PCA readout. Most points use sign-flipped commands on
        held-out windows; positive throttle instead uses sampled forward commands
        on held-out no-op contexts. Dashed lines mark the approximate training
        support.
    }
    \label{fig:response_flip}
\end{figure}

\paragraph{Response follows the data support.}
Figure~\ref{fig:response_flip} shows approximately linear steering, while
throttle saturates outside the observed training range. As batch size or the number of critic-supervised PCA
dimensions is reduced, the reverse curves also bend back toward zero at
progressively larger negative command magnitudes. This highlights a limitation of our data-derived
approach: response strength depends strongly on the coverage of the observed
motion distribution. Conversely, the absence of the same contraction on the forward branch is consistent with its much denser representation in the training data.
Nevertheless, every functional variant recovers the sign of backward motion
despite its extreme scarcity.

\begin{figure}[t]
\centering
\includegraphics[
    width=\columnwidth
]{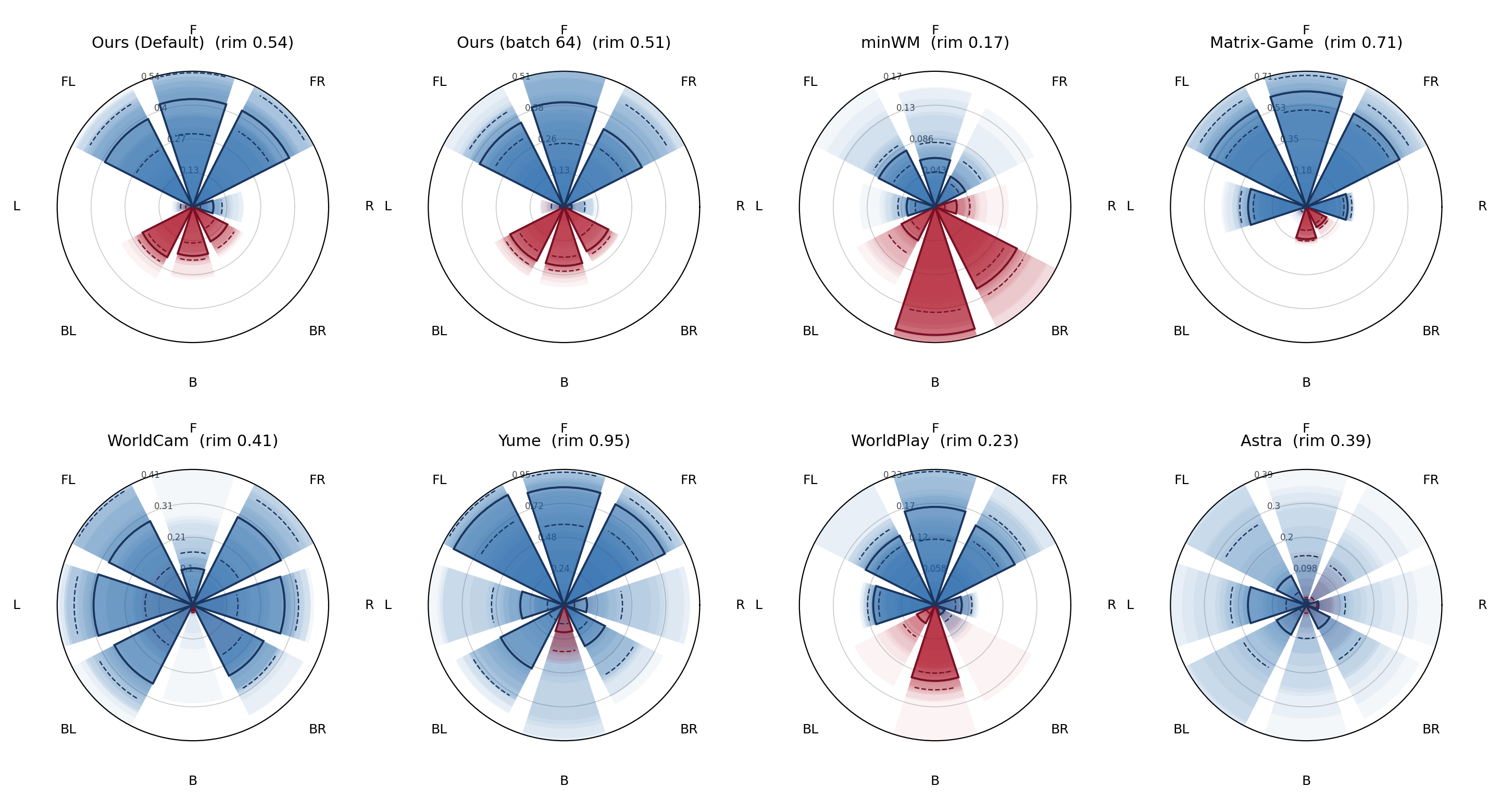}
\caption{
    Realised throttle $g_0$ under eight commands. Each wedge is a collation of one direction; opacity shows density, solid outlines medians, and dashed arcs
    interquartile ranges. Blue denotes forward and red backward motion. Panels
    use separate $95$th-percentile rims, so compare sign and shape rather than
    absolute radius.
}
\label{fig:wedge_all_g0}
\end{figure}

\paragraph{Backward composition distinguishes the models.}
Figure~\ref{fig:wedge_all_g0} applies the same eight-direction action-forcing
test to all models. Some baselines (Astra, WorldCam) fail to produce backward motion at all.
Others reverse under the pure backward command but lose or flip that component
when steering is added, showing that backward, backward-left, and
backward-right are not represented compositionally. Our model also reflects
the severe forward--backward data imbalance, producing asymmetric wedges,
but it gives the strongest and most consistent reverse response across both
pure and diagonal commands. minWM also preserves the correct structure, but
with substantially weaker realized motion.

\begin{figure}[t]
\centering
\includegraphics[width=\columnwidth]{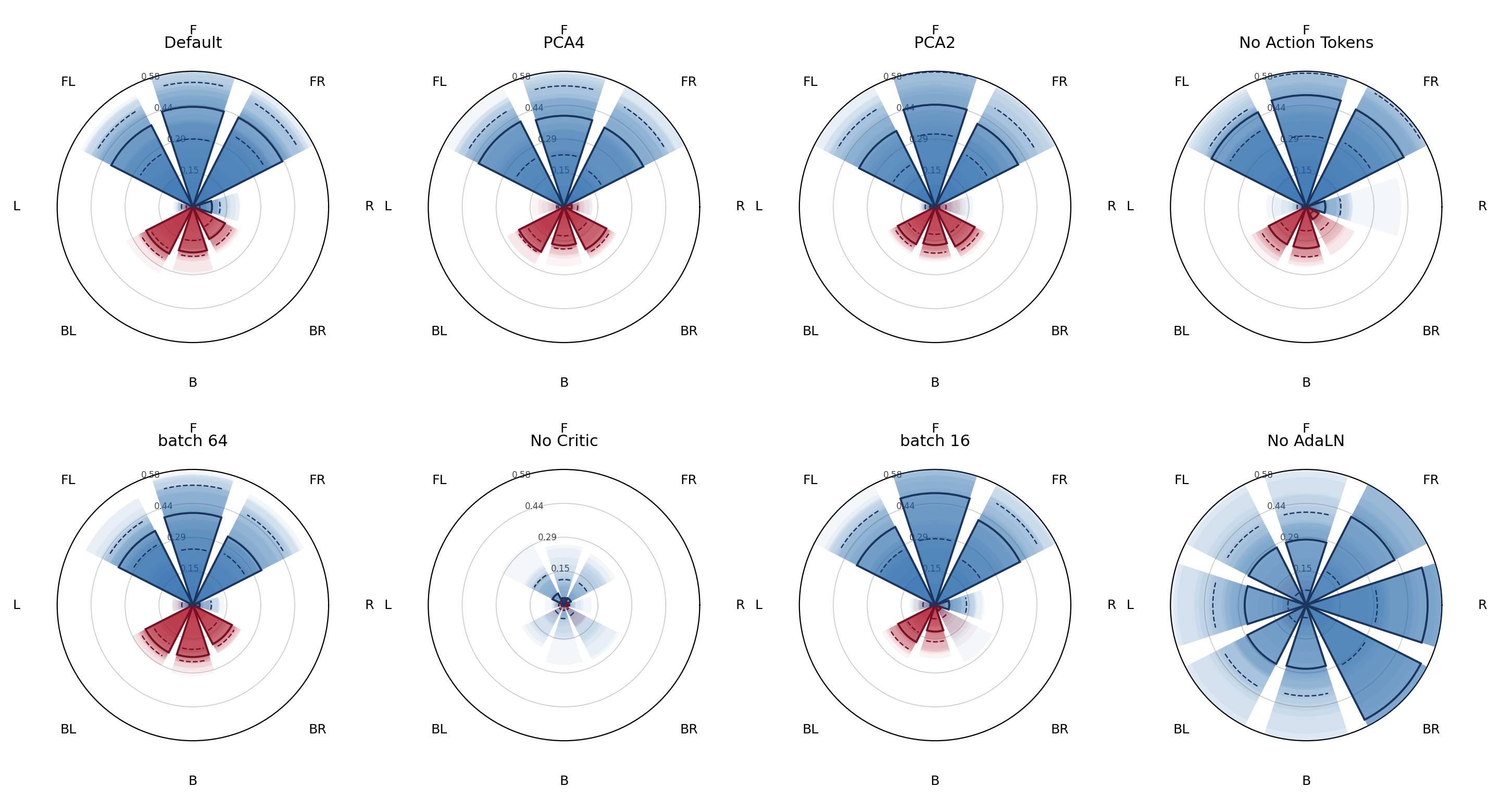}
\caption{
    Wedge plots for the ablation suite. The
    \texttt{No Critic} and the \texttt{No AdaLN} model exhibit markedly weaker and less coherent
    realised responses than the other models, showing that control performance is strongly dependent on these axes.
}
\label{fig:no_critic_wedges}
\end{figure}

\paragraph{Control failure by command.}
Table~\ref{tab:control_external} shows that backward control is the main failure mode for WorldCam, Astra and Yume, which fail $93\%$, $87\%$ and $40\%$ of backward commands respectively. In contrast, minWM struggles across all directions, with $84/87$ control failures caused by near-stasis.

Table~\ref{tab:control_ablation} meanwhile shows that removing AdaLN or
critic supervision raises overall control failure from $3\%$ to $58\%$
and $88\%$, respectively, demonstrating that both are critical for
reliable action following. The remaining ablations retain low overall
failure ($4$--$8\%$), but reduced PCA supervision, smaller batches and
removal of action tokens weaken reversal: backward failure reaches
$27$--$37\%$, versus $7\%$ for \texttt{Default} and $10\%$ for
\texttt{Batch64}.

\begin{table}[t]

\centering
{\footnotesize
\setlength{\tabcolsep}{3pt}
\begin{tabular}{@{}l*{7}{r}@{}}
\toprule
Command       & \rotatebox{90}{Default} & \rotatebox{90}{WorldPlay} & \rotatebox{90}{Matrix-Game} & \rotatebox{90}{WorldCam} & \rotatebox{90}{Astra} & \rotatebox{90}{Yume} & \rotatebox{90}{minWM} \\
\midrule
$\uparrow$    & 7 & 13 & 7 & 23 & 67 & 3  & 67 \\
$\downarrow$  & 7 & 3  & 0 & 93 & 87 & 40 & 60 \\
$\leftarrow$  & 3 & 0  & 0 & 0  & 7  & 0  & 23 \\
$\rightarrow$ & 3 & 0  & 0 & 0  & 3  & 3  & 30 \\
$\nwarrow$    & 7 & 0  & 0 & 3  & 3  & 0  & 27 \\
$\nearrow$    & 0 & 0  & 3 & 0  & 3  & 0  & 27 \\
$\swarrow$    & 0 & 0  & 0 & 60 & 43 & 63 & 27 \\
$\searrow$    & 0 & 0  & 0 & 77 & 33 & 63 & 30 \\
\midrule
All           & 3 & 2  & 1 & 32 & 31 & 22 & 36 \\
\bottomrule
\end{tabular}}
\caption{
    Control failure by command (\%) for our flagship and the external baselines; lower is better.
}
\label{tab:control_external}

\end{table}

\begin{table}[t]

\centering
{\footnotesize
\setlength{\tabcolsep}{3pt}
\begin{tabular}{@{}l*{8}{r}@{}}
\toprule
Command       & \rotatebox{90}{Default} & \rotatebox{90}{pca4} & \rotatebox{90}{pca2} & \rotatebox{90}{Batch64} & \rotatebox{90}{Batch16} & \rotatebox{90}{No Act Tok} & \rotatebox{90}{No AdaLN} & \rotatebox{90}{No Critic} \\
\midrule
$\uparrow$    & 7 & 3  & 10 & 17 & 7  & 7  & 47 & 87 \\
$\downarrow$  & 7 & 27 & 33 & 10 & 27 & 37 & 97 & 100 \\
$\leftarrow$  & 3 & 0  & 0  & 0  & 0  & 3  & 77 & 97 \\
$\rightarrow$ & 3 & 0  & 0  & 3  & 0  & 0  & 13 & 80 \\
$\nwarrow$    & 7 & 3  & 7  & 7  & 7  & 7  & 40 & 77 \\
$\nearrow$    & 0 & 0  & 0  & 0  & 0  & 0  & 7  & 77 \\
$\swarrow$    & 0 & 0  & 10 & 7  & 0  & 3  & 90 & 97 \\
$\searrow$    & 0 & 0  & 7  & 0  & 3  & 0  & 93 & 87 \\
\midrule
All           & 3 & 4  & 8  & 5  & 5  & 7  & 58 & 88 \\
\bottomrule
\end{tabular}}
\caption{
    Control failure by command (\%) for the ablation grid, with the same definitions and populations as Table~\ref{tab:control_external}.
}
\label{tab:control_ablation}

\end{table}

\subsection{Quality}
\label{sec:quality}

Action-forced rollouts do not have paired ground-truth continuations,
making standard reference-based video metrics unsuitable. We therefore
evaluate five complementary failure signals: \emph{style shift}
(changes in rendering appearance), \emph{geometric corruption}
(distorted or implausible scene structure), \emph{scene relocation}
(loss of the original place identity), \emph{conjuration}
(persistent objects appearing without a plausible visual history),
and \emph{progressive high-frequency degradation}
(loss of fine detail over time). Each instrument is selected
for its agreement with human judgement. Style and geometry are assessed against real context frames,
while conjuration is detected from within-rollout object histories.
Only HF and relocation use peers: HF normalises sharpness loss by
the same-input peer median to account for scene difficulty, while
relocation uses peer endpoints alongside real context frames as
alternative views for checking place identity. The reference contains
one representative per family, with each ablation replacing Default
when scored. Metric construction
is described in Appendix~\ref{app:quality_metrics}, and validation
against human annotations is reported in
Appendix~\ref{app:quality_validation}.

Table~\ref{tab:quality_all} shows distinct failure profiles. Yume preserves
style and geometry but frequently relocates, while minWM is visually stable
but often freezes or conjures objects. Our \texttt{Default} model records $3\%$ style shift, $17\%$
geometric corruption, $20\%$ relocation, $1\%$ conjuration, and $16\%$ high-frequency degradation. Reducing critic supervision to \texttt{pca4}/\texttt{pca2}
raises geometric corruption from $17\%$ to $38\%$/$40\%$, while
removing action tokens increases HF degradation from $16\%$ to
$29\%$. Smaller batches also lose detail: HF degradation is $44\%$
for \texttt{Batch16} versus $14\%$ for \texttt{Batch64}. Compared with \texttt{Default}, \texttt{Batch64} reduces relocation
($20\%$ to $17\%$) and HF degradation ($16\%$ to $14\%$),
but increases geometric corruption ($17\%$ to $26\%$).
Conversely, \texttt{No AdaLN}/\texttt{No Critic} reduce geometric
corruption to $11\%$/$2\%$, and \texttt{No Critic} lowers relocation
to $5\%$, but these gains in individual quality metrics accompany
poor action following (Table~\ref{tab:quality_ablation_main}).

\begin{table}[t]
\centering
{
\small
\setlength{\tabcolsep}{4pt}
\begin{tabular}{lrrrrr}
\toprule
Model          & Style\,\% & Geom.\,\% & Reloc.\,\%              & Conj.\,\% & HF\,\% \\
\midrule
WorldPlay      & 16 & 80 & 78 & 0   & 13 \\
Matrix-Game    & 62 & 71 & 96 & 0   & 9 \\
WorldCam       & 62 & 68 & 88                      & 0.4 & 45 \\
Astra          & 13 & 52 & 68 & 0.4 & 11 \\
Yume           & 5  & 9  & 59 & 0.4 & 13 \\
minWM          & 4  & 1  & 4                       & 20  & 5 \\
Ours (Default) & 3  & 17 & 20 & 1   & 16 \\
\bottomrule
\end{tabular}
}
\caption{
    Generation-quality failure rates (\%); lower is better.
}
\label{tab:quality_all}
\end{table}

\begin{table}[t]
\centering
{
\small
\setlength{\tabcolsep}{4pt}
\begin{tabular}{lrrrrr}
\toprule
Model & Style\,\% & Geom.\,\% & Reloc.\,\% & Conj.\,\% & HF\,\% \\
\midrule
Default & 3 & 17 & 20 & 1 & 16 \\
pca4 & 3 & 38 & 28 & 0 & 22 \\
pca2 & 3 & 40 & 25 & 0 & 28 \\
Batch64 & 3 & 26 & 17 & 0 & 14 \\
Batch16 & 5 & 32 & 42 & 1 & 44 \\
No Act Tok & 3 & 29 & 31 & 1 & 29 \\
No AdaLN & 4 & 11 & 30 & 2 & 53 \\
No Critic & 0 & 2 & 5 & 0 & 23 \\
\bottomrule
\end{tabular}
}
\caption{
    Generation-quality failure rates (\%) for the ablations. Each variant replaces Default as the sole family representative when scored.
}
\label{tab:quality_ablation_main}
\label{tab:quality_ablation}
\end{table}

Full degradation analyses are reported in Appendix~\ref{app:quality_results}.

\subsection{Stationarity}
\label{sec:stationarity}

A correct no-op should hold the viewpoint without freezing
independent scene dynamics. We evaluate the same $32$ contexts under
stationary commands, flagging rollouts whose CoTracker--PCA action
magnitude reaches $0.1$ in squashed units. We use the motion readout
rather than frame matching because geometric corruption can break
correspondences without moving the viewpoint. Stationary quality and
residual animation are evaluated separately in
Appendix~\ref{app:stationary_results}.

Thus, directional control does not automatically extend to
stationarity. Despite receiving stationary commands, Astra, Yume and
WorldCam move in $75\%$, $44\%$ and $22\%$ of rollouts, respectively
(Table~\ref{tab:stationary_external}). Default and all functional
ablations have no flagged no-op rollouts, whereas removing AdaLN raises
failure to $78\%$ (Table~\ref{tab:stationary_ablation}).
Taken together with minWM's static failures under motion commands,
these results highlight the value of a grounded, affine action space:
it supports interpolation between observed actions, enabling both a
calibrated no-op and composable directional control within the
training support.

\begin{table}[t]

\centering
{\footnotesize
\setlength{\tabcolsep}{3pt}
\begin{tabular}{@{}l*{8}{r}@{}}
\toprule
                & \rotatebox{90}{Default} & \rotatebox{90}{WorldPlay} & \rotatebox{90}{Matrix-Game} & \rotatebox{90}{WorldCam} & \rotatebox{90}{Astra} & \rotatebox{90}{Yume} & \rotatebox{90}{minWM} & \rotatebox{90}{REAL} \\
\midrule
No-op fail (\%) & 0 & 0 & 0 & 22 & 75 & 44 & 3 & 6 \\
\bottomrule
\end{tabular}}
\caption{
    No-op control failure (\%) over $32$ contexts, using the $0.1$ action-magnitude threshold defined in the text. REAL reports false positives from noisy action labelling on real no-op continuations, not a lower bound on model failure rates.
}
\label{tab:stationary_external}

\end{table}

\begin{table}[t]

\centering
{\footnotesize
\setlength{\tabcolsep}{3pt}
\begin{tabular}{@{}l*{8}{r}@{}}
\toprule
                & \rotatebox{90}{Default} & \rotatebox{90}{pca4} & \rotatebox{90}{pca2} & \rotatebox{90}{Batch64} & \rotatebox{90}{Batch16} & \rotatebox{90}{No Act Tok} & \rotatebox{90}{No AdaLN} & \rotatebox{90}{No Critic} \\
\midrule
No-op fail (\%) & 0 & 0 & 0 & 0 & 0 & 0 & 78 & 3 \\
\bottomrule
\end{tabular}}
\caption{
    Stationary control failure (\%) for the ablation grid, with the same definition as Table~\ref{tab:stationary_external}.
}
\label{tab:stationary_ablation}

\end{table}

\subsection{Combined Metrics}
\label{sec:combined}
We call a rollout \emph{legitimate} when it passes the
control criterion and all five quality tests.

\begin{figure}[t]

\centering

\begin{subfigure}{\columnwidth}
    \centering
    \includegraphics[width=0.8\columnwidth]{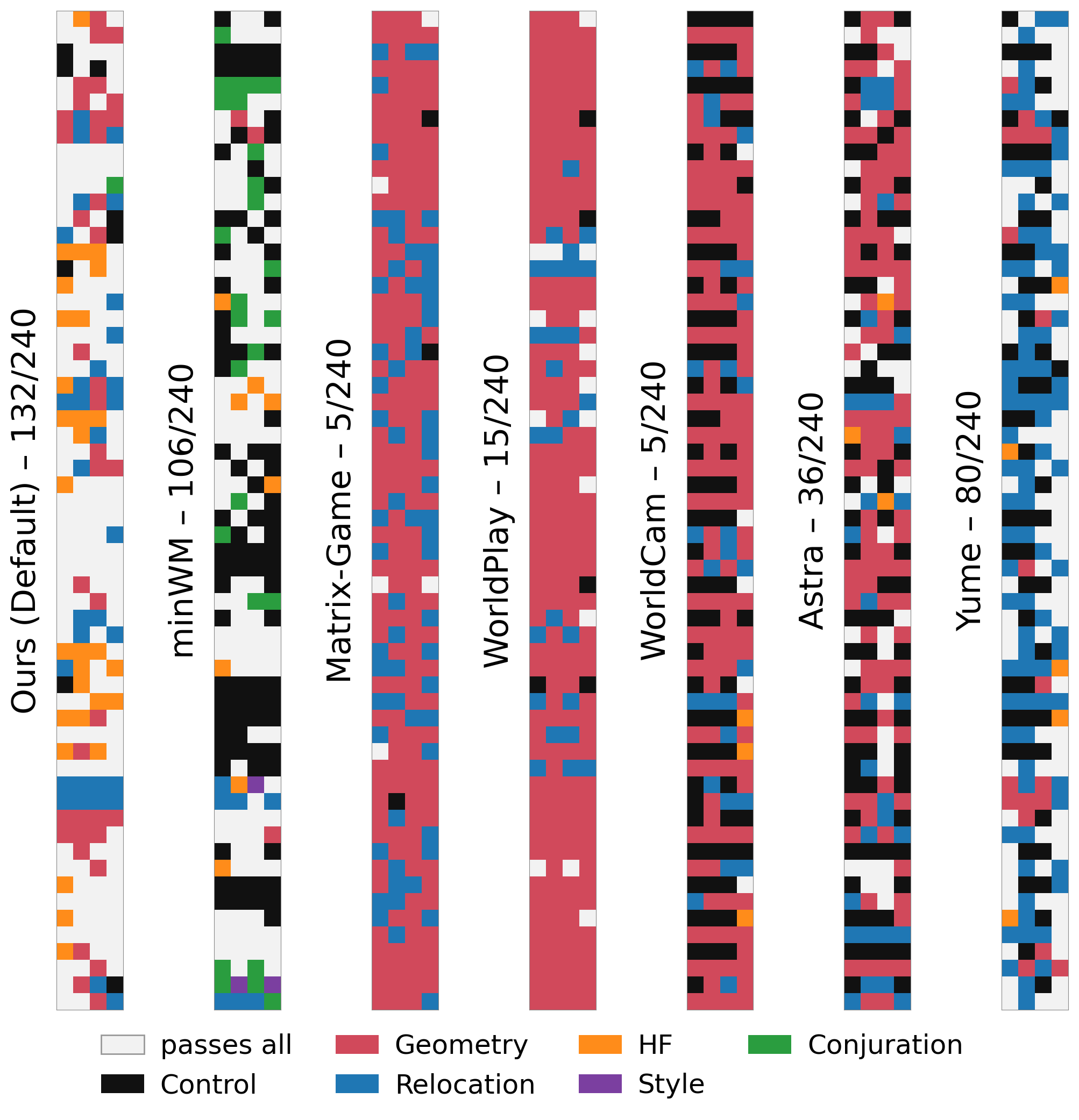}
    \phantomsubcaption
    \label{fig:failure_grid_external}
\end{subfigure}

\begin{subfigure}{\columnwidth}
    \centering
    \includegraphics[width=0.8\columnwidth]{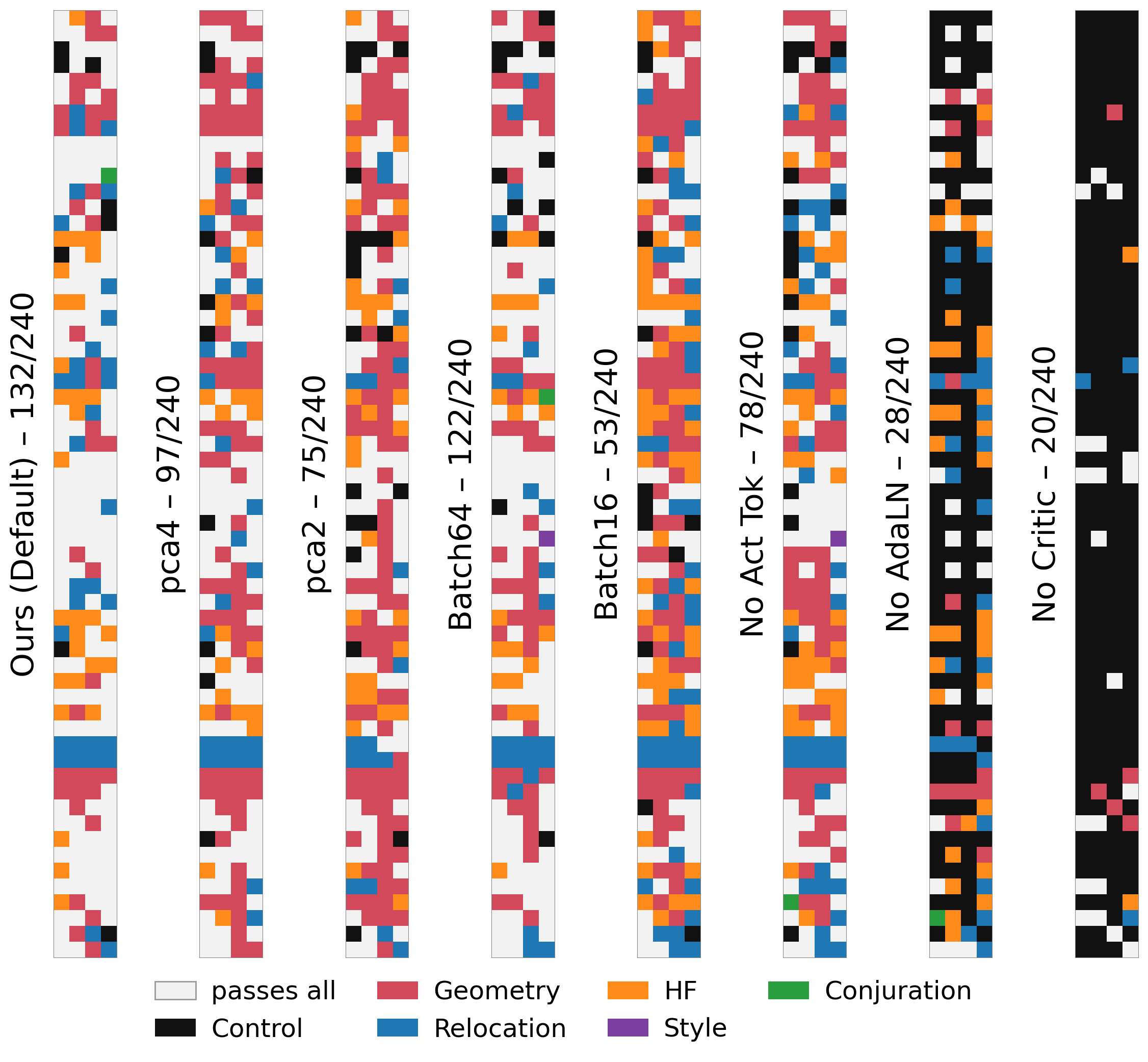}
    \phantomsubcaption
    \label{fig:failure_grid_ablations}
\end{subfigure}

\caption{
    Per-rollout failure profiles for (a) Default and external baselines and
    (b) our ablations. Neutral cells pass all criteria; coloTurs show the first
    failure in priority order: control $>$ geometry $>$ relocation $>$ HF $>$
    style $>$ conjuration. Headings report the number passing all criteria.
}

\end{figure}

Figures~\ref{fig:failure_grid_external}
and~\ref{fig:failure_grid_ablations} reveal distinct failure profiles.
minWM preserves visual quality comparatively well, but frequent control
failures (black) and conjuration (green) limit its usefulness as a
controllable world model. Geometric corruption (red) dominates the
displayed failures of Matrix-Game, WorldPlay and WorldCam, while
relocation (blue) dominates Yume; Astra combines geometry and control
failures. Default achieves the highest joint pass rate, represented
by light-grey cells, although HF degradation (orange) remains a
notable source of error.

These failure profiles also motivate our use of real-world
video. We hypothesise that third-person gameplay can encourage
foreground-actor priors underlying conjuration, while training on
generated video can propagate rendering biases and style shifts
across successive models. Learning from noisy, low-resolution footage
remains challenging, and our model still exhibits HF degradation.
However, these deployment challenges must ultimately be addressed;
synthetic-only training risks introducing additional biases rather
than resolving them.

\section{Conclusion \& Limitations}

Learning grounded actions from unlabelled video is challenging because observed
motion entangles camera egomotion with scene geometry, independent object
motion, and noise, yet solving this problem is essential for scaling world
models beyond costly instrumented datasets. We show that using a simple pixel tracker to derive pixel vector fields and finding their principal components results in recovering a reversible, composable, and linear action space with which we learn reliable backward control and composable steering despite scarce backward footage. Crucially, this approach can exploit the scale and diversity of real-world
footage rather than restricting world-model training to simulated environments
or videos collected within games.

Our method is both strengthened and limited by its core data dependency - any axes of egomotion present in the data can be recovered but motion not present in the data cannot be learnt in this manner. We also find that while our model interpolates well within the support provided (and better than the other baselines), extrapolation outside of the given support is somewhat lacking as shown by Fig~\ref{fig:response_forward}.

Action forcing evaluation also shows that conventional evaluation of reference-based quality is insufficient as a metric to evaluate world models. Our controllability wedges and quality metrics offer one potential manner in which WM evaluation can be done.

As a final takeaway, we wish to highlight that world model training no longer needs to be bounded by the quality of associated action labels.

\section*{Acknowledgements}

This work made use of the Isambard high-performance computing facility.
We thank the FrodoBots team for making the FrodoBots-2K dataset available
under the CC BY-SA 4.0 licence.

\clearpage
\newpage
\bibliographystyle{unsrtnat}
\bibliography{ActionForcing}

@article{relic,
  title={Relic: Interactive video world model with long-horizon memory},
  author={Hong, Yicong and Mei, Yiqun and Ge, Chongjian and Xu, Yiran and Zhou, Yang and Bi, Sai and Hold-Geoffroy, Yannick and Roberts, Mike and Fisher, Matthew and Shechtman, Eli and others},
  journal={arXiv preprint arXiv:2512.04040},
  year={2025}
}

@article{minwm,
  title={minWM: A full-stack open-source framework for real-time interactive video world models},
  author={Zhao, Min and Zhu, Hongzhou and Yan, Bokai and Zhou, Zihan and Chen, Yimin and Sun, Wenqiang and Zheng, Kaiwen and He, Guande and Yang, Xiao and Li, Chongxuan and others},
  journal={arXiv preprint arXiv:2605.30263},
  year={2026}
}

@article{internvl3,
  title={Internvl3: Exploring advanced training and test-time recipes for open-source multimodal models},
  author={Zhu, Jinguo and Wang, Weiyun and Chen, Zhe and Liu, Zhaoyang and Ye, Shenglong and Gu, Lixin and Tian, Hao and Duan, Yuchen and Su, Weijie and Shao, Jie and others},
  journal={arXiv preprint arXiv:2504.10479},
  year={2025}
}

@article{goyal2018focal,
  title={Focal loss for dense object detection},
  author={Goyal, Priya},
  journal={IEEE transactions on pattern analysis and machine intelligence},
  year={2018}
}

@inproceedings{girshick2015fast,
  title={Fast r-cnn},
  author={Girshick, Ross},
  booktitle={Proceedings of the IEEE international conference on computer vision},
  pages={1440--1448},
  year={2015}
}

@article{dinov2,
  title={Dinov2: Learning robust visual features without supervision},
  author={Oquab, Maxime and Darcet, Timoth{\'e}e and Moutakanni, Th{\'e}o and Vo, Huy and Szafraniec, Marc and Khalidov, Vasil and Fernandez, Pierre and Haziza, Daniel and Massa, Francisco and El-Nouby, Alaaeldin and others},
  journal={Transactions on Machine Learning Research Journal},
  year={2024}
}

@inproceedings{vqascore,
  title={Evaluating text-to-visual generation with image-to-text generation},
  author={Lin, Zhiqiu and Pathak, Deepak and Li, Baiqi and Li, Jiayao and Xia, Xide and Neubig, Graham and Zhang, Pengchuan and Ramanan, Deva},
  booktitle={European Conference on Computer Vision},
  pages={366--384},
  year={2024},
  organization={Springer}
}

@inproceedings{viescore,
  title={Viescore: Towards explainable metrics for conditional image synthesis evaluation},
  author={Ku, Max and Jiang, Dongfu and Wei, Cong and Yue, Xiang and Chen, Wenhu},
  booktitle={Proceedings of the 62nd Annual Meeting of the Association for Computational Linguistics (Volume 1: Long Papers)},
  pages={12268--12290},
  year={2024}
}

@inproceedings{zhao2024detrs,
  title={Detrs beat yolos on real-time object detection},
  author={Zhao, Yian and Lv, Wenyu and Xu, Shangliang and Wei, Jinman and Wang, Guanzhong and Dang, Qingqing and Liu, Yi and Chen, Jie},
  booktitle={Proceedings of the IEEE/CVF conference on computer vision and pattern recognition},
  pages={16965--16974},
  year={2024}
}

@inproceedings{gatys2016image,
  title={Image style transfer using convolutional neural networks},
  author={Gatys, Leon A and Ecker, Alexander S and Bethge, Matthias},
  booktitle={Proceedings of the IEEE conference on computer vision and pattern recognition},
  pages={2414--2423},
  year={2016}
}

@inproceedings{videophy,
  title={Videophy: Evaluating physical commonsense for video generation},
  author={Bansal, Hritik and Lin, Zongyu and Xie, Tianyi and Zong, Zeshun and Yarom, Michal and Bitton, Yonatan and Jiang, Chenfanfu and Sun, Yizhou and Chang, Kai-Wei and Grover, Aditya},
  booktitle={International Conference on Learning Representations},
  volume={2025},
  pages={102075--102121},
  year={2025}
}

@inproceedings{videoscore,
  title={Videoscore: Building automatic metrics to simulate fine-grained human feedback for video generation},
  author={He, Xuan and Jiang, Dongfu and Zhang, Ge and Ku, Max and Soni, Achint and Siu, Sherman and Chen, Haonan and Chandra, Abhranil and Jiang, Ziyan and Arulraj, Aaran and others},
  booktitle={Proceedings of the 2024 Conference on Empirical Methods in Natural Language Processing},
  pages={2105--2123},
  year={2024}
}

@inproceedings{videobench,
  title={Video-bench: Human-aligned video generation benchmark},
  author={Han, Hui and Li, Siyuan and Chen, Jiaqi and Yuan, Yiwen and Wu, Yuling and Deng, Yufan and Leong, Chak Tou and Du, Hanwen and Fu, Junchen and Li, Youhua and others},
  booktitle={Proceedings of the Computer Vision and Pattern Recognition Conference},
  pages={18858--18868},
  year={2025}
}

@inproceedings{lapa,
  title={Latent action pretraining from videos},
  author={Ye, Seonghyeon and Jang, Joel and Jeon, Byeongguk and Joo, Se June and Yang, Jianwei and Peng, Baolin and Mandlekar, Ajay and Tan, Reuben and Chao, Yu-Wei and Lin, Bill Yuchen and others},
  booktitle={International Conference on Learning Representations},
  volume={2025},
  pages={28213--28239},
  year={2025}
}

@inproceedings{pal,
  title={Perceptual artifacts localization for image synthesis tasks},
  author={Zhang, Lingzhi and Xu, Zhengjie and Barnes, Connelly and Zhou, Yuqian and Liu, Qing and Zhang, He and Amirghodsi, Sohrab and Lin, Zhe and Shechtman, Eli and Shi, Jianbo},
  booktitle={Proceedings of the IEEE/CVF International Conference on Computer Vision},
  pages={7579--7590},
  year={2023}
}

@article{QWEN,
  title={Qwen3-vl technical report},
  author={Bai, Shuai and Cai, Yuxuan and Chen, Ruizhe and Chen, Keqin and Chen, Xionghui and Cheng, Zesen and Deng, Lianghao and Ding, Wei and Gao, Chang and Ge, Chunjiang and others},
  journal={arXiv preprint arXiv:2511.21631},
  year={2025}
}

@article{williams1989learning,
  title={A learning algorithm for continually running fully recurrent neural networks},
  author={Williams, Ronald J and Zipser, David},
  journal={Neural computation},
  volume={1},
  number={2},
  pages={270--280},
  year={1989},
  publisher={MIT Press}
}

@article{adaworld,
  title={Adaworld: Learning adaptable world models with latent actions},
  author={Gao, Shenyuan and Zhou, Siyuan and Du, Yilun and Zhang, Jun and Gan, Chuang},
  journal={arXiv preprint arXiv:2503.18938},
  year={2025}
}

@inproceedings{lapo,
  title={Learning to act without actions},
  author={Schmidt, Dominik and Jiang, Minqi},
  booktitle={International Conference on Learning Representations},
  volume={2024},
  pages={9379--9395},
  year={2024}
}

@article{vid2world,
  title={Vid2world: Crafting video diffusion models to interactive world models},
  author={Huang, Siqiao and Wu, Jialong and Zhou, Qixing and Miao, Shangchen and Long, Mingsheng},
  journal={arXiv preprint arXiv:2505.14357},
  year={2025}
}

@inproceedings{dws,
  title={Pre-trained video generative models as world simulators},
  author={He, Haoran and Zhang, Yang and Lin, Liang and Xu, Zhongwen and Pan, Ling},
  booktitle={Proceedings of the AAAI Conference on Artificial Intelligence},
  volume={40},
  pages={4645--4653},
  year={2026}
}

@article{cosmos,
  title={Cosmos world foundation model platform for physical ai},
  author={Agarwal, Niket and Ali, Arslan and Bala, Maciej and Balaji, Yogesh and Barker, Erik and Cai, Tiffany and Chattopadhyay, Prithvijit and Chen, Yongxin and Cui, Yin and Ding, Yifan and others},
  journal={arXiv preprint arXiv:2501.03575},
  year={2025}
}

@article{hyworldplay,
  title={HY-World 1.5: A systematic framework for interactive world modeling with real-time latency and geometric consistency},
  author={HunyuanWorld, Team},
  journal={arXiv preprint},
  year={2025}
}

@article{yume,
  title={Yume: An interactive world generation model},
  author={Mao, Xiaofeng and Lin, Shaoheng and Li, Zhen and Li, Chuanhao and Peng, Wenshuo and He, Tong and Pang, Jiangmiao and Chi, Mingmin and Qiao, Yu and Zhang, Kaipeng},
  journal={arXiv preprint arXiv:2507.17744},
  year={2025}
}

@article{matrixgame2,
  title={Matrix-game 2.0: An open-source real-time and streaming interactive world model},
  author={He, Xianglong and Peng, Chunli and Liu, Zexiang and Wang, Boyang and Zhang, Yifan and Cui, Qi and Kang, Fei and Jiang, Biao and An, Mengyin and Ren, Yangyang and others},
  journal={arXiv preprint arXiv:2508.13009},
  year={2025}
}

@article{astra,
  title={Astra: General Interactive World Model with Autoregressive Denoising},
  author={Zhu, Yixuan and Feng, Jiaqi and Zheng, Wenzhao and Gao, Yuan and Tao, Xin and Wan, Pengfei and Zhou, Jie and Lu, Jiwen},
  journal={arXiv preprint arXiv:2512.08931},
  year={2025}
}

@article{worldcam,
  title={Worldcam: Interactive autoregressive 3d gaming worlds with camera pose as a unifying geometric representation},
  author={Nam, Jisu and Hong, Yicong and Huang, Chun-Hao Paul and Liu, Feng and Lee, JoungBin and Kim, Jiyoung and Jin, Siyoon and Lee, Yunsung and Jung, Jaeyoon and Choi, Suhwan and others},
  journal={arXiv preprint arXiv:2603.16871},
  year={2026}
}

@inproceedings{onlyflow,
  title={Onlyflow: Optical flow based motion conditioning for video diffusion models},
  author={Koroglu, Mathis and Caselles-Dupr{\'e}, Hugo and Jeanneret, Guillaume and Cord, Matthieu},
  booktitle={Proceedings of the Computer Vision and Pattern Recognition Conference},
  pages={6226--6236},
  year={2025}
}

@inproceedings{motionprompting,
  title={Motion prompting: Controlling video generation with motion trajectories},
  author={Geng, Daniel and Herrmann, Charles and Hur, Junhwa and Cole, Forrester and Zhang, Serena and Pfaff, Tobias and Lopez-Guevara, Tatiana and Aytar, Yusuf and Rubinstein, Michael and Sun, Chen and others},
  booktitle={Proceedings of the Computer Vision and Pattern Recognition Conference},
  pages={1--12},
  year={2025}
}

@inproceedings{flovd,
  title={Flovd: Optical flow meets video diffusion model for enhanced camera-controlled video synthesis},
  author={Jin, Wonjoon and Dai, Qi and Luo, Chong and Baek, Seung-Hwan and Cho, Sunghyun},
  booktitle={Proceedings of the Computer Vision and Pattern Recognition Conference},
  pages={2040--2049},
  year={2025}
}

@inproceedings{gamengen2024,
  title={Diffusion Models Are Real-Time Game Engines},
  author={Valevski, Dani and Leviathan, Yaniv and Arar, Moab and Fruchter, Shlomi},
  booktitle={International Conference on Learning Representations (ICLR)},
  year={2025}
}

@article{mineworld,
  title={Mineworld: a real-time and open-source interactive world model on minecraft},
  author={Guo, Junliang and Ye, Yang and He, Tianyu and Wu, Haoyu and Jiang, Yushu and Pearce, Tim and Bian, Jiang},
  journal={arXiv preprint arXiv:2504.08388},
  year={2025}
}

@inproceedings{gamefactory,
  title={Gamefactory: Creating new games with generative interactive videos},
  author={Yu, Jiwen and Qin, Yiran and Wang, Xintao and Wan, Pengfei and Zhang, Di and Liu, Xihui},
  booktitle={Proceedings of the IEEE/CVF International Conference on Computer Vision},
  pages={11590--11599},
  year={2025}
}

@article{diamond,
  title={Diffusion for world modeling: Visual details matter in atari},
  author={Alonso, Eloi and Jelley, Adam and Micheli, Vincent and Kanervisto, Anssi and Storkey, Amos and Pearce, Tim and Fleuret, Fran{\c{c}}ois},
  journal={Advances in Neural Information Processing Systems},
  volume={37},
  pages={58757--58791},
  year={2024}
}

@inproceedings{navwormod,
  title={Navigation World Models},
  author={Bar, Amir and Zhou, Gaoyue and Tran, Danny and Darrell, Trevor and LeCun, Yann},
  booktitle={Proceedings of the IEEE/CVF Conference on Computer Vision and Pattern Recognition (CVPR)},
  pages={15791--15801},
  month={June},
  year={2025}
}

@article{uniwm,
  title={Towards Unified World Models for Visual Navigation via Memory-Augmented Planning and Foresight},
  author={Dong, Yifei and Wu, Fengyi and Chen, Guangyu and Kong, Lingdong and Zhu, Xu and Hu, Qiyu and Zhou, Yuxuan and Sun, Jingdong and He, Jun-Yan and Dai, Qi and Hauptmann, Alexander G. and Cheng, Zhi-Qi},
  journal={arXiv preprint arXiv:2510.08713},
  year={2025}
}

@inproceedings{genie,
  title={Genie: Generative Interactive Environments},
  author={Bruce, Jake and Dennis, Michael D. and Edwards, Ashley and Parker-Holder, Jack and Shi, Yuge and Hughes, Edward and Lai, Matthew and Mavalankar, Aditi and Steigerwald, Richie and Apps, Chris and Aytar, Yusuf and Bechtle, Sarah Maria Elisabeth and Behbahani, Feryal and Chan, Stephanie C. Y. and Heess, Nicolas and Gonzalez, Lucy and Osindero, Simon and Ozair, Sherjil and Reed, Scott and Zhang, Jingwei and Zolna, Konrad and Clune, Jeff and de Freitas, Nando and Singh, Satinder and Rockt{\"a}schel, Tim},
  booktitle={Proceedings of the 41st International Conference on Machine Learning},
  pages={4603--4623},
  year={2024},
  volume={235},
  series={Proceedings of Machine Learning Research},
  publisher={PMLR}
}

@misc{frodobots,
  title        = {{FrodoBots-2K Dataset}},
  author       = {{FrodoBots Lab} and Cho, Michael and Cho, Sam and Tung, Aaron and Dravin, Niresh and Pravisani, Santiago},
  year         = {2024},
  publisher    = {Hugging Face},
  howpublished = {\url{https://huggingface.co/datasets/frodobots/FrodoBots-2K}},
  doi          = {10.57967/hf/3042},
  note         = {Licensed under CC-BY-SA-4.0}
}

@inproceedings{cameracontrol2,
  title={CameraCtrl II: Dynamic Scene Exploration via Camera-controlled Video Diffusion Models},
  author={He, Hao and Yang, Ceyuan and Lin, Shanchuan and Xu, Yinghao and Wei, Meng and Gui, Liangke and Zhao, Qi and Wetzstein, Gordon and Jiang, Lu and Li, Hongsheng},
  booktitle={Proceedings of the IEEE/CVF International Conference on Computer Vision (ICCV)},
  pages={13416--13426},
  month={October},
  year={2025}
}

@inproceedings{cotracker,
  title={CoTracker3: Simpler and Better Point Tracking by Pseudo-Labelling Real Videos},
  author={Karaev, Nikita and Makarov, Yuri and Wang, Jianyuan and Neverova, Natalia and Vedaldi, Andrea and Rupprecht, Christian},
  booktitle={Proceedings of the IEEE/CVF International Conference on Computer Vision (ICCV)},
  pages={6013--6022},
  month={October},
  year={2025}
}

@article{cosmos2,
  title={World simulation with video foundation models for physical ai},
  author={Ali, Arslan and Bai, Junjie and Bala, Maciej and Balaji, Yogesh and Blakeman, Aaron and Cai, Tiffany and Cao, Jiaxin and Cao, Tianshi and Cha, Elizabeth and Chao, Yu-Wei and others},
  journal={arXiv preprint arXiv:2511.00062},
  year={2025}
}

@article{olafworld,
  title={Olaf-World: Orienting Latent Actions for Video World Modeling},
  author={Jiang, Yuxin and Gu, Yuchao and Tsang, Ivor W and Shou, Mike Zheng},
  journal={arXiv preprint arXiv:2602.10104},
  year={2026}
}

@article{oasis2,
  title={Worldmem: Long-term consistent world simulation with memory},
  author={Xiao, Zeqi and Lan, Yushi and Zhou, Yifan and Ouyang, Wenqi and Yang, Shuai and Zeng, Yanhong and Pan, Xingang},
  journal={Advances in Neural Information Processing Systems},
  volume={38},
  pages={49632--49652},
  year={2026}
}

@article{xworld,
  title={X-World: Controllable Ego-Centric Multi-Camera World Models for Scalable End-to-End Driving},
  author={Zheng, Chaoda and Li, Sean and Deng, Jinhao and Wang, Zhennan and Chen, Shijia and Xiao, Liqiang and Chi, Ziheng and Lin, Hongbin and Chen, Kangjie and Wang, Boyang and others},
  journal={arXiv preprint arXiv:2603.19979},
  year={2026}
}

@article{gaia2,
  title={Gaia-2: A controllable multi-view generative world model for autonomous driving},
  author={Russell, Lloyd and Hu, Anthony and Bertoni, Lorenzo and Fedoseev, George and Shotton, Jamie and Arani, Elahe and Corrado, Gianluca},
  journal={arXiv preprint arXiv:2503.20523},
  year={2025}
}
\clearpage
\newpage


\appendix

\newpage
\section{Example Generations}
\label{app:ex}

\begin{figure*}[t]
    \centering
    \includegraphics[
        width=0.95\textwidth
    ]{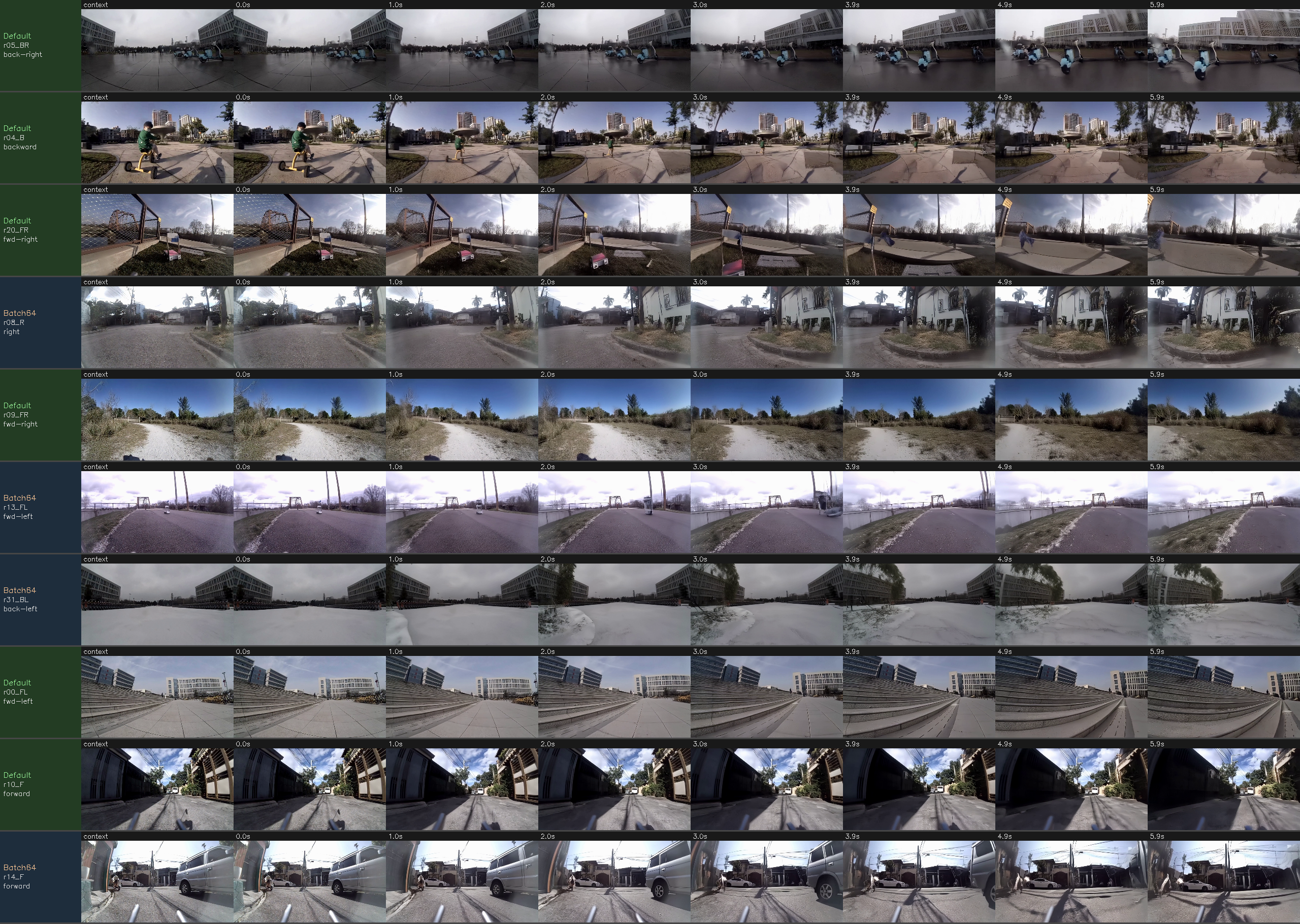}
    \caption{
        Representative rollouts across environments and cities. The commanded
        action is shown in the leftmost panel of each row.
    }
    \label{fig:ex}
\end{figure*}

Representative continuations seeded with approximately $0.5$ seconds of real
video are shown in Figure~\ref{fig:ex}.

\section{Recovered Motion Basis}
\label{app:pca_components}

\begin{figure*}[t]
    \centering
    \includegraphics[
        width=0.95\textwidth
    ]{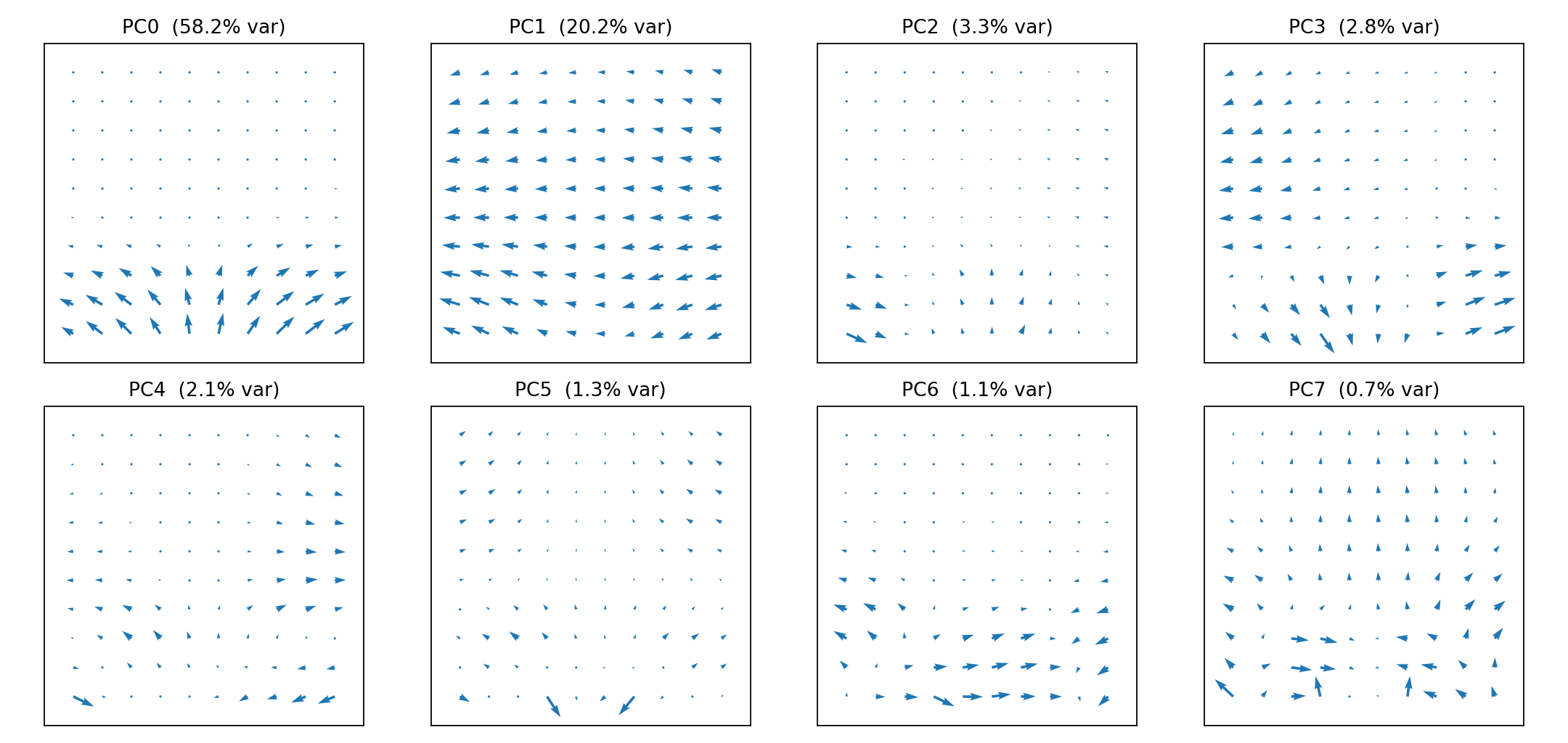}
    \caption{
        Principal motion components recovered from the fixed
        CoTracker--PCA pipeline. Each panel shows the displacement field
        associated with one PCA coordinate. The exposed control space uses
        $g_0$ for longitudinal motion and $g_1$ for yaw; $g_6$ captures
        roll-like motion. The remaining components describe secondary
        camera and scene dynamics and are used only for critic supervision.
    }
    \label{fig:pca_act}
\end{figure*}

Figure~\ref{fig:pca_act} visualizes the leading motion directions recovered
by the fixed CoTracker--PCA pipeline. The first component, $g_0$, captures
longitudinal forward--backward motion; $g_1$ captures yaw; and $g_6$ captures
roll-like camera motion.

\section{Implementation Details}
\label{app:implementation}

Each training video contains seven chunks and we train using the teacher forcing paradigm \cite{williams1989learning}. We decompose the seven chunk sequence into seven teacher-forced prediction sequences of varying lengths, each targeting one chunk with a different
amount of preceding ground-truth context.

When inference is run with 48 diffusion steps like in training, it takes 86s (on a GH200) to generate 7 chunks (or roughly 6 seconds) of content autoregressively, which at 16 FPS gives a generation speed of 1.11 FPS. Since Wan is a flow model, 20 diffusion steps suffice to generate with minimal loss in quality which subsequently increases the generation FPS to 2.72 FPS.

\paragraph{Critic conditioning and targets.}
The training configurations use PCA components 0 and 1 for throttle and
steering and components 0--7 for the critic; in Section~\ref{sec:critic}, these
are indexed as $a=y_{1:2}$ and $y\in\mathbb{R}^{8}$.
The target $y$ is the real training window's per-frame PCA projection,
scaled and tanh-squashed componentwise, then mean-pooled within the target
chunk. Only $a$ conditions the DiT. The critic receives the full $y$, the
latent estimate and the sampled flow timestep in both updates.

During the critic update, its target is $y^{\mathrm{teach}}$, obtained by
applying the frozen decoder--tracker--PCA teacher to the detached generated
latent. It is therefore trained to predict rendered motion, not to copy its
conditioning vector. During the generator update, the critic parameters are
frozen and its outputs are regressed onto the real-window target $y$:
\[
\begin{aligned}
\hat y &= D_{\bar\phi}(\hat{x}_{\mathrm{clean}},\tau,y),\\
\mathcal L_{\mathrm{act}}
&= \operatorname{MSE}(\hat y_{1:2},y_{1:2})\\
&\quad + 0.3333\,\operatorname{MSE}(\hat y_{3:8},y_{3:8}).
\end{aligned}
\]
The first term supervises the commanded throttle and steering; the second
uses the real continuation's six auxiliary motion components. Both use the
same PCA basis and tanh scales as the teacher. The matrix $W_g$ in
Eq.~\ref{eq:latent_action_loss} incorporates these weights and mean reductions.
The overall coefficient is $\lambda_{\mathrm{act}}=0.3$ in
Eq.~\ref{eq:generator_objective} and $0$ for No Critic; it is applied once.

\section{Cross-Model Evaluation Protocol}
\label{app:evaluation_protocol}

\subsection{Action-Forcing Benchmark}
\label{app:action_forcing_details}

\paragraph{Low-egomotion context selection.}
The benchmark uses $32$ contexts drawn from distinct held-out FrodoBots rides.
For each candidate window $w$, we compute the mean ego-motion magnitude across
its $N_w$ chunks:
\[
e(w)
=
\frac{1}{N_w}
\sum_{t=1}^{N_w}
\sqrt{
a_{t,\mathrm{throttle}}^2
+
a_{t,\mathrm{yaw}}^2
},
\]
where $a_{t,\mathrm{throttle}}$ and $a_{t,\mathrm{yaw}}$ are the
tanh-squashed PCA action coordinates for chunk $t$.

Let $q_{25}$ denote the $25$th percentile of the ego-motion scores across all
candidate windows. We define the eligibility threshold as
\[
\tau_{\mathrm{stat}}
=
\min(0.12,q_{25}),
\]
and retain candidate windows satisfying
\[
e(w)\leq\tau_{\mathrm{stat}}.
\]
This selects at most the lowest-motion quarter of the candidate population
while also imposing an absolute upper limit of $0.12$ in squashed PCA units.

The selection criterion restricts camera ego-motion rather than all visible
motion. A context may still contain independently moving pedestrians,
vehicles, vegetation, or other scene elements. A separate point-track
scene-activity score is used elsewhere for evaluation but is not part of
this low-egomotion selection criterion.

For our model, actions are ordered as
\[
a=(a_{\mathrm{throttle}},a_{\mathrm{yaw}}).
\]
All commands have Euclidean norm $s=0.5$. The cardinal commands are
\[
\uparrow=(s,0),\qquad
\downarrow=(-s,0),\qquad
\rightarrow=(0,s),\qquad
\leftarrow=(0,-s),
\]
and the diagonal commands are
\[
\begin{aligned}
\nearrow&=\frac{s}{\sqrt{2}}(1,1),&
\nwarrow&=\frac{s}{\sqrt{2}}(1,-1),\\
\searrow&=\frac{s}{\sqrt{2}}(-1,1),&
\swarrow&=\frac{s}{\sqrt{2}}(-1,-1).
\end{aligned}
\]
Thus, each active diagonal coordinate has magnitude
$s/\sqrt{2}\approx0.354$. Normalizing the diagonals in this way tests
translation--steering composition without increasing total command magnitude
relative to the cardinal directions.

\paragraph{Context and rollout duration.}
Our model receives one temporal context chunk containing three latent frames,
corresponding to $12$ RGB frames or $0.75$\,s at $16$\,fps. It then generates
eight autoregressive chunks, corresponding to $96$ RGB frames or $6.0$\,s.
The complete context-plus-generation clip therefore contains $108$ RGB frames
and spans $6.75$\,s.

\paragraph{Sampling and reproducibility.}
Our model uses the rectified flow-matching sampler with $48$ denoising steps
per generated chunk, \texttt{timestep\_shift}$=5.0$, and
\texttt{real\_guidance\_scale}$=0.0$, corresponding to no classifier-free
guidance. We evaluate the step-$5{,}000$ checkpoint
\texttt{causal\_lora\_step0005000.pt}. The same sampling configuration is used
for all of our ablations.

Sampling is deterministic. For generated block
$b\in\{0,\ldots,7\}$, the initial noise is drawn using
\[
\mathrm{seed}(b)=1234+b.
\]
The same fixed eight-block noise sequence is reused for every context and
command, holding sampling noise constant across the action-forcing
comparisons.

Each baseline uses its released evaluation checkpoint, sampler, inference-step
count, guidance configuration, and random-seed handling without modification.
The only extensions are the control-interface mappings described in
Section~\ref{app:baseline_controls}.

\paragraph{Evaluation resolution and motion readout.}
Each model is evaluated at its native output resolution; no shared resizing is
applied before the CoTracker--PCA controllability readout. The resolutions are
listed in Table~\ref{tab:action_forcing_resolutions}.

\begin{table}[t]
\centering
{
\small
\begin{tabular}{lc}
\toprule
Model                  & Output resolution \\
\midrule
Ours                   & $832\times480$ \\
minWM                  & $832\times480$ \\
WorldCam               & $832\times480$ \\
HY-World 1.5 WorldPlay & $832\times480$ \\
Astra                  & $832\times480$ \\
Matrix-Game 2.0        & $640\times352$ \\
Yume-1.5               & $1280\times704$ \\
\bottomrule
\end{tabular}
}
\caption{
    Native output resolutions used by the action-forcing benchmark.
}
\label{tab:action_forcing_resolutions}
\end{table}

Because the conditioning interfaces and native output
resolutions are not metrically matched, raw PCA response magnitudes are not
treated as directly comparable across models. Cross-model analysis instead
emphasizes response sign, requested direction, within-model
forward--backward asymmetry, and translation--steering composition. For the
separate cross-model image-translation analysis, only width-normalized
horizontal and vertical translations are compared.

\paragraph{Failed generations.}
Each context--command rollout is executed independently. If generation fails,
the rollout is skipped entirely: no video or metrics row is written, and the
missing result is not zero-filled or replaced. In practice, after the
control-interface implementations were finalized, all $256$ rollouts
completed successfully for every evaluated model. No benchmark rows were
therefore removed because of generation failure.

\subsection{Baseline Control Interfaces}
\label{app:baseline_controls}

\paragraph{Common protocol.}
Each compass command is translated into the closest operation supported by the
released model and repeated throughout generation. For pose-conditioned
models, the running camera pose is composed with a fixed relative translation
and/or yaw increment. For keyboard--mouse and text-conditioned models, the
corresponding control input or caption is held constant throughout the
rollout. For a given model, all parameters are fixed across scenes and
directions.

\paragraph{minWM}
minWM accepts an autoregressive sequence of relative $\mathrm{SE}(3)$ camera
poses. We use the released trajectory-generator increments of $0.08$
translation and $3^\circ$ yaw per generation step. Forward and backward use
positive and negative longitudinal translation, left and right use signed
yaw, and diagonal commands combine the corresponding translation and yaw
increments. The no-op command uses zero translation and zero yaw, leaving the relative
pose at identity for every generation step.

\paragraph{HY-World 1.5 WorldPlay}
WorldPlay is also driven using relative $\mathrm{SE}(3)$ camera poses and uses
the released trajectory scale of $0.08$ translation and $3^\circ$ yaw per
step. Its pose-string interface cannot express simultaneous translation and
yaw, so diagonal commands are supplied through the released JSON pose
interface. Backward and backward-diagonal commands use negative longitudinal
translation. The no-op command supplies an identity extrinsic for every frame through the
same JSON pose interface.

\paragraph{Astra}
Astra conditions on a per-frame relative camera pose represented by a
flattened $3\times4$ matrix. We follow the scale used in the released
demonstration code, including its translation increment of $0.03$. The release
provides forward and signed-steering primitives. We construct backward motion
by reversing the longitudinal translation and construct backward-diagonal
commands by combining this translation with the released signed-yaw
primitive. The no-op command holds the identity relative pose $\mathbf{I}_{3\times4}$ for
every frame. This matches the release's own convention: its trajectory
generator already assigns identity poses to the conditioning frames, annotated
as zero-motion camera pose, and our no-op extends that convention across the
generated span rather than switching to a translation increment at the
hand-off.

\paragraph{WorldCam}
WorldCam is conditioned on per-pixel-frame camera-to-world trajectories in
OpenCV convention. Its pose scale is inherited from its training trajectories
and is not metric. Because the release does not include an eight-direction
trajectory generator, we construct one through the released pose interface.

The conditioning span uses identity poses. During generation, the running pose
is composed multiplicatively as
\[
C_{t+1}=C_tD,
\]
where $D$ contains a longitudinal translation of
\[
\Delta z=\pm0.02
\]
camera-to-world units per pixel frame and/or a yaw rotation of
$\pm1.5^\circ$ about the $y$-axis. Positive and negative translations are used
for the forward and backward command families, respectively. Positive yaw is
used for right-family commands and negative yaw for left-family commands.
Diagonal commands combine the corresponding translation and yaw within the
same increment.

The released runner performs $50$ autoregressive steps and produces $200$
generated frames. We use the fixed intrinsics
\[
f_x=f_y=723,\qquad c_x=960,\qquad c_y=540.
\]
The translation and yaw increments were fixed
during interface calibration before the action-forcing benchmark and were
then applied unchanged across every scene and direction. The no-op command sets $D=I$, that is $\Delta z=0$ with zero yaw, so the
running pose $C_{t+1}=C_tD$ remains at the identity of the conditioning span
throughout generation.

\paragraph{Matrix-Game 2.0}
Matrix-Game accepts a six-dimensional binary keyboard stream and a
two-dimensional mouse stream, both held constant throughout generation. We
use its released benchmark mouse-yaw magnitude of $0.1$ per pixel frame. The
eight command mappings are given in
Table~\ref{tab:matrix_game_commands}.

\begin{table}[t]
\centering
{
\small
\begin{tabular}{ccc}
\toprule
Command        & Keyboard input      & Mouse yaw \\
\midrule
$\uparrow$     & \texttt{forward}=1  & $0$ \\
$\downarrow$   & \texttt{backward}=1 & $0$ \\
$\rightarrow$  & none                & $+0.1$ \\
$\leftarrow$   & none                & $-0.1$ \\
$\nearrow$     & \texttt{forward}=1  & $+0.1$ \\
$\nwarrow$     & \texttt{forward}=1  & $-0.1$ \\
$\searrow$     & \texttt{backward}=1 & $+0.1$ \\
$\swarrow$     & \texttt{backward}=1 & $-0.1$ \\
$\varnothing$  & none                & $0$ \\
\bottomrule
\end{tabular}
}
\caption{
    Matrix-Game 2.0 controls used for the eight-direction benchmark.
}
\label{tab:matrix_game_commands}
\end{table}

Keyboard controls are binary, whereas mouse yaw has a continuous magnitude.
Backward input is supported by the released interface but is not exercised in
the model's released benchmark protocol.

\paragraph{Yume-1.5.}
Yume is controlled through natural-language captions drawn from its supported
training vocabulary. Each command uses the common prefix
\begin{quote}
\small
\emph{This video depicts a city walk scene with a first-person view (FPV).}
\end{quote}
followed by the corresponding person and camera clauses in
Table~\ref{tab:yume_commands}. The complete caption is repeated throughout
the rollout. Yume exposes directional commands but no numerical command
magnitude and therefore cannot be evaluated using a continuous dose sweep.

\begin{table}[t]
\centering
{
\small
\begin{tabular}{lll}
\toprule
Command        & Person clause                  & Camera clause \\
\midrule
$\uparrow$     & moves forward (W)              & remains still \\
$\downarrow$   & moves backward (S)             & remains still \\
$\leftarrow$   & stands still                   & turns left \\
$\rightarrow$  & stands still                   & turns right \\
$\nwarrow$     & moves forward and left (W+A)   & turns left \\
$\nearrow$     & moves forward and right (W+D)  & turns right \\
$\swarrow$     & moves backward and left (S+A)  & turns left \\
$\searrow$     & moves backward and right (S+D) & turns right \\
$\varnothing$  & stands still ($\cdot$)         & remains still ($\cdot$) \\
\bottomrule
\end{tabular}
}
\caption{
    Yume-1.5 caption clauses used for the eight-direction benchmark.
}
\label{tab:yume_commands}
\end{table}

\paragraph{Text-conditioning neutrality.}
Several baselines accept a text prompt alongside their action interface. Under
a directional command, motion-referring language in that prompt is harmless;
under a no-op it contradicts the commanded behaviour. Three of the prompts we
used for the directional benchmark contain such language: Astra's scene prompt
ends with ``\emph{The camera moves smoothly through the scene}''; Yume's
common prefix describes a ``\emph{city walk}'' scene; and WorldCam inherits
its base model's default negative prompt, which lists three terms meaning
\emph{static}, \emph{motionless}, and \emph{a still, unmoving picture} among
the artefacts to be avoided, thereby penalising the behaviour under test.

For the stationary evaluation we therefore repeat these rollouts with the
motion-referring language removed and all other settings unchanged.
Neutralisation reduces camera drift substantially for Yume and Astra, leaves
minWM and WorldPlay essentially unchanged, and increases drift for WorldCam.
We use the neutral prompts for Yume and Astra and retain the released defaults
elsewhere, following the rule ``use the released default unless it explicitly
contradicts the command''. The stationary results in
Appendix~\ref{app:stationary_results} use these settings.

\paragraph{Authored interface extensions.}
Our additions operate only through the models' released conditioning
interfaces. We construct backward and backward-diagonal pose sequences for
Astra, backward--yaw combinations for Matrix-Game diagonals, an
eight-direction trajectory generator for WorldCam, and simultaneous
translation--yaw commands through WorldPlay's JSON interface. For the
stationary evaluation we further supply each model with a no-op command
through its native interface: identity pose sequences for the four
camera-conditioned models (minWM, WorldPlay, Astra, WorldCam), an all-zero
keyboard and mouse-control stream for Matrix-Game, and the composed
still-clause caption for Yume. In each case the no-op is the zero element of
the model's own conditioning representation, and for the pose-conditioned
models it coincides with the identity convention those releases already apply
to their conditioning spans.

These inputs are representable by the released conditioning tensors but are
not part of the models' published benchmark protocols, which expose no named
or evaluated no-op command; we therefore report the resulting rollouts as each
model's response to an authored zero-input extension. No model architecture,
learned weight, or conditioning representation is modified.

PCA centres the tracked-motion vectors before projection, so the mean raw PCA
score is zero by construction. However, the numerical score origin corresponds
to the training-set mean tracked-motion field rather than to zero displacement.
We therefore obtain our stationary command by passing an all-zero displacement
field through the same action encoder, giving the calibrated value
$a_{\mathrm{null}}=(-0.0234,-0.0013)$ defined in
Eq.~\eqref{eq:null_action}. This correction cancels the mildly forward/right
mean motion and is used only for the no-op evaluation; all training actions and
all active commands retain their original encoded values. No analogous
calibration is applied to the baselines, whose stationary inputs are the
untuned zero or identity elements of their released interfaces.

\subsection{Response-Curve Protocol}
\label{app:response_details}

\paragraph{Ablation curves.}
The reverse-throttle and steering branches are measured using held-out ride
windows whose recorded commands are replaced by their rotated
counterfactuals. We use evaluations from the final $1{,}000$ training steps,
restrict the window offset to at most $81$, and measure only settled chunks by
discarding the first two generated chunks of each rollout. This is because there is a finite acceleration in the data and rapid direction changes induce a penalty from this acceleration. We find this in a lot of the other baseline models too, so all models use this protocol.

For each action axis, requested values are divided into $13$ linearly spaced
bins spanning the $2$nd to the $98$th percentile of the requested-command
distribution. A bin is plotted only when it contains at least $12$ samples.
Each plotted point is the mean realized CoTracker--PCA response within that
bin. Shaded regions show one standard error of the mean:
\[
\mathrm{SEM}
=
\frac{\operatorname{std}(x)}{\sqrt{n}}.
\]

Genuine reverse travel is too rare to produce a sufficiently populated
rotated forward branch. We therefore measure forward response using
held-out no-op contexts and the fixed positive throttle strengths
\[
\{0.1,0.2,\ldots,0.8\}.
\]
Settled response is again measured after discarding the first two generated
chunks. When more than $132$ samples are available at a command strength, we
subsample $132$ examples using a fixed random seed of $0$. This cap matches
the size of the largest FLIP bin.

The forward and reverse branches therefore probe the same throttle coordinate
but do not use identical starting-context distributions. They are used to
evaluate response sign, smoothness, saturation, and approximate gain rather
than as a paired forward--backward comparison. Dashed plot boundaries mark
the approximate support of the training actions, extending to approximately
$+0.5$ for forward throttle and $-0.3$ for reverse throttle.

\paragraph{Cross-model curves.}
Cross-model response curves use the same $32$ held-out contexts as the
eight-direction action-forcing benchmark. The compass commands produce the
five signed per-axis protocol levels
\[
\left\{
-0.5,\,
-\frac{0.5}{\sqrt{2}},\,
0,\,
\frac{0.5}{\sqrt{2}},\,
0.5
\right\}
\approx
\{-0.5,-0.354,0,0.354,0.5\}.
\]
For each level, the plotted point is the mean response over the $32$ scenes
and the shaded region shows $\pm1$ SEM.

Each baseline receives the corresponding fixed-strength command through its
native interface. The horizontal coordinate represents the signed protocol
level rather than a common physical control magnitude. The curves are
therefore interpreted through response sign and within-model
forward--backward asymmetry rather than absolute cross-model gain. The dotted
diagonal denotes unit gain.

\section{Extended Controllability Results}
\label{app:extended_control}

\subsection{Conditioning-Pathway Ablation}
\label{app:conditioning_ablation}

\begin{figure*}[t]
    \centering
    \includegraphics[
        width=0.82\textwidth
    ]{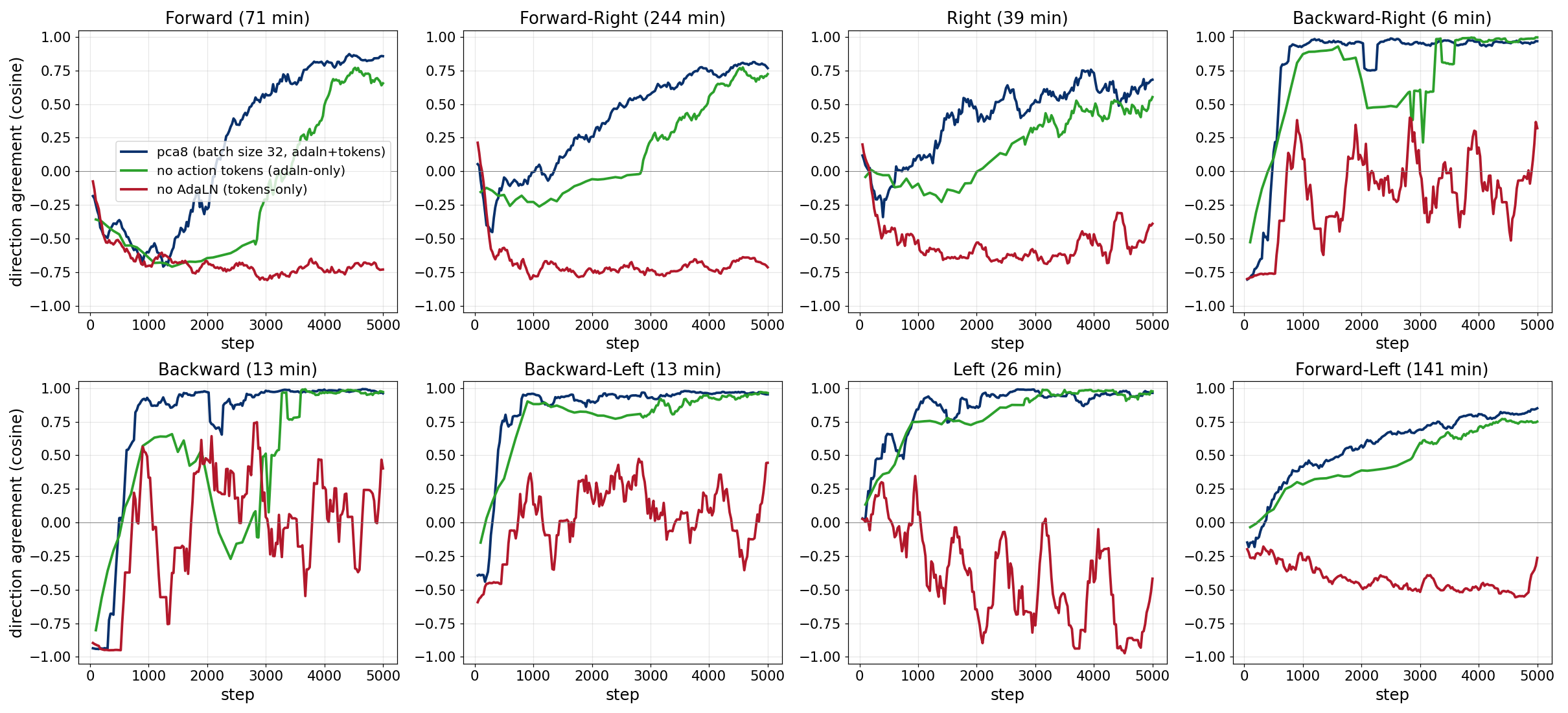}
    \caption{
        Conditioning-pathway ablation under counterfactual FLIP evaluation,
        grouped by source-ride motion direction. Dual injection through AdaLN
        and action tokens converges fastest and reaches the strongest response.
        AdaLN-only conditioning learns control more slowly, while tokens-only
        conditioning remains anti-aligned on throttle and unstable on steering.
        Panels and smoothing follow Fig.~\ref{fig:following_dir}.
    }
    \label{fig:abl_injection}
\end{figure*}

Figure~\ref{fig:abl_injection} isolates the effect of the action-conditioning
pathway. Injecting actions through both AdaLN and action tokens produces the
fastest convergence and strongest counterfactual response. AdaLN alone still
establishes control, but converges more slowly and reaches a weaker response.
Action tokens alone appear competitive under GT evaluation but fail under
FLIP, showing that continuation of the source motion can be mistaken for
command following.

\subsection{Training-Time Command Following}
\label{app:extended_control_following}

\begin{figure*}[t]
    \centering
    \includegraphics[
        width=0.82\textwidth
    ]{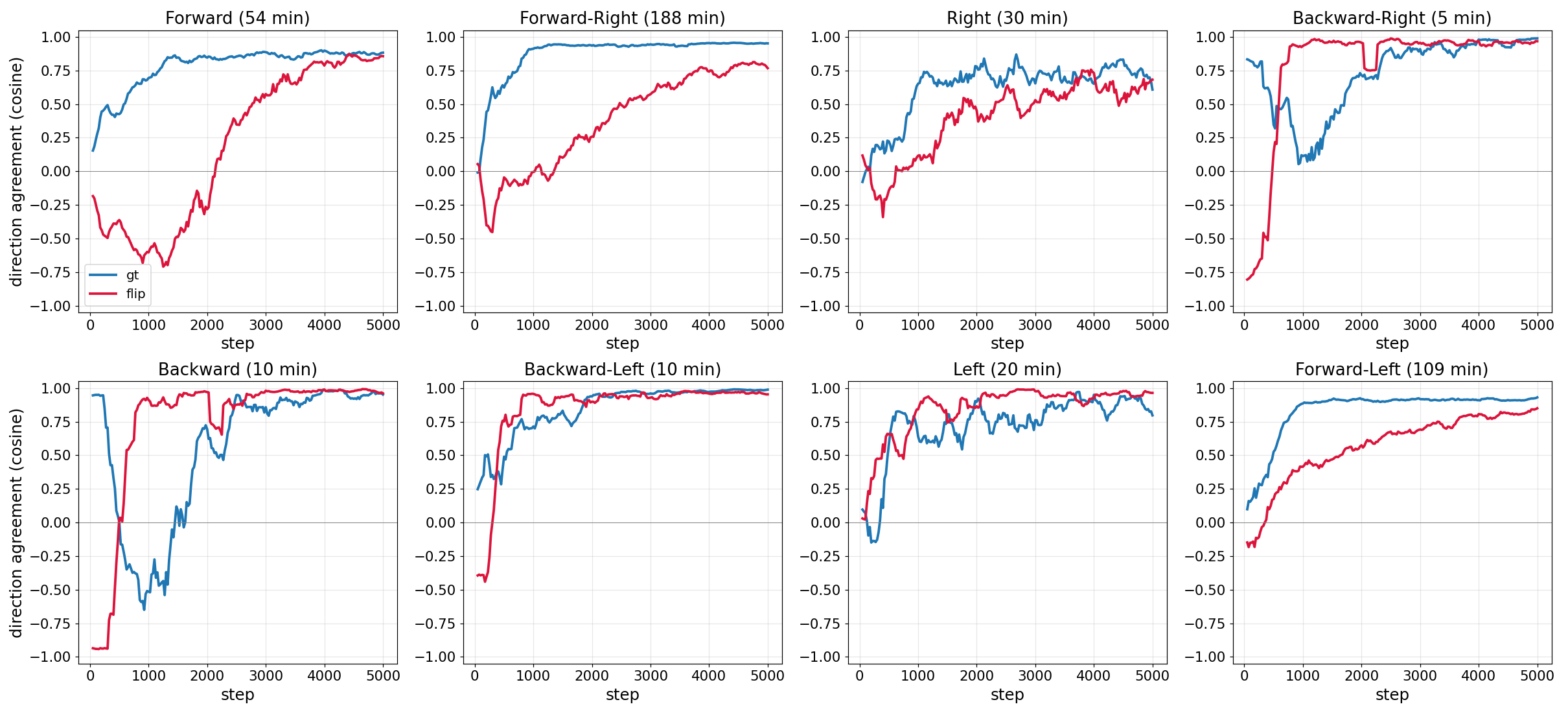}
    \caption{
        Training-time command following for the default model across all eight
        source-motion groups. Blue uses the recorded command (GT), while red applies
        the sign-flipped counterfactual (FLIP) to the same context. Positive FLIP
        agreement indicates genuine command response. Panel titles denote the
        source-context motion, and curves use a $9$-point
        ($\approx450$-step) centred rolling mean.
    }
    \label{fig:following_8pca}
\end{figure*}

\paragraph{Default model.}
Figure~\ref{fig:following_8pca} extends the main-paper training analysis to all
eight source-motion groups for the default pca8 model with global batch size
$32$.

\begin{figure*}[t]
    \centering
    \includegraphics[
        width=0.82\textwidth
    ]{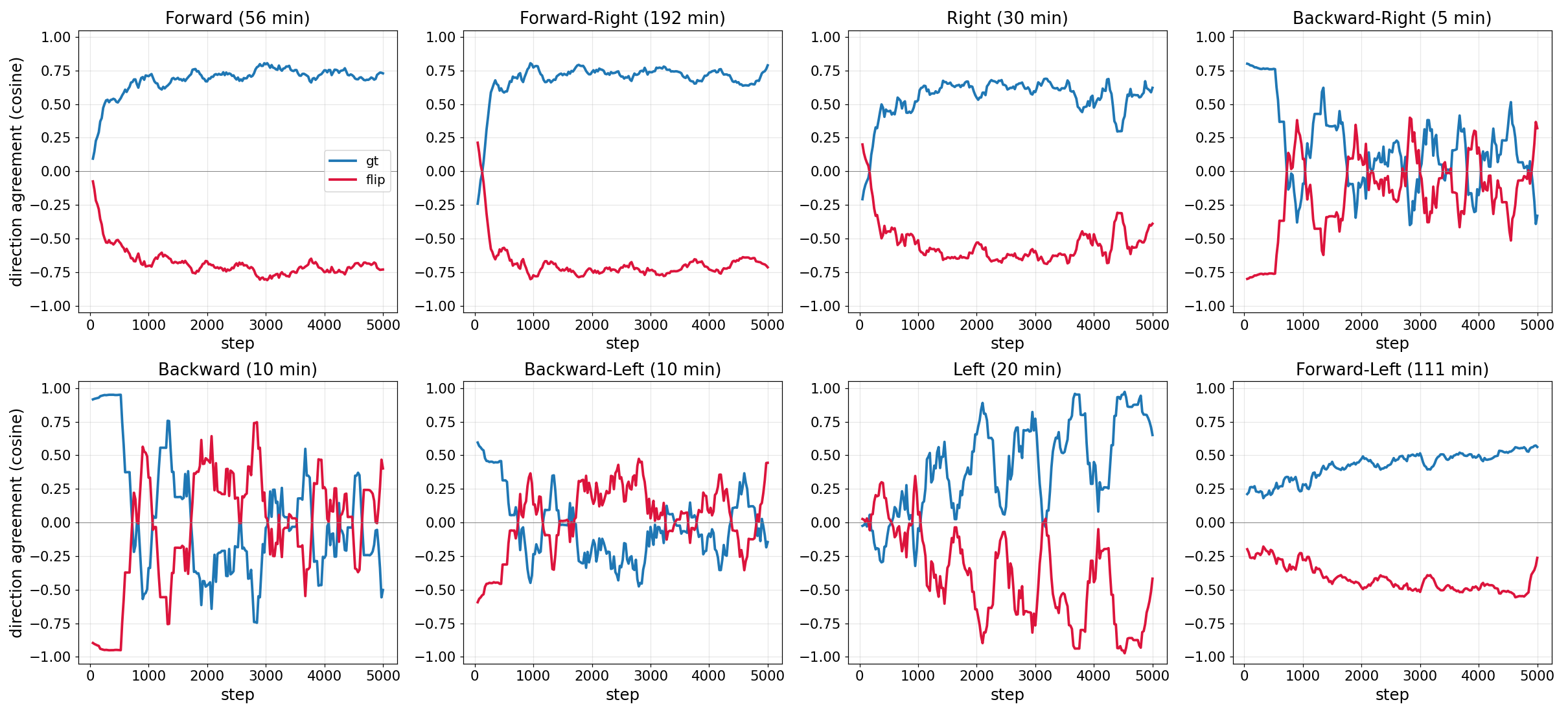}
    \caption{
        GT and FLIP agreement for the no-AdaLN, tokens-only variant. High GT
        agreement but negative or unstructured FLIP response shows that the model
        continues source motion without acquiring reliable command control. Panel
        titles denote source-context motion.
    }
    \label{fig:noadaln_gt_flip}
\end{figure*}

\paragraph{Tokens-only conditioning.}
Figure~\ref{fig:noadaln_gt_flip} illustrates why GT agreement is insufficient
on its own. The no-AdaLN, tokens-only variant obtains high GT agreement in
several common source-motion contexts, but its FLIP response remains negative
or unstructured. It therefore continues motion inherited from the context
without acquiring reliable command control.

\begin{figure*}[t]
    \centering
    \includegraphics[
        width=0.82\textwidth
    ]{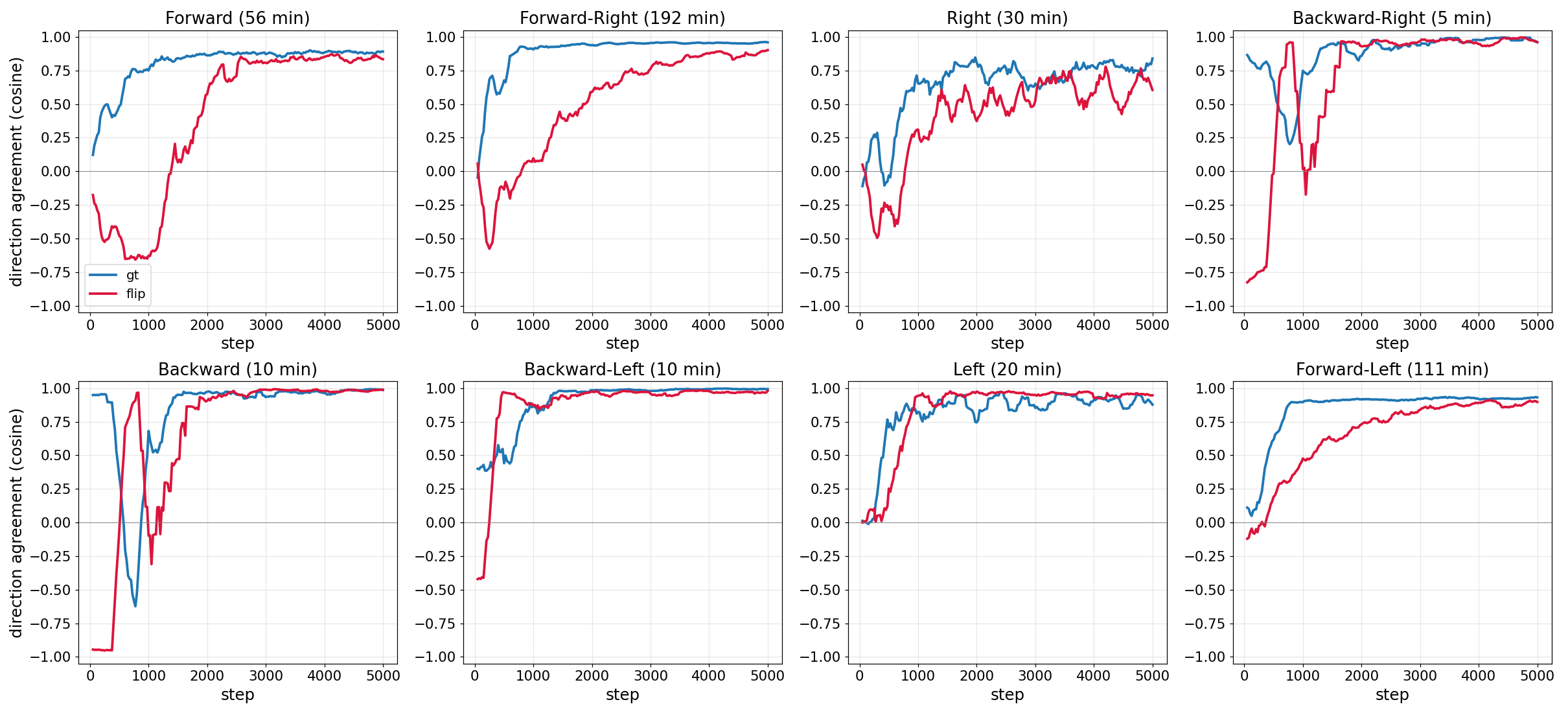}
    \caption{
        GT and FLIP agreement for the Batch64 model. FLIP becomes positive across
        all eight directions, demonstrating genuine command response. Panel titles
        denote source-context motion.
    }
    \label{fig:following_16pca}
\end{figure*}

\paragraph{Larger-batch model.}
Figure~\ref{fig:following_16pca} shows the corresponding training curves for
the Batch64 model. GT agreement again rises rapidly, but FLIP agreement also
becomes consistently positive across all eight directions. Relative to the
default batch size, the larger batch produces smoother training dynamics and
a stronger response in the harder throttle directions.

\subsection{Batch-Size and Critic-Supervision Ablations}
\label{app:control_training_ablations}

\begin{figure*}[t]
    \centering
    \includegraphics[
        width=0.82\textwidth
    ]{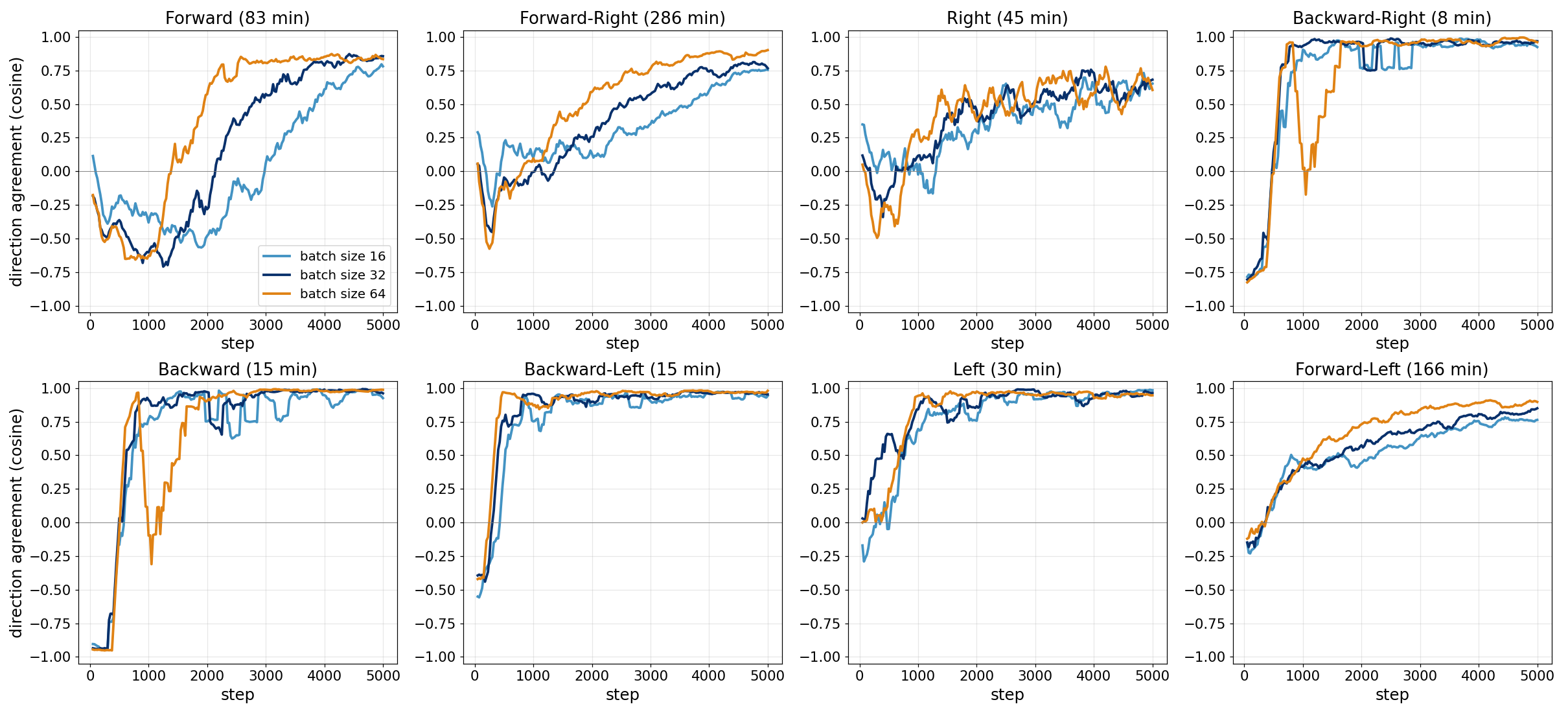}
    \caption{
        Batch-size ablation using global batch sizes $16$, $32$, and $64$
        under FLIP evaluation. All batch sizes recover the same directional
        structure by step $5{,}000$, but larger batches acquire the hardest
        throttle responses earlier and retain stronger reverse motion.
    }
    \label{fig:abl_nodes}
\end{figure*}

\begin{figure*}[t]
    \centering
    \includegraphics[
        width=0.82\textwidth
    ]{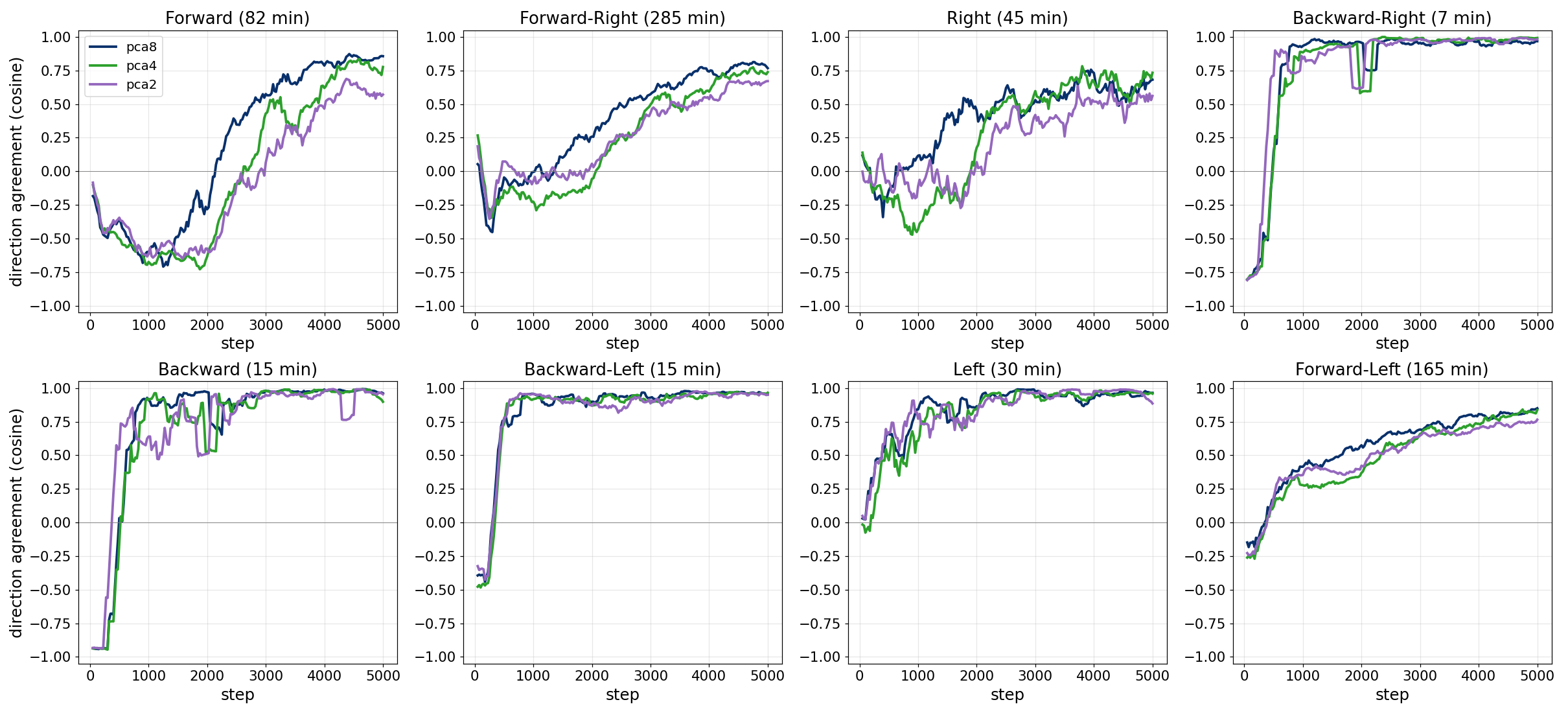}
    \caption{
        Critic-supervision ablation using the leading $2$, $4$, or $8$ PCA
        components. Two supervised components suffice to recover directional
        control, while richer supervision improves convergence and realized
        response strength.
    }
    \label{fig:abl_encoders}
\end{figure*}

Reducing either global batch size
(Fig.~\ref{fig:abl_nodes}) or the number of PCA components supervised by the
critic (Fig.~\ref{fig:abl_encoders}) delays convergence and weakens the
realized reverse response, but does not prevent recovery of the backward
direction.

\subsection{Settled Ablation Responses}
\label{app:settled_ablation_responses}

\begin{figure*}[t]
    \centering
    \includegraphics[
        width=0.82\textwidth
    ]{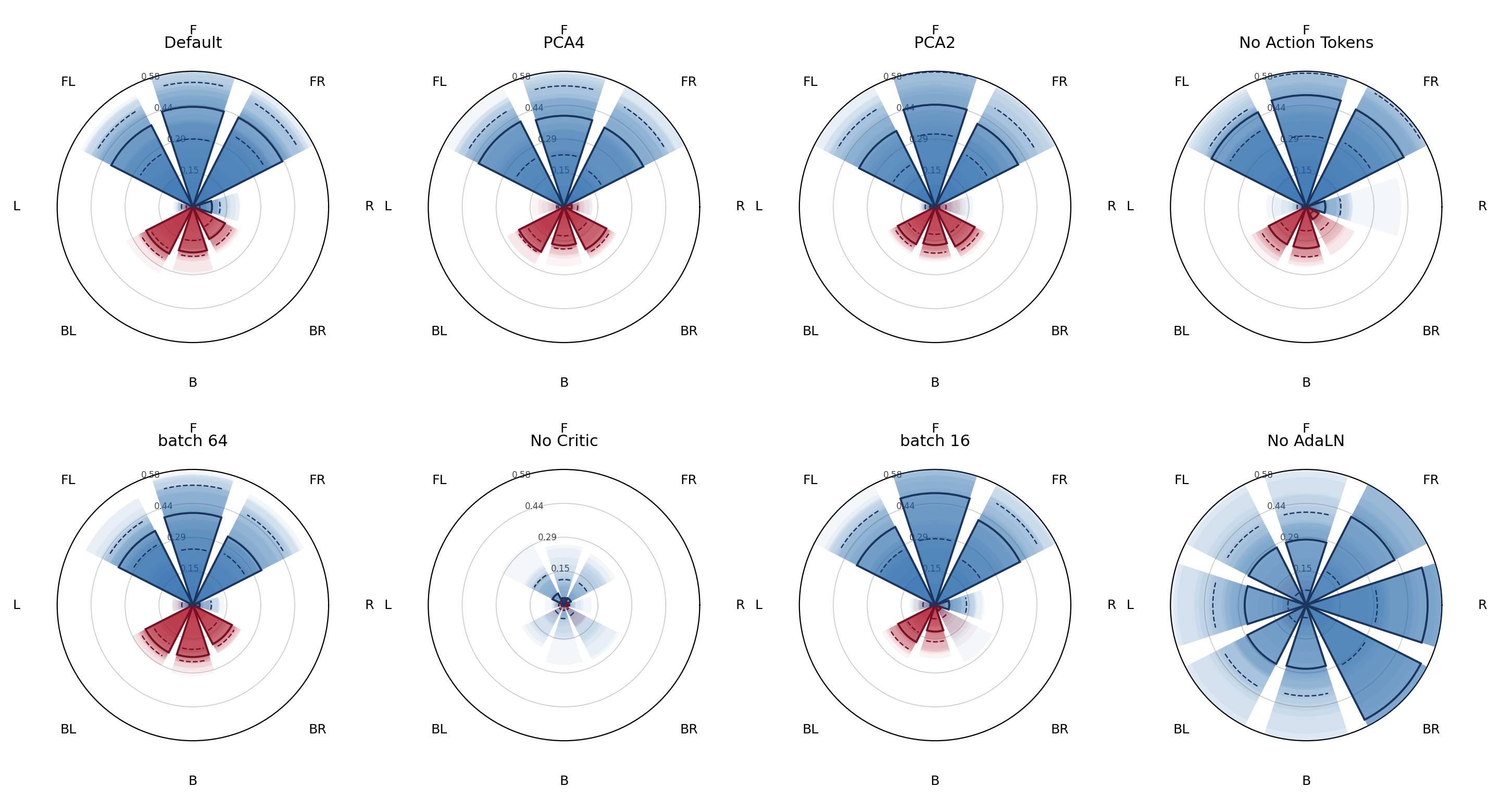}
    \caption{
        Settled throttle response $g_0$ by commanded direction for the ablation
        variants. Each translucent wedge represents one of $256$ rollouts; blue
        denotes realised forward motion and red realised backward motion. Solid
        outlines show medians, and panel rims show the local $95$th percentile. The
        tokens-only model renders every command as forward motion.
    }
    \label{fig:wedge_g0}
\end{figure*}

\begin{figure*}[t]
    \centering
    \includegraphics[
        width=0.82\textwidth
    ]{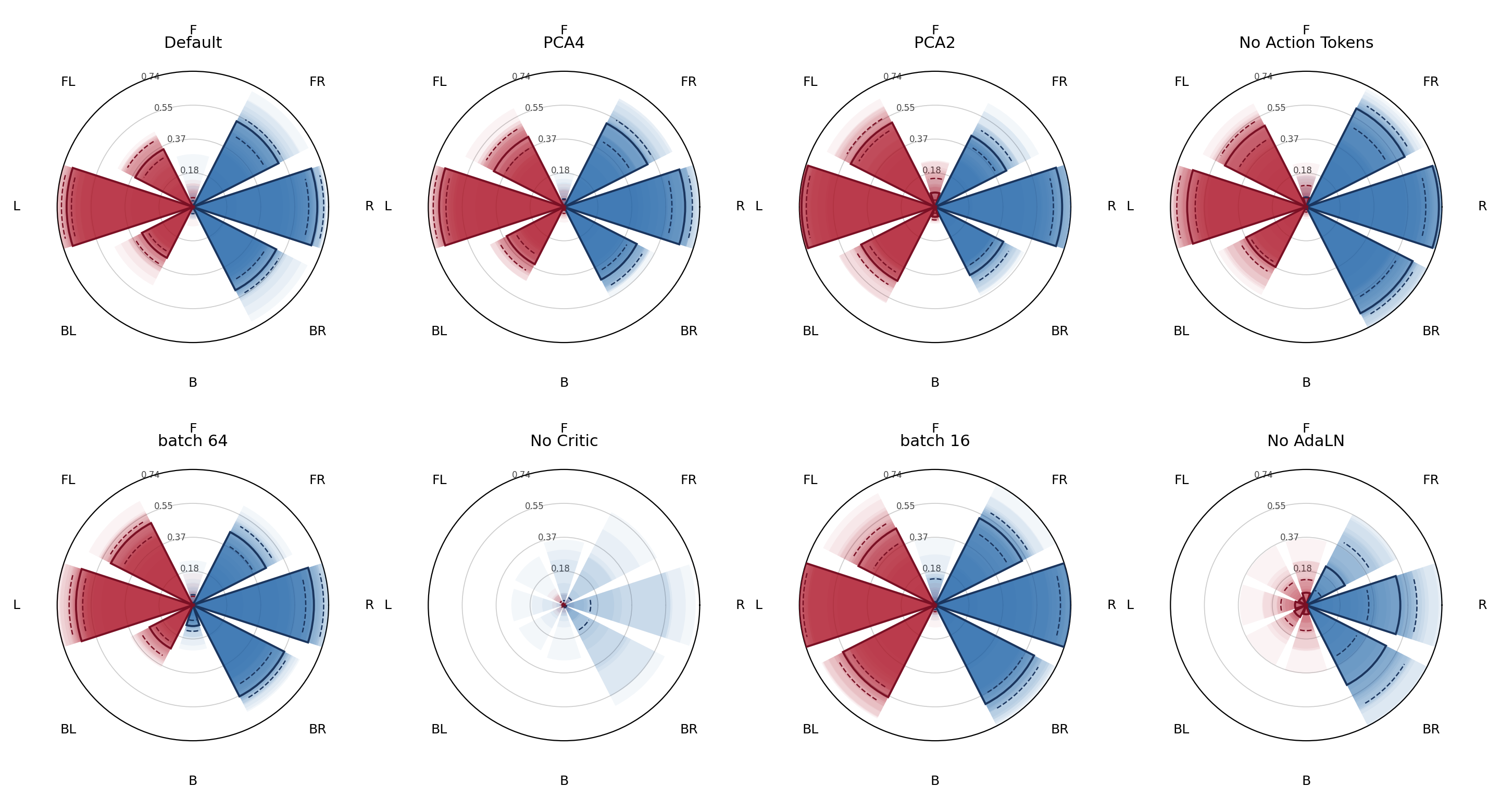}
    \caption{
        Settled steering response $g_1$ by commanded direction for the ablation
        variants. Blue denotes realised rightward yaw and red realised leftward yaw;
        opacity, outlines, and panel normalization otherwise follow
        Fig.~\ref{fig:wedge_g0}. Every functional variant shows the expected
        four-lobe structure.
    }
    \label{fig:wedge_g1}
\end{figure*}

Figures~\ref{fig:wedge_g0} and~\ref{fig:wedge_g1} compare the settled
directional responses of the ablation variants. Batch size has its largest
effect on reverse throttle: Batch16 is substantially weaker under backward and
backward-diagonal commands than Batch32 and Batch64. Steering is less
sensitive to batch size and critic-supervision dimensionality, with every
functional variant retaining the expected signed four-lobe structure. 

\subsection{Cross-Model Response Curves}
\label{app:cross_model_response}

\begin{figure}[t]
    \centering

    \begin{subfigure}[t]{\columnwidth}
        \centering
        \includegraphics[
            width=0.82\linewidth
        ]{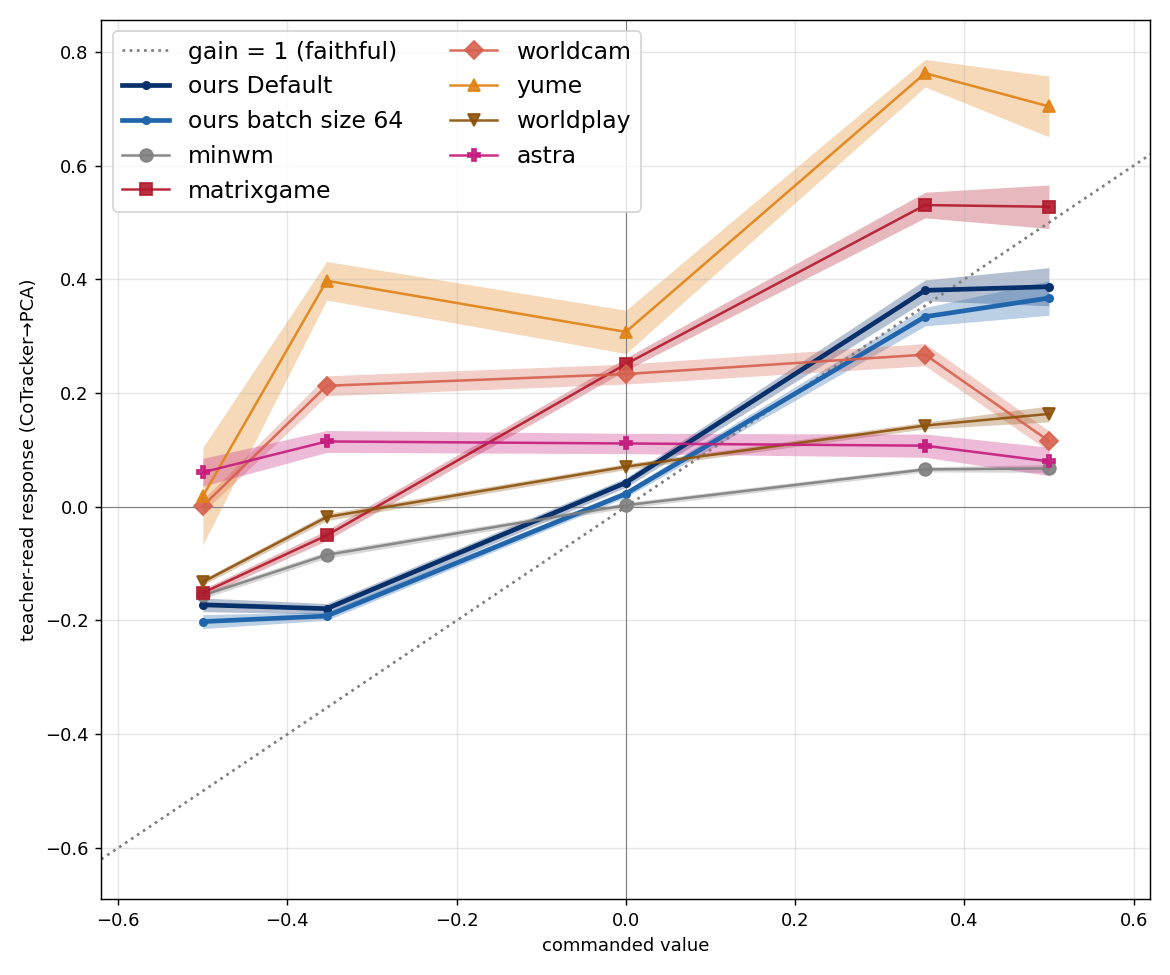}
        \caption{Throttle response.}
        \label{fig:response_eval_forward}
    \end{subfigure}

    \begin{subfigure}[t]{\columnwidth}
        \centering
        \includegraphics[
            width=0.82\linewidth
        ]{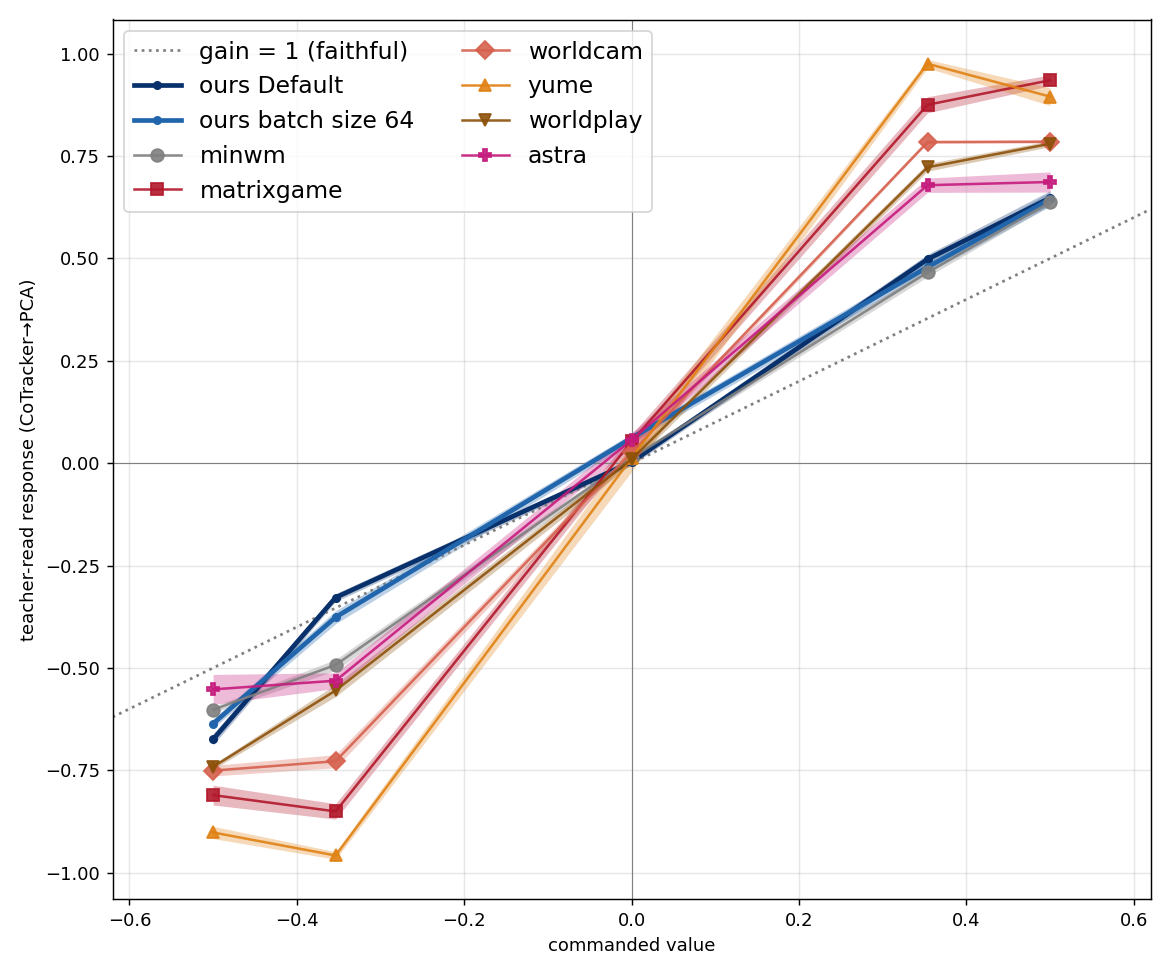}
        \caption{Steering response.}
        \label{fig:response_eval_steer}
    \end{subfigure}

    \caption{
        Cross-model action-forcing response curves from $32$ low-egomotion
        held-out contexts. Realized motion is measured using the common
        CoTracker--PCA readout, and shaded bands show $\pm1$ SEM. Baselines are
        shown only at strengths supported by their released interfaces.
        The dotted diagonal denotes unit gain as a visual reference.
    }
    \label{fig:response_gain_all}
\end{figure}

Figure~\ref{fig:response_gain_all} provides a scalar view of the behavior
summarized by the main-paper directional wedges. The curve construction and
uncertainty estimates are defined in
Section~\ref{app:response_details}.

Our model supports continuous-valued actions, whereas several baselines expose
only fixed pose increments, binary controls, or text commands without a
numerical strength parameter; see
Section~\ref{app:baseline_controls},
Table~\ref{tab:matrix_game_commands}, and
Table~\ref{tab:yume_commands}. Baselines are therefore shown only at strengths
supported by their released interfaces, and unsupported intermediate values
are not interpolated.

\subsection{Cross-Model Steering and Yaw--Roll Coupling}
\label{app:cross_model_steering_roll}

\begin{figure*}[t]
    \centering
    \includegraphics[
        width=0.82\textwidth
    ]{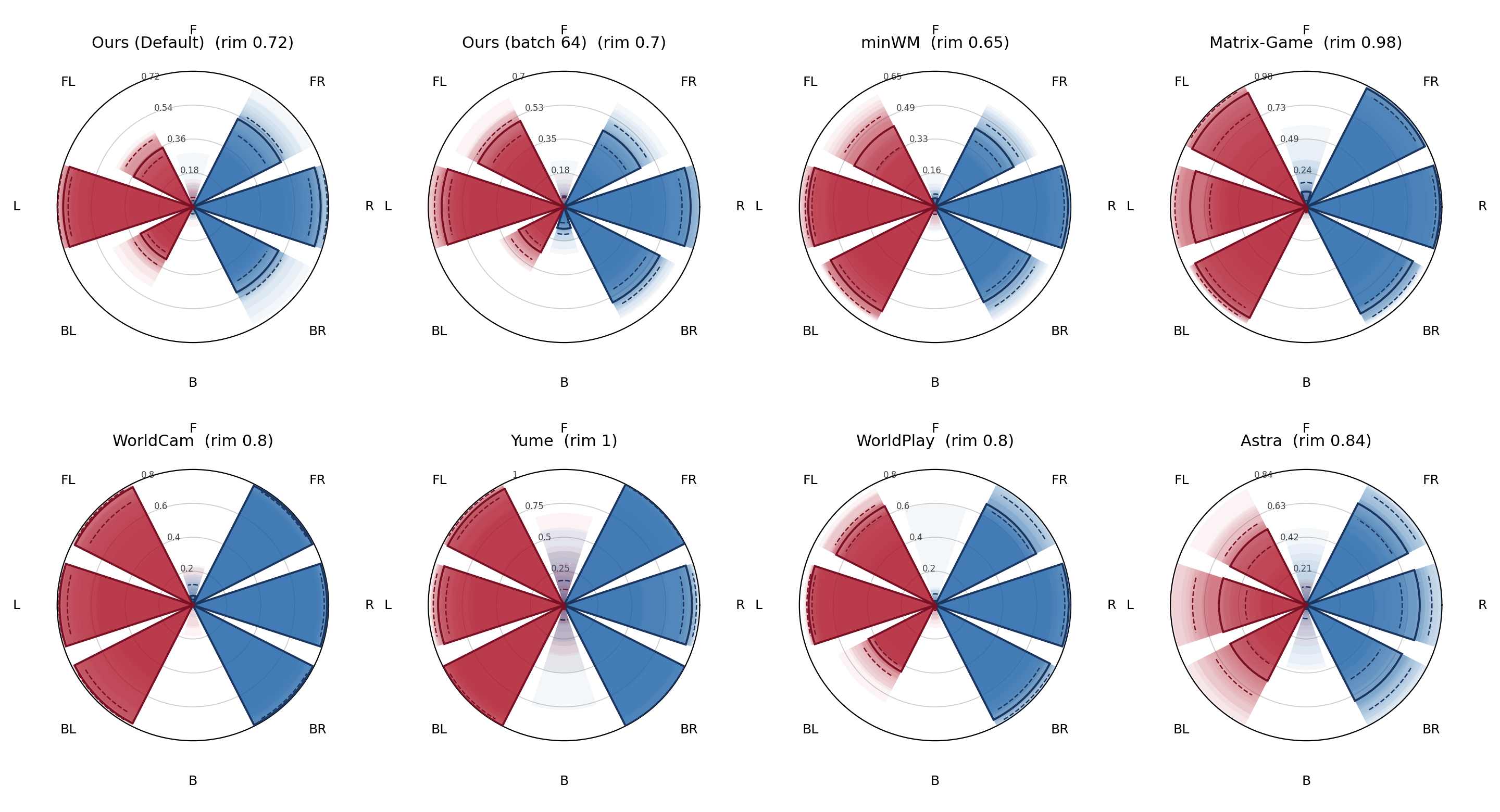}
    \caption{
        Steering response $g_1$ by commanded direction for all models. Blue denotes
        realised rightward yaw and red realised leftward yaw; other wedge
        conventions follow the main-paper throttle figure. Every model recovers the
        expected steering sign, although response magnitude and variance differ.
    }
    \label{fig:wedge_all_g1}
\end{figure*}

\begin{figure*}[t]
    \centering
    \includegraphics[
        width=0.82\textwidth
    ]{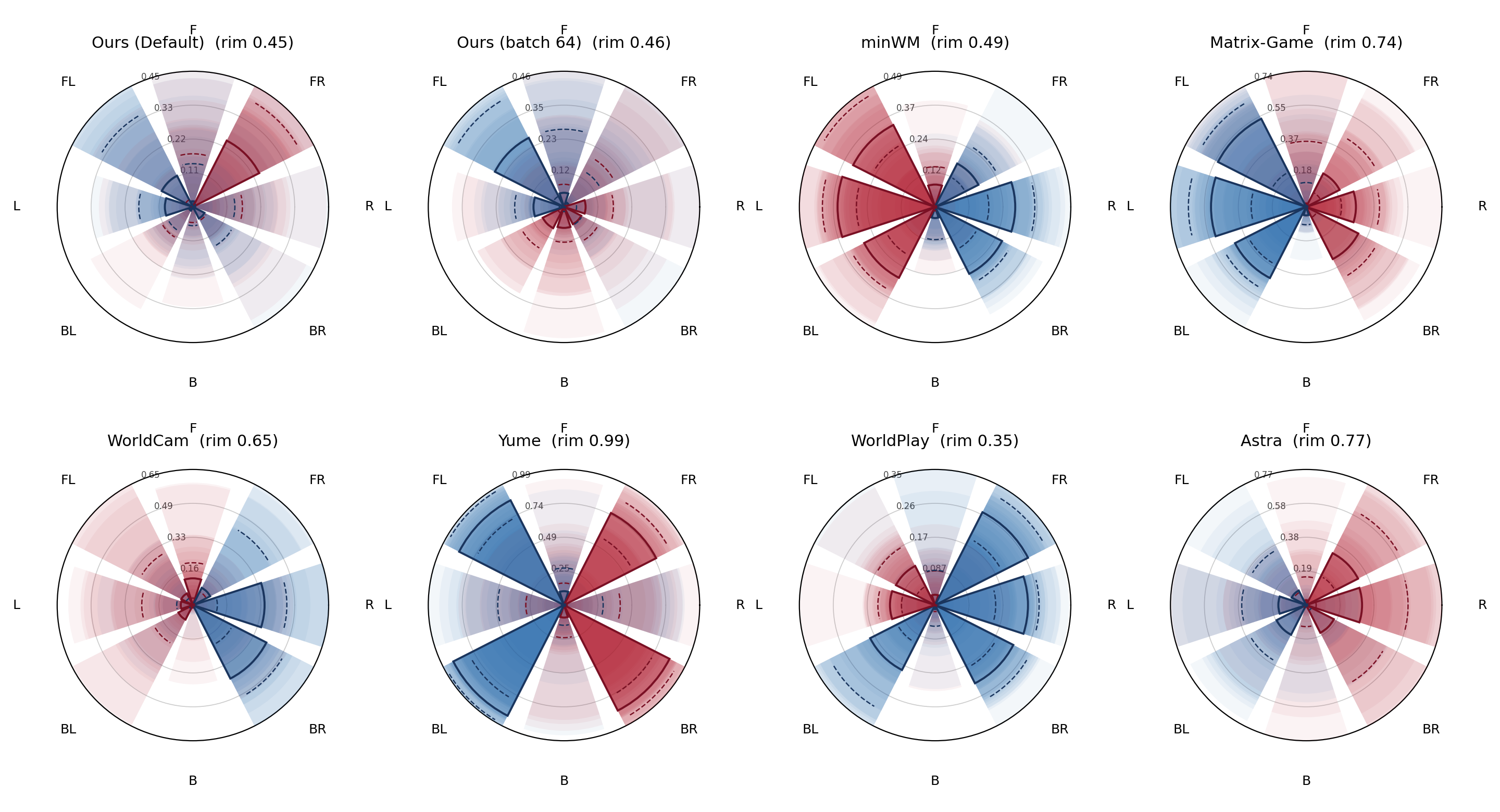}
    \caption{
        Uncommanded roll response $g_6$ under each planar command. Radius indicates
        roll magnitude; blue denotes clockwise and red counter-clockwise motion.
        Direction-dependent responses reveal coupling between commanded planar
        motion and out-of-plane camera motion.
    }
    \label{fig:wedge_all_g6}
\end{figure*}

Figure~\ref{fig:wedge_all_g1} shows that every evaluated model recovers the
requested steering sign, although response magnitude and consistency across
scenes differ. Figure~\ref{fig:wedge_all_g6} measures the roll-like component
$g_6$, which is not directly commanded but responds systematically to yaw
across several models.

A non-zero $g_6$ response is not necessarily a failure. Real egocentric turns
are rarely pure yaw rotations: vehicle dynamics, suspension, camera mounting,
and body motion can introduce a correlated roll component. Because the PCA
basis is recovered from real video, $g_6$ may capture this naturally coupled
motion. We therefore use it as a diagnostic of yaw--roll coupling rather than
assuming that all roll should vanish. Large or inconsistent responses remain
undesirable, particularly when they exceed the coupling observed in real
motion.

\section{Reference-Free Quality Evaluation Methodology}
\label{app:quality_metrics}

Action-forced rollouts do not admit a unique ground-truth continuation, making
paired reference metrics unsuitable. We instead evaluate complementary
failure modes that can be measured without a target video: visual style shift, geometric corruption, scene relocation, conjuration (spurious object appearance), and progressive loss of high-frequency detail.

\paragraph{Family-balanced reference.}
Only HF and relocation use other systems' outputs. Under \texttt{HF-Cal-v2}
and \texttt{Reloc-Cal-v2}, the reference contains the six external baselines
and one representative of our family. Default occupies that seat when an
external is scored; each ablation replaces Default when scored. Relocation
excludes the candidate endpoint and also uses real context references, so
each candidate is matched against six other-family systems. HF uses all
seven entries, including the candidate. No variant receives same-family
peer support; Default has the same score in the cross-model and ablation
tables. The reference values and scoring rule are fixed, so adding an
ablation does not change existing scores. These are peer-referenced
diagnostics, not ground-truth comparisons.

\paragraph{Released reference.}
The files \texttt{hf\_reference\_dblur\_v2\_directional.csv} ($256$ inputs)
and \texttt{hf\_reference\_dblur\_v2\_stationary.csv} ($32$ contexts) contain
per-input $d_{\mathrm{blur}}$ values for the six externals and Default.
Forming the median under the rule above reproduces the stored HF scores
exactly (maximum absolute error $0.0000$, no threshold-flag differences).
Relocation instead requires the ORB keypoints and descriptors in
\texttt{reloc\_reference\_pack\_3000.npz}, which also stores descriptors for four real context
frames. \texttt{Reloc-Cal-v2.json} identifies the eligible members and
verification settings. These artifacts, rather than model weights, define
the reference.

\subsection{Evaluation Population}
\label{app:quality_population}

The complete action-forcing benchmark contains $32\times8=256$ directional
rollouts per model. Two source contexts are unsuitable for feature-dependent
evaluation because water droplets and lens smearing leave their reference
frames nearly featureless. ORB detects only $371$ and $104$ keypoints in these
contexts, compared with approximately $2{,}900$--$3{,}000$ in representative
clean contexts. Reliable ORB--RANSAC correspondence counts are therefore
undefined for rollouts seeded from either context.

We exclude these two contexts uniformly from feature-dependent analyses. This
removes $2\times8=16$ rollouts from every model, leaving a common
$240$-rollout feature-valid population. The exclusion is a property of the
shared source footage rather than a model failure, and exactly the same
contexts and commands are removed for every model.

A model may avoid visible generation failures by producing little or no
motion. Within the feature-valid population, we therefore identify
near-static rollouts using the ORB matching and homography-verification
pipeline defined in Section~\ref{app:scene_relocation}. A rollout is classified
as static when its first generated frame and six-second-horizon frame retain
more than $600$ RANSAC-verified feature matches.

Style shift, scene relocation, and the deployed high-frequency-degradation
rate are reported on the model-specific active population, defined as
feature-valid and non-static. Geometry and conjuration do not require a
feature-rich reference and are therefore reported over all $256$ directional
rollouts. The resulting feature-valid and active populations are reported in
Table~\ref{tab:quality_denominators}.

\begin{table}[t]
\centering
{
\small
\begin{tabular}{lrr}
\toprule
Model or variant        & Feature-valid & Active \\
\midrule
WorldPlay               & 240 & 238 \\
Matrix-Game             & 240 & 240 \\
WorldCam                & 240 & 216 \\
Astra                   & 240 & 215 \\
Yume                    & 240 & 237 \\
minWM                   & 240 & 187 \\
\midrule
Ours (Default)          & 240 & 239 \\
Ours (pca4)             & 240 & 237 \\
Ours (pca2)             & 240 & 232 \\
Ours (Batch 64)         & 240 & 237 \\
Ours (Batch 16)         & 240 & 236 \\
Ours (No Action Tokens) & 240 & 229 \\
Ours (No AdaLN)         & 240 & 200 \\
Ours (No Critic)        & 240 & 75 \\
\bottomrule
\end{tabular}
}
\caption{
    Quality-evaluation populations. The complete directional fleet contains
    $256$ rollouts per model. Two wet-lens source contexts, corresponding to
    $16$ rollouts per model, are excluded uniformly because reference-based
    feature matching is undefined, leaving $240$ feature-valid rollouts.
    Active further excludes near-static continuations using the ORB-based
    criterion. Style, relocation, and deployed HF rates use the Active
    denominator.
}
\label{tab:quality_denominators}
\end{table}

\subsection{Geometric Corruption Probe}
\label{app:vlm_quality_probes}

We detect geometric corruption---melted, dissolving, or reality-breaking
structure---with a single binary probe implemented with Qwen3-VL-8B~\cite{QWEN}.
The probe receives one real reference frame followed by $16$ generated frames
distributed evenly over the available continuation, allowing it to assess
changes across the rollout rather than at a single time point.

The reference is frame $8$ of the shared real context. Generated-frame
selection is deterministic: the first and final available frames within a
six-second window are included, with the remaining frames placed uniformly
between them. The first
generated-frame indices are $9$ for ours, $4$ for Astra, $1$ for Matrix-Game,
$13$ for minWM, $65$ for WorldCam, and $1$ for WorldPlay and Yume as they are all designed to take different amounts of context. For our
$108$-frame rollouts, the selected indices are
$\{9,15,22,28,35,41,47,54,60,67,73,79,86,92,99,105\}$. Shorter baseline
rollouts are sampled over their available duration. All images are resized to
$640\times352$ and visibly labelled as \textsc{Reference} or
\textsc{Generated}.

The probe begins with the following preamble:

\begin{quote}
\small
\emph{The first image labelled REFERENCE is the last real frame of a driving
video---the true scene. The remaining images labelled GENERATED are an AI
world model's continuation of that exact scene, in temporal order. The model
was supposed to continue the SAME scene in the SAME visual style with
plausible content.}
\end{quote}

followed by the geometric-corruption question:

\begin{quote}
\small
\emph{Is there a SIGNIFICANT uncanny or reality-breaking failure in the
generated frames---impossible geometry, surfaces dissolving into abstract
patterns, large corrupted regions---that a casual viewer would notice within
one second? Ignore small local artifacts and minor blur. Answer Yes or No.}
\end{quote}

The question ends with the literal instruction
\texttt{Answer with ONLY \{"answer": "Yes"\} or \{"answer": "No"\}.}
The assistant response is force-prefixed with
\verb|{"answer": "|. No free-form text is decoded. The probe score is the
softmax probability of \emph{Yes} against \emph{No} at the next-token
position, and a rollout is flagged when $P(\mathrm{Yes})>0.5$.

\subsection{Style-Shift Instrument}
\label{app:style_instrument}

Style shift is a change in rendering rather than in content, so we measure it as
drift in a self-supervised feature space that responds to texture and rendering
statistics rather than to scene layout. We embed frames with
DINOv2 ViT-S/14~\cite{dinov2}: each frame is resized to $224\times224$ and
normalized with the standard ImageNet mean and standard deviation.

Two windows are compared. The anchor window is the first $\min(16,\text{ctx})$
real context frames; the end window is the last $16$ generated frames within the
six-second horizon. Each frame is embedded, the embeddings are
$\ell_2$-normalized, and the normalized embeddings are averaged over the window,
giving a single vector per window, $\bar e_{\mathrm{anchor}}$ and
$\bar e_{\mathrm{end}}$. The style-drift score is
\[
\texttt{dino\_drift}
=
1-\cos\!\left(\bar e_{\mathrm{anchor}},\,\bar e_{\mathrm{end}}\right),
\]
so that a continuation re-rendered in a different visual register moves far in
embedding space and scores high even when the geometry it depicts is unchanged.
A rollout is flagged when its DINOv2 drift exceeds $0.72$. This fixed
high-drift threshold is descriptive and is not tuned on the labelled
evaluation set. Although DINOv2 does not itself require keypoint matching, we
apply the common feature-valid mask because severe wet-lens corruption also
makes embedding drift relative to the source frame uninterpretable.

We also evaluated a VLM style probe as a candidate---the geometric-corruption
probe of Section~\ref{app:vlm_quality_probes} re-asked under the identical
protocol, querying whether the rendering style departs from the reference
(becoming painted, game-like, cartoonish, oversaturated, or otherwise
re-rendered), with ordinary lighting and exposure changes excluded. It is scored
comparatively and further explored in Section~\ref{app:val_style}.

\subsection{Scene-Relocation Instrument}
\label{app:scene_relocation}

Scene identity is measured geometrically and using feature matching instead of with VLMs. Frames are
resized to $640\times352$, converted to grayscale, and processed using
\[
\texttt{cv2.ORB\_create(3000)}.
\]
Descriptors are matched using
\[
\texttt{BFMatcher(NORM\_HAMMING, crossCheck=True)}.
\]
Matching therefore uses brute-force Hamming distance with mutual cross-check;
no Lowe ratio test is applied.

Matches are geometrically verified using a single homography:
\[
\texttt{cv2.findHomography(src,dst,cv2.RANSAC,5.0)},
\]
where $5.0$ is the reprojection threshold in pixels. The verified-inlier
count is the sum of the returned RANSAC mask.

For each source scene, \texttt{Reloc-Cal-v2} supplies a frozen reference comprising one representative per family: our flagship \texttt{pca8}, the six external baselines, and the four real context references used by the released descriptor pack. The relocation score is the maximum number of RANSAC-verified ORB matches to any eligible reference, and a rollout is classified as relocated when this best match contains fewer than $50$ verified correspondences. 

The quality-population static filter uses the same ORB, descriptor-matching, and homography
pipeline. It compares each rollout's first generated frame with its
six-second-horizon end frame and classifies the rollout as static when the
verified-inlier count exceeds $600$.

\subsection{Conjuration Instrument}
\label{app:conjuration_instrument}

A strange by-product of training on third-person video games is that there will always be an
actor in the foreground who is controlled via commands, with the camera view rotating around
them. Sometimes these models create such an actor even when the scene does not have one, and we
label these as videos with conjured objects present, as they seem to apparate as if by magic.

In real world footage new objects appear constantly and legitimately: they enter from the side of
the frame, they approach from the distance until they cross the detector's resolution limit,
and they emerge from behind occluders. Reporting any object new to the frame is therefore
useless, as it flags every rollout. Requiring in addition
that the object be confidently detected, born inside the central portion of the frame rather
than at its border, and still present at the end removes most of these, but only reaches $0.84$
AUC at $0.59$ precision. Because conjuration is uncommon, low specificity is especially
punishing: with a base rate of a few percent, even a small false-positive rate dominates the
flagged set.

Objects are detected in every frame and linked into tracks. A candidate
must begin after the context boundary, remain away from the left and right
borders during its first half second, reach the confidence and minimum-area
criteria, and persist near the rollout's end. Three temporal tests then
distinguish abrupt appearance from previously visible content.

The \emph{onset} is the pixel change at the birth box divided by the change over the whole
frame, so that the egomotion which moves every pixel is divided out and only a relative spike
counts; the ratio must exceed $5.0$.

The \emph{zoom probe} crops where the object is about to be, in the frames before it exists,
upscales and re-detects: a distant real car is a few pixels at native resolution and looks like
absence, but resolves into a confident vehicle when magnified, whereas empty ground stays empty
at any magnification. A re-detection confidence of $0.50$ or above counts as the object having
already been there.

The \emph{back-match} correlates the object's own settled patch into pre-birth frames at the box
its trajectory extrapolates to, catching partial occlusion reveals; a correlation of $0.65$ or
above likewise counts as prior appearance.

Detections come from RT-DETR R50~\cite{zhao2024detrs}, run over all classes at a low confidence
threshold so that the weak-detection tail is available to the prior-appearance tests, with
tracks linked from detections scoring at least $0.30$; only classes that can plausibly be
conjured are reportable (car, truck, bus, motorcycle, and a small number of rigid stationary
items). Each test is expressed as a signed margin from its threshold, and a rollout is flagged
when the smallest of the three is positive, so all three must be satisfied independently.

The conjuration decision rule is \texttt{score > 0}. Only $13/3328$ fleet rollouts lie within $0.10$ of this boundary.

\subsection{High-Frequency Degradation Instrument}
\label{app:high_frequency_metric}
\label{app:quality_results}
\label{app:high_frequency_results}

Some variants progressively lose contrast and fine image structure during
long rollouts. We measure this effect using the change in Laplacian variance,
a standard indicator of high-frequency image content.

For a temporal window \(W\), let
\[
L_r(W)
=
\frac{1}{|W|}
\sum_{t\in W}
\operatorname{Var}\!\left(\nabla^2 g_{r,t}\right),
\]
where \(g_{r,t}\) is the grayscale frame from rollout \(r\) at time \(t\).
The base window contains four frames beginning near \(t=1\)\,s, while the
end window contains eight frames sampled with stride two near the end of
the rollout.

The loss of high-frequency detail in rollout \(r\) is
\[
\Delta_r
=
L_r(W_{\mathrm{base}})
-
L_r(W_{\mathrm{end}}),
\]
so that larger positive values indicate a greater reduction in sharpness.

To account for scene difficulty and source-image quality,
\texttt{HF-Cal-v2} blocks by source scene and expresses each loss relative
to the frozen one-vote-per-family reference, the six external baselines
plus a single representative of our family:
\[
\mathcal B_r
=
\Delta_r
-
\operatorname{median}_{s\in\mathcal S_r}\Delta_s.
\]
Positive $\mathcal B_r$ means that the rollout loses more high-frequency
detail than the reference median for that scene, while negative values
mean better relative retention. Thus an ``HF 16\%'' entry means that
16\% of active rollouts exceed a \emph{peer-relative} threshold, not
that only 16\% visibly degrade in an absolute sense.

On $112$ ablation rollouts ($16$ scenes, $13$ haze positives), the
family-balanced score achieves AUC $0.68$. To summarize the continuous
score as a failure rate, we classify a rollout as high-frequency degraded
when $\mathcal B_r>150$. This fixed reporting threshold was not optimized
on the labelled evaluation set. The resulting percentage is therefore a
descriptive summary of the score distribution, while the AUC evaluates
the continuous metric independently of any particular threshold.
Dark-channel and contrast measurements are reported only as alternative
candidate metrics and do not contribute to this classification.

\begin{figure}[t]
    \centering
    \includegraphics[width=\columnwidth]
    {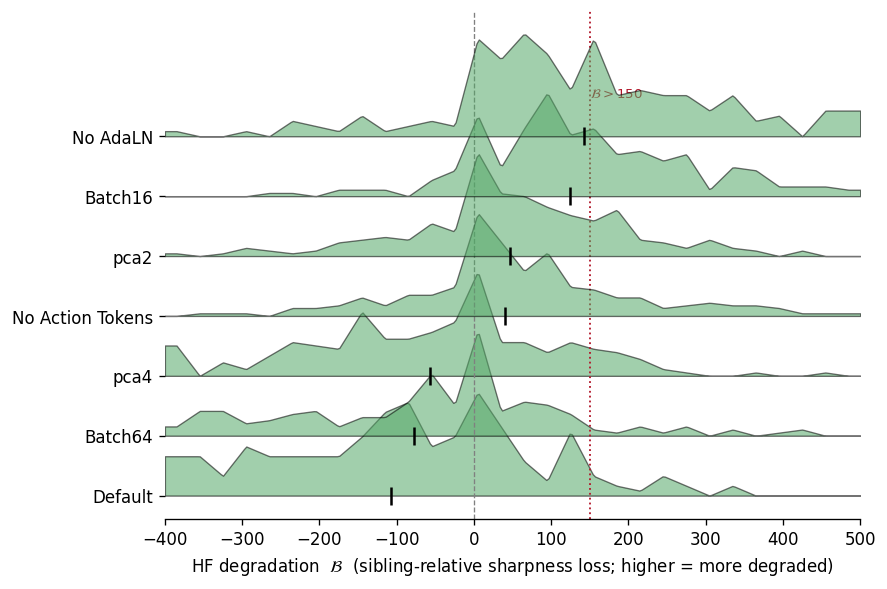}
    \caption{
        Distribution of the high-frequency-degradation score $\mathcal B$
        under \texttt{HF-Cal-v2}
        (reference-relative sharpness loss; higher is more degraded).
        \texttt{No AdaLN} and \texttt{Batch16} degrade most, while
        \texttt{Default}, \texttt{Batch64}, and \texttt{pca4} lie at or below zero and
        preserve detail better than the same-scene reference continuations.
    }
    \label{fig:hf_distribution}
\end{figure}

\begin{figure}[t]
    \centering
    \includegraphics[width=\columnwidth]
    {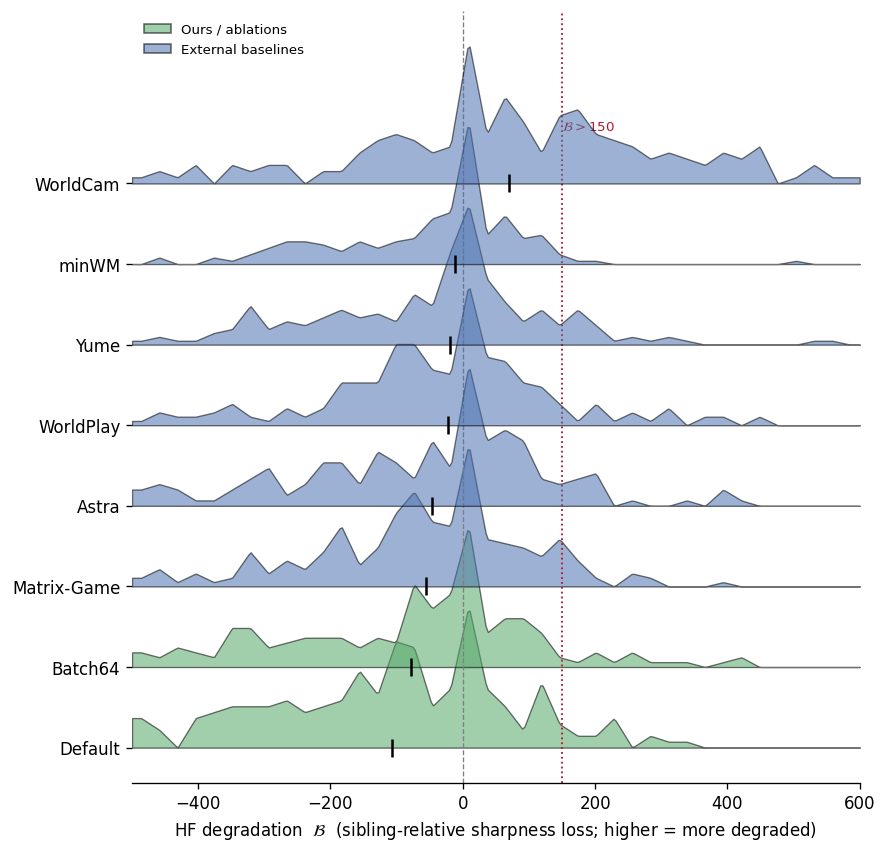}
    \caption{
        Cross-model distribution of the high-frequency-degradation score
        $\mathcal B$ under \texttt{HF-Cal-v2} for the six external
        baselines, Default and Batch64, using their active evaluation
        populations ($1{,}809$ rollouts in total). Higher values indicate
        greater reference-relative loss of sharpness. Green denotes our
        models, while blue denotes external baselines. Black ticks mark
        medians, the grey dashed line marks $\mathcal B=0$, and the red
        dotted line marks the descriptive threshold $\mathcal B>150$.
    }
    \label{fig:hf_fleet_distribution}
\end{figure}

\paragraph{Score distributions.}
Table~\ref{tab:hf_losses} reports median scores and threshold
exceedances under \texttt{HF-Cal-v2}.
Figures~\ref{fig:hf_distribution} and~\ref{fig:hf_fleet_distribution}
show the corresponding ablation and cross-model distributions.

\begin{table*}[t]
\centering
{
\small
\begin{tabular}{lrrr}
\toprule
Variant & Median $\mathcal B$ & $\mathcal B>150$ (\%) & $n/N$ \\
\midrule
Default
  & $-78$
  & $16$
  & $39/239$ \\
Batch64
  & $-51$
  & $14$
  & $33/237$ \\
pca4
  & $-10$
  & $22$
  & $52/237$ \\
No Action Tokens
  & $+48$
  & $29$
  & $66/229$ \\
pca2
  & $+61$
  & $28$
  & $66/232$ \\
Batch16
  & $+125$
  & $44$
  & $104/236$ \\
No AdaLN
  & $+154$
  & $53$
  & $106/200$ \\
No Critic
  & $+40$
  & $23$
  & $17/75$ \\
\bottomrule
\end{tabular}
}
\caption{
    High-frequency degradation under \texttt{HF-Cal-v2}. $\mathcal B$
    measures sharpness loss relative to the same-scene median of the six
    external baselines and the scored variant. Rates use model-specific
    active populations; $n/N$ counts rollouts with $\mathcal B>150$.
    This is a peer-relative exceedance rate, not absolute degradation
    prevalence. No additional sharpness-quantile exclusion is applied.
}
\label{tab:hf_losses}
\end{table*}

\section{Reference-Free Quality Evaluation Validation}
\label{app:quality_validation}

We validate all five quality instruments against human annotations. Style
shift, geometric corruption, scene relocation, and conjuration are validated
on shared or enriched labelled sets. High-frequency degradation is validated
separately on the ablation set, where positive examples are more concentrated,
and then evaluated on the active cross-model populations defined in
Section~\ref{app:high_frequency_metric}.

It should be noted that evaluating these rollouts qualitatively without a reference requires nuance and subtlety; which means that some rollouts will inevitably be wrongly evaluated, however this is an unavoidable consequence of such an evaluation process, and we leave further performance gains and ensembling approaches.

\subsection{Geometric Corruption}
\label{app:val_geometry}

Egomotion models re-render scene geometry as the viewpoint changes, and their
characteristic failure is localized over-warping: a surface keeps plausible
colour and coarse shape while its internal structure melts, duplicates, or
becomes impossible. Whole-frame pixel and line-segment statistics miss this, so
we compared a VLM probe that judges melted or geometrically impossible structure
against a generative artifact segmenter (PAL4VST~\cite{pal}) and a melt-specific
VLM prompt evaluated on left/right crops. Each candidate was scored against $36$
human ``mangle'' labels on the $100$-rollout subset
(Table~\ref{tab:val_geometry}).

The VLM geometric-corruption probe agrees best with human judgement at $0.78$
AUC. PAL4VST is close ($0.77$) but redundant with it. We evaluated the same
probe across three open VLMs to confirm the choice of backbone: Qwen3-VL-8B is
strongest, ahead of InternVL3-8B and Cosmos-Reason1-7B
(Table~\ref{tab:val_geometry}). We therefore deploy the single Qwen3-VL probe
and flag a rollout when $P(\mathrm{Yes})>0.5$; combining it with the segmenter
improves AUC only within noise.

\begin{figure*}[t]
    \centering
    \includegraphics[
        width=0.82\textwidth
    ]{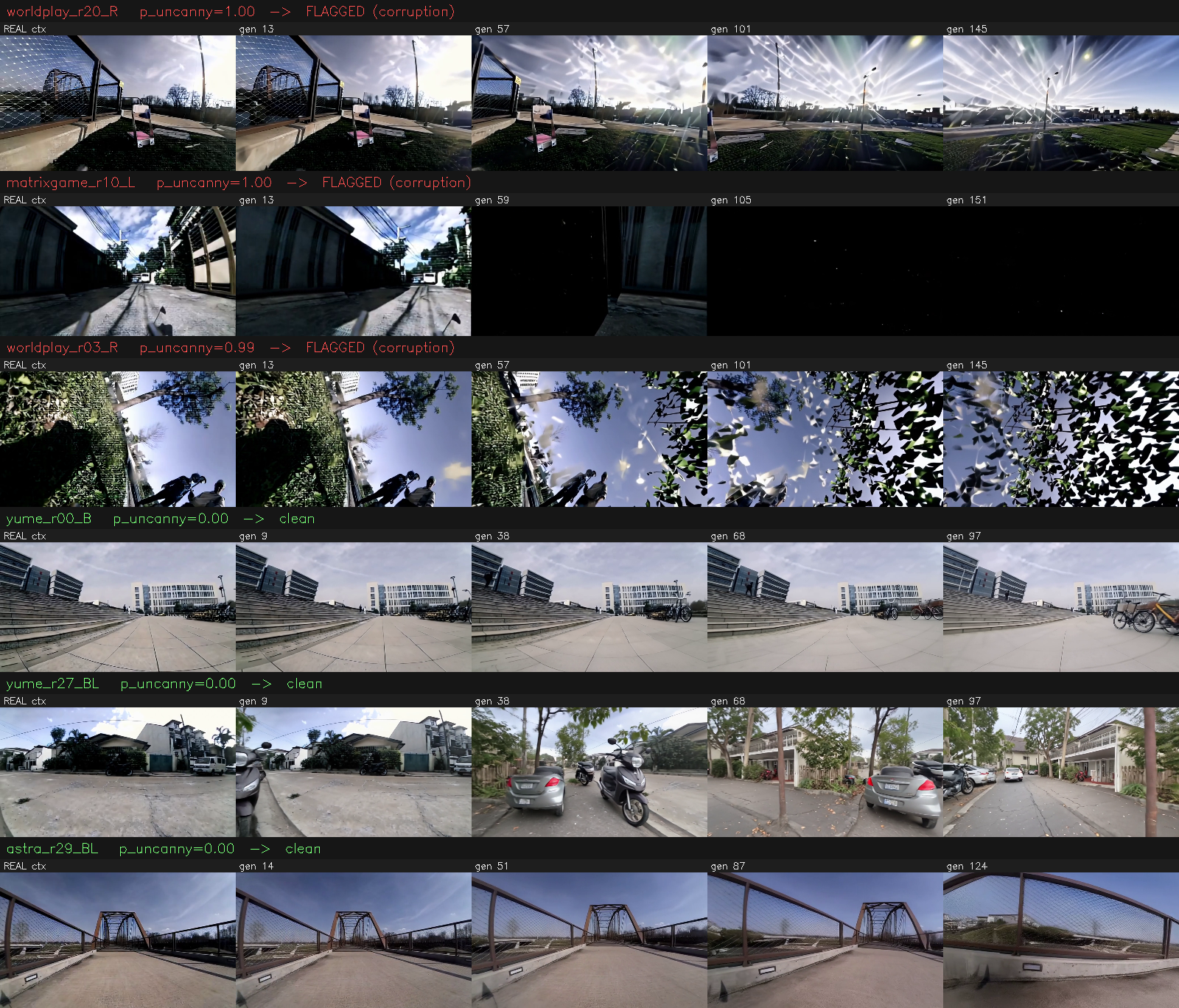}
    \caption{
        Geometry-instrument examples for the external baselines, showing
        representative human-positive and human-negative cases. Yume r27 relocates
        without visible geometric corruption, illustrating why relocation needs a separate instrument.
    }
    \label{fig:val_geometry_external}
\end{figure*}

\begin{figure*}[t]
    \centering
    \includegraphics[
        width=0.82\textwidth
    ]{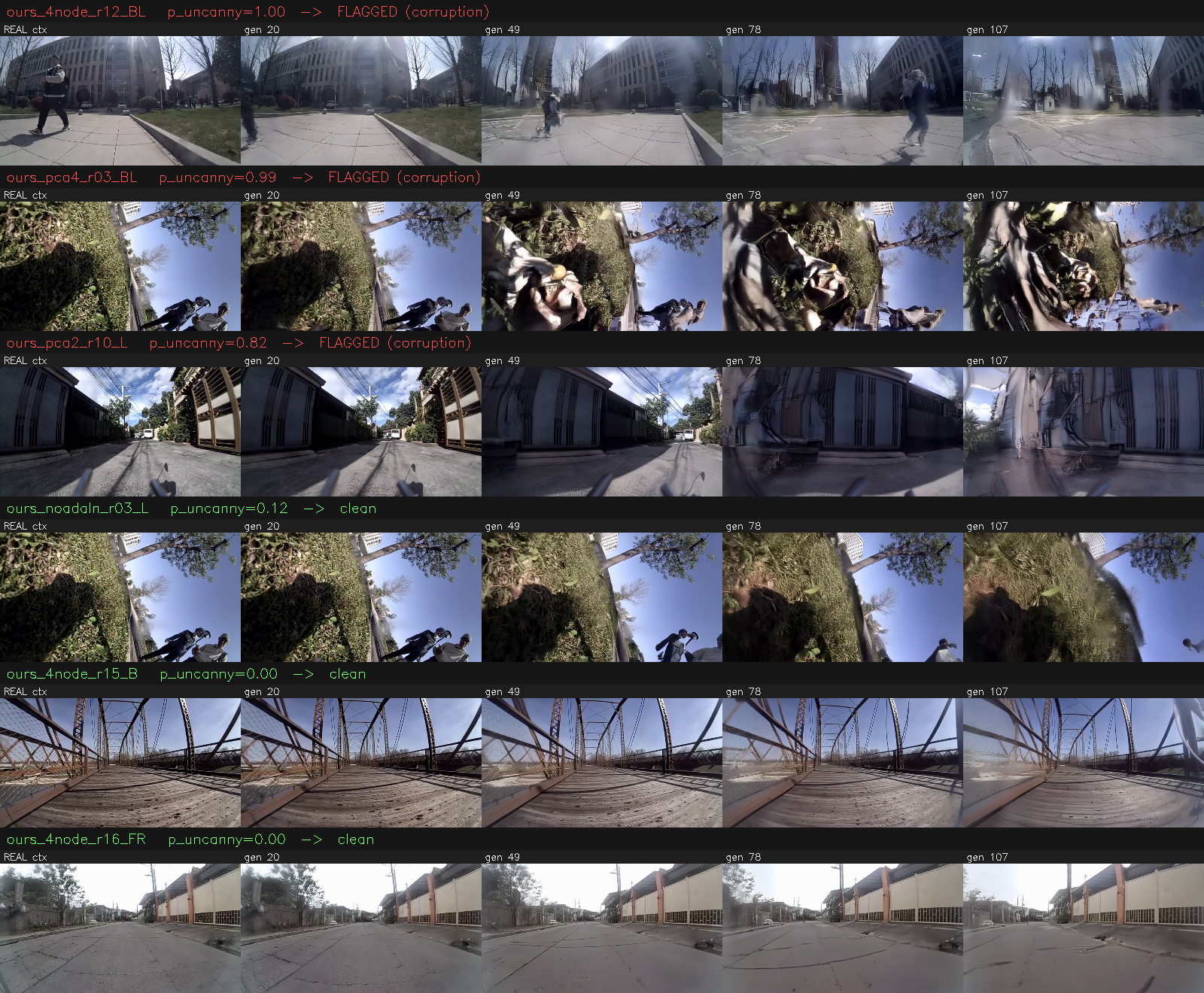}
    \caption{
        Geometry-instrument examples for our ablation variants, showing
        representative human-positive and human-negative cases.
    }
    \label{fig:val_geometry_internal}
\end{figure*}

As shown in Figures~\ref{fig:val_geometry_external} and ~\ref{fig:val_geometry_internal}, the geometry / warping metric earns its keep but it classifies one of the scene relocated rollouts (yume\_r27) as clean, we note that this is because the metric does not find any geometry warping and it is not designed to identify scene relocation. This further necessitates separate axes controls.

\begin{table}[t]
\centering
{
\small
\begin{tabular}{lc}
\toprule
Detector                                              & AUC \\
\midrule
\textbf{VLM geometric-corruption probe (Qwen3-VL-8B)} & \textbf{0.78} \\
\quad same probe, InternVL3-8B                        & 0.75 \\
\quad same probe, Cosmos-Reason1-7B                   & 0.74 \\
PAL4VST artifact fraction                             & 0.77 \\
VLM melt probe, Cosmos-Reason1-7B                     & 0.71 \\
VLM melt probe, Qwen2.5-VL-7B                         & 0.68 \\
Depth-field curvature (Depth-Anything-V2)             & 0.52 \\
\bottomrule
\end{tabular}
}
\caption{
    Geometric-corruption detectors scored against $36$ human mangle labels on
    the $100$-rollout subset. The deployed VLM probe (Qwen3-VL-8B) agrees best
    with human judgement; the same probe is evaluated across three open VLMs.
    The depth-curvature signal, part of an earlier ensemble, performs near
    chance ($0.52$) and is not used.
}
\label{tab:val_geometry}
\end{table}

\subsection{Style Shifts}
\label{app:val_style}

A style shift is a change in rendering rather than in content:
the continuation departs from the reference by becoming painted, game-like,
cartoonish, oversaturated, or otherwise re-rendered in a different visual
register, while the underlying scene layout may stay the same.

We compared a DINOv2 embedding drift against a CLIP embedding distance, a direct
style probe evaluated across three open VLMs (Qwen3-VL-8B~\cite{QWEN},
InternVL3-8B~\cite{internvl3}, and Cosmos-Reason1-7B~\cite{cosmos2}), and a
classical VGG Gram-matrix style distance~\cite{gatys2016image}, scoring each against
$20$ human style labels on the $100$-rollout subset
(Table~\ref{tab:style_metric_auc}). DINOv2 drift agrees best with human judgement
at $0.72$ AUC: the VLM probes asked directly whether the rendering changed land
lower and disagree with one another ($0.55$--$0.67$), the CLIP distance is close
but weaker ($0.70$), and the Gram-matrix distance is near chance. We therefore
deploy DINOv2 drift.

\begin{table}[t]
\centering
{
\small
\begin{tabular}{lc}
\toprule
Signal                                     & AUC \\
\midrule
\textbf{DINOv2 embedding drift (deployed)} & $\mathbf{0.72}$ \\
CLIP embedding distance                    & $0.70$ \\
Cosmos-Reason1-7B style probe              & $0.67$ \\
Qwen3-VL-8B style probe                    & $0.66$ \\
InternVL3-8B style probe                   & $0.55$ \\
VGG Gram-matrix distance                   & $0.55$ \\
\bottomrule
\end{tabular}
}
\caption{
    Candidate style-shift measures evaluated by AUC on the human-labelled set
    containing $20$ positives among $100$ examples.
}
\label{tab:style_metric_auc}
\end{table}

\begin{figure*}[t]
    \centering
    \includegraphics[
        width=0.82\textwidth
    ]{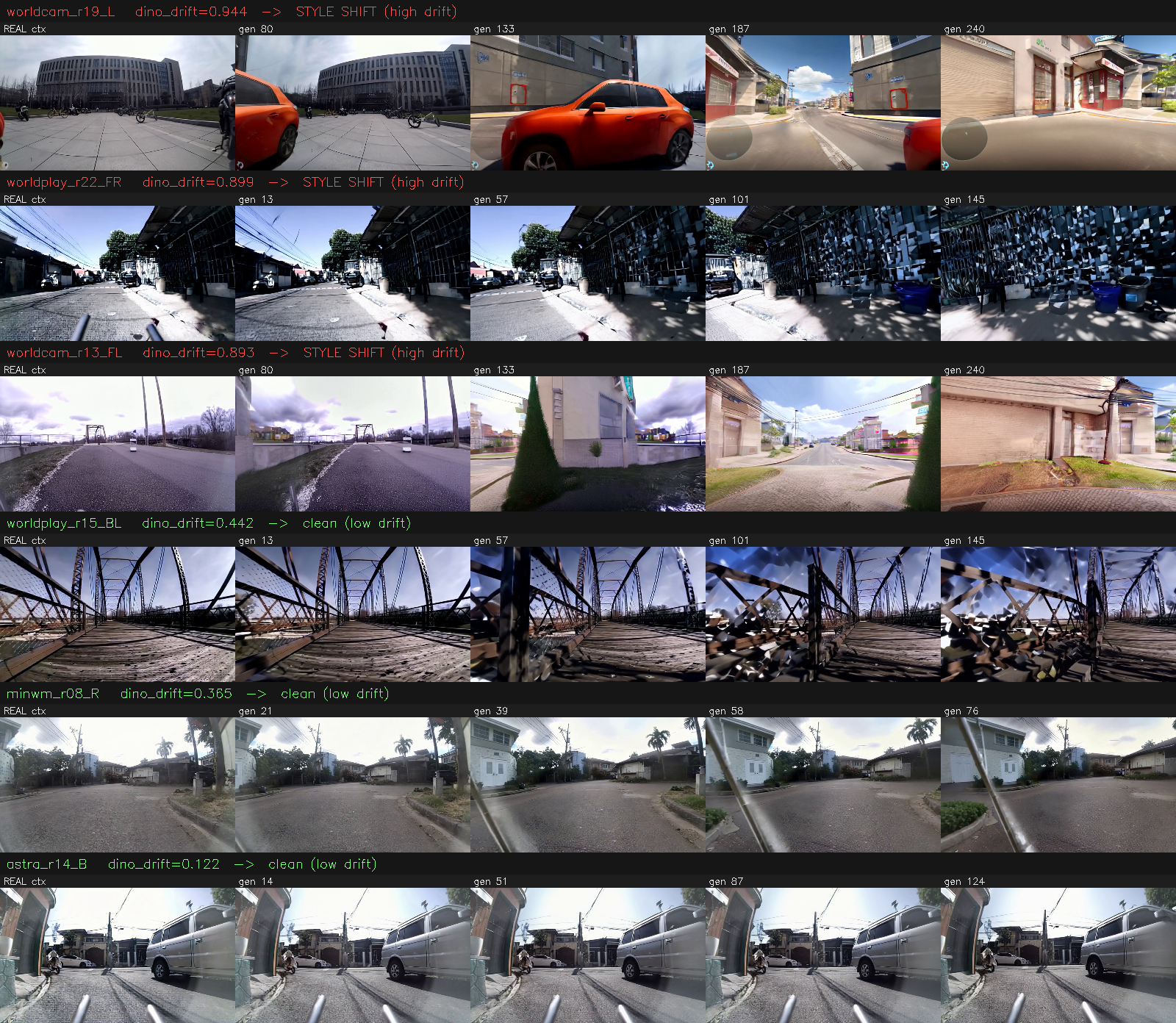}
    \caption{
        Style-shift-instrument examples for the external baselines, showing
        representative flagged and clean rollouts.
    }
    \label{fig:style_external}
\end{figure*}

\begin{figure*}[t]
    \centering
    \includegraphics[
        width=0.82\textwidth
    ]{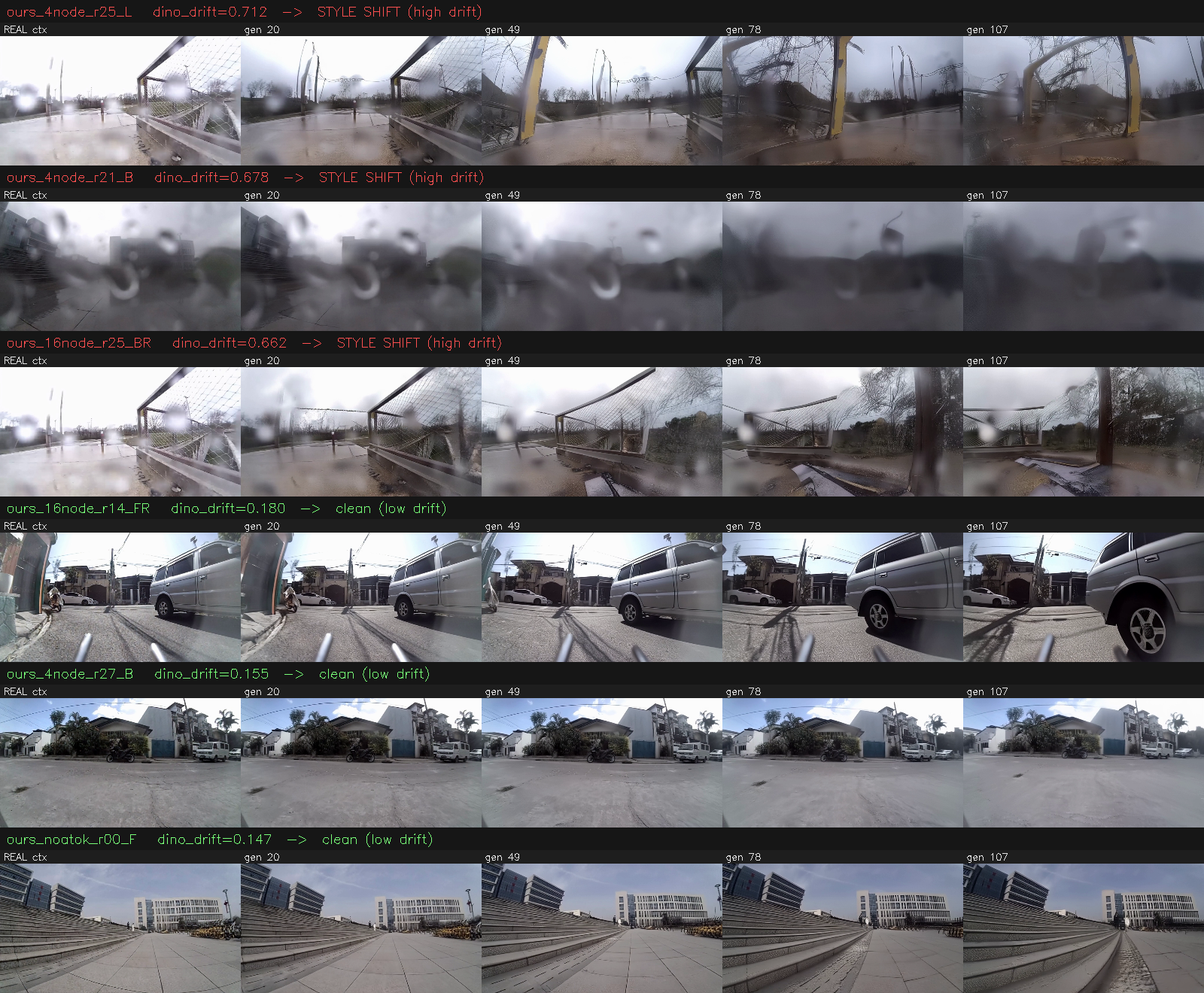}
    \caption{
        Style-shift-instrument examples for our ablation variants, showing
        representative flagged and clean rollouts.
    }
    \label{fig:style_internal}
\end{figure*}

As Figures~\ref{fig:style_external} and~\ref{fig:style_internal}
show, changes in weather or lighting are classified as style shifts which is permissible depending on the definition of style. However, it should be noted that sudden weather changes are possible and plausible, but it takes nuance to judge whether the model is arbitrarily changing the weather or if it is done in a correct way, so we permit it here.

\subsection{Scene Relocation}
\label{app:val_scene}

A model can abandon the input scene and generate a plausible but different
place. We detect this geometrically rather than semantically: ORB keypoints are
matched between the generated six-second frame and the real reference, as well
as the end frames of sibling rollouts. A projective homography is fitted with
RANSAC, and the number of geometrically consistent inliers is counted. A
continuation that still depicts the same place retains many verified
correspondences, whereas a relocated continuation does not. A rollout is
flagged when fewer than $50$ geometrically verified inliers survive against
every available reference.

We compared feature detectors and embedding-drift alternatives against $19$
human relocation labels in a $100$-video subset
(Table~\ref{tab:val_scene}). Among classical detectors, ORB with RANSAC agrees
best with human judgement ($0.70$ AUC), ahead of AKAZE ($0.63$) and SIFT
($0.62$). RANSAC verification is important, as raw ORB matching drops to
$0.54$. DINOv2 embedding drift scores marginally higher ($0.72$), but provides
only a relative, uncalibrated distance.
We deploy ORB+RANSAC because its verified-inlier count is directly interpretable and requires no learned model, and because the same primitive supports static-rollout detection.

\begin{table}[t]
\centering
{
\small
\begin{tabular}{lc}
\toprule
Detector                        & AUC \\
\midrule
ORB + RANSAC inliers (deployed) & 0.70 \\
\textbf{DINOv2 embedding drift} & \textbf{0.72} \\
AKAZE + RANSAC inliers          & 0.63 \\
SIFT + RANSAC inliers           & 0.62 \\
ORB matches, no RANSAC          & 0.54 \\
\bottomrule
\end{tabular}
}
\caption{
    Scene-relocation detectors evaluated on a $100$-video human-labelled subset
    containing $19$ relocation positives. Higher AUC is better. ORB+RANSAC is
    deployed because it provides an absolute inlier count and also supports
    static-rollout detection. The deployed relocation decision uses this inlier count against the frozen peer panel and is therefore panel-relative rather than a standalone absolute rate.
}
\label{tab:val_scene}
\end{table}

\begin{figure*}[t]
    \centering
    \includegraphics[
        width=0.82\textwidth
    ]{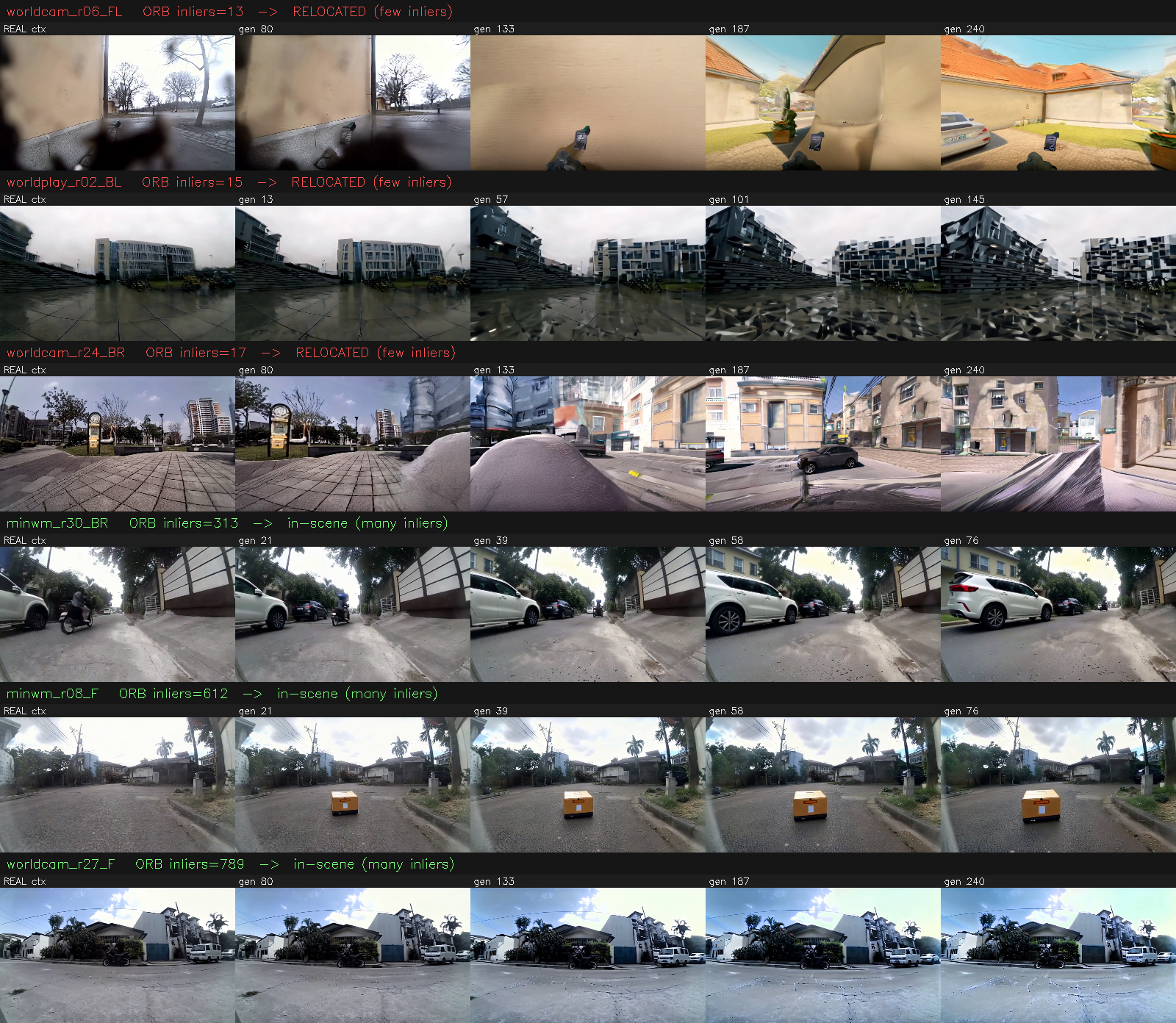}
    \caption{
        Scene-relocation-instrument examples for the external baselines,
        showing representative relocated and retained-scene rollouts.
    }
    \label{fig:scene_external}
\end{figure*}

\begin{figure*}[t]
    \centering
    \includegraphics[
        width=0.82\textwidth
    ]{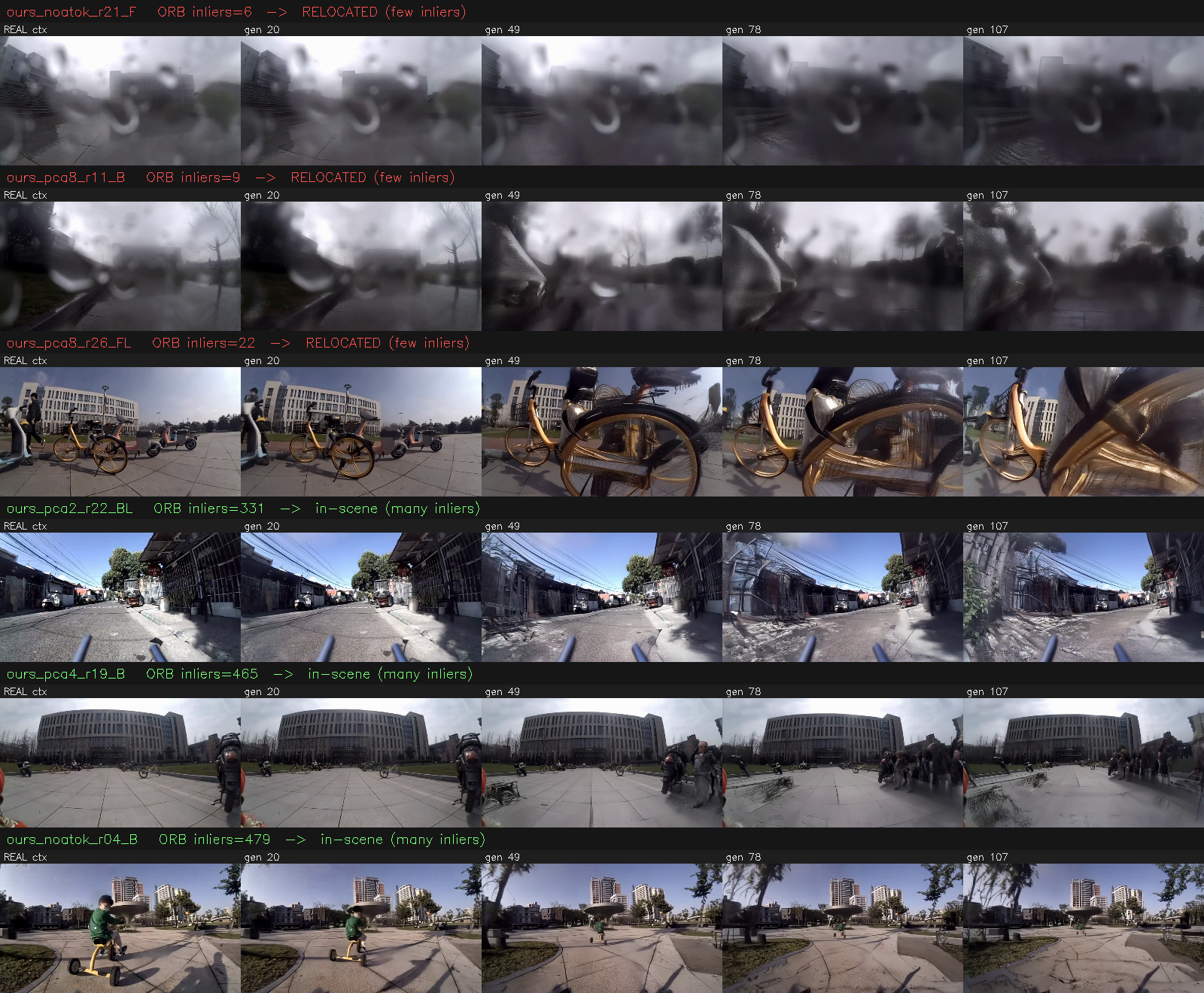}
    \caption{
        Scene-relocation-instrument examples for our ablation variants.
        Post-collision content and severe lens occlusion illustrate ambiguous
        flagged cases that suppress geometric correspondences without
        necessarily changing the underlying scene.
    }
    \label{fig:scene_internal}
\end{figure*}

Representative successes and failure cases are shown in
Figures~\ref{fig:scene_external} and~\ref{fig:scene_internal}.
Figure~\ref{fig:scene_internal} includes ambiguous flagged cases caused by
post-collision content and severe lens occlusion. Both can suppress geometric
correspondences without necessarily indicating true scene relocation. We retain
these rare edge cases rather than tune the detector specifically around them.

\subsection{Conjuring}
\label{app:conj}

This is rare in our clips
and uncommon in most other baselines video, but certain models are prolific in it, so rather than reuse the
$100$-rollout subset (which contains too few positives to separate detectors) we assemble
a larger set enriched with true positives: $15$ confirmed
conjurations drawn from several models together with $22$ rollouts confirmed
clean, so that both precision and recall are measurable.

The temporal logic is agnostic to the detector backbone, so we validate that
choice across three architecturally distinct backbones under an identical
pipeline (Table~\ref{tab:val_conj}). RT-DETR and Faster R-CNN \cite{girshick2015fast}
are indistinguishable on this set---identical precision, recall and $F_1$, differing on no
rollout---so we deploy RT-DETR because it is $1.7\times$ faster. RetinaNet \cite{goyal2018focal}
loses one further rollout, and does so only because that rollout falls exactly on the decision
threshold rather than because it ranks it differently.

\begin{table}[t]
\centering
{
\small
\begin{tabular}{lcccc}
\toprule
Backbone                        & Prec. & Rec. & $F_1$ & fps \\
\midrule
\textbf{RT-DETR R50 (deployed)} & 1.00 & 0.80 & 0.89 & \textbf{94.8} \\
Faster R-CNN R50-FPNv2          & 1.00 & 0.80 & 0.89 & 56.3 \\
RetinaNet R50-FPNv2             & 1.00 & 0.73 & 0.85 & 69.6 \\
\bottomrule
\end{tabular}
}
\caption{
    Detection backbones under an identical temporal pipeline, scored on the
    enriched evaluation set: $15$ confirmed conjurations drawn from several
    models together with $22$ rollouts confirmed clean, so that both precision
    and recall are measurable. Throughput is end-to-end detection plus temporal
    analysis on a single RTX 5090.
}
\label{tab:val_conj}
\end{table}

\begin{figure*}[t]
    \centering
    \includegraphics[width=0.82\textwidth]{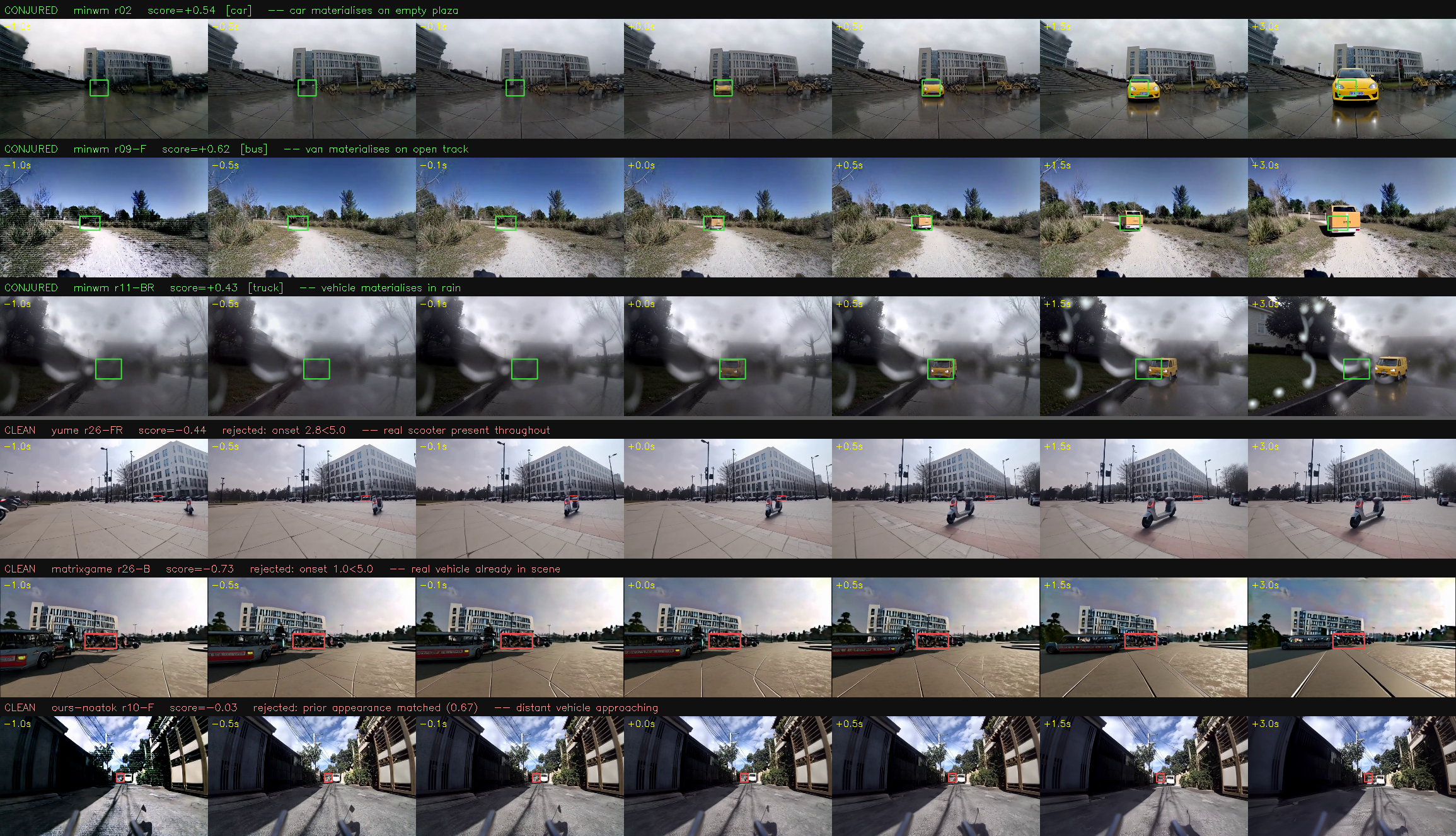}
    \caption{
        Conjuration-instrument examples. Top: flagged rollouts centred on object
        birth, showing empty ground beforehand and a persistent object afterwards.
        Bottom: hard negatives in which new objects arrive legitimately; annotations
        identify the prior-appearance test that rejected each candidate.
    }
    \label{fig:conj_validation}
\end{figure*}

Figure~\ref{fig:conj_validation} shows the instrument at work: flagged rollouts
(top) have empty ground for the full second before the birth frame, after which a
crisp vehicle occupies it and persists; the clean rollouts (bottom) are the hard
cases a novel-object detector would fail on---a scooter present throughout, a
vehicle already in the scene, and a distant vehicle approaching---each annotated
with the test that rejected it.

Deployed over the full directional fleet, capped at six seconds of generation
per rollout, the instrument flags $2.1\%$ of rollouts. The rate is highly
non-uniform: minWM is flagged in $19.9\%$ of its rollouts, compared with
$0.84\%$ across our seven variants and $0.23\%$ across the other five external
baselines. Two controls indicate this reflects the model rather than the instrument. First,
the same scenes as real footage are flagged $0$ times out of $32$, so the detector does not
fire on genuine driving video. Second, and more telling, the gap is already present
\emph{before} any prior-appearance evidence is applied: minWM produces a persistent, crisp,
newborn interior object in $52.7\%$ of its rollouts, against $6.6$--$16.4\%$ for every other
model. The instrument is not simply more sensitive on minWM; minWM spawns far more new
persistent objects, and a sixth of its rollouts contain one that no prior-appearance test can
account for.

\subsection{High-Frequency Degradation}
\label{app:val_hf}

High-frequency degradation is most pronounced in some of our ablations, so we
validate the instrument on the $16$ ablation scenes containing all seven
variants ($112$ rollouts), of which $13$ are human-labelled as degraded.

We compared the drift in Laplacian variance (a standard sharpness measure), the
drift in a dark-channel haze estimate, and the drift in global contrast, each
scored against those $13$ haze labels (Table~\ref{tab:val_hf}).

\begin{table*}[t]
\centering
{
\small
\begin{tabular}{lc}
\toprule
Detector & AUC \\
\midrule
Laplacian drift, original analysis                    & $0.85$ \\
Laplacian drift, original-panel rescore               & $0.78$ \\
Laplacian drift, family-balanced (deployed)           & $0.68$ \\
Dark-channel haze drift, sibling-relative             & $0.82$ \\
Dark-channel haze drift, raw                          & $0.77$ \\
Laplacian-variance drift, raw                         & $0.68$ \\
Contrast drift                                        & $0.64$ \\
\bottomrule
\end{tabular}
}
\caption{
    High-frequency-degradation validation on $112$ rollouts with $13$ haze positives.
    The first three rows separate the original Laplacian analysis from rescoring under
    the original and family-balanced references. Other candidate results are from the
    original comparison, not a new matched comparison against \texttt{v2}.
}
\label{tab:val_hf}
\end{table*}

Figure~\ref{fig:hf_internal} shows representative rollouts to which the
high-frequency-degradation instrument is applied.

\begin{figure*}[t]
\color{red}
    \centering
    \includegraphics[width=0.82\textwidth]{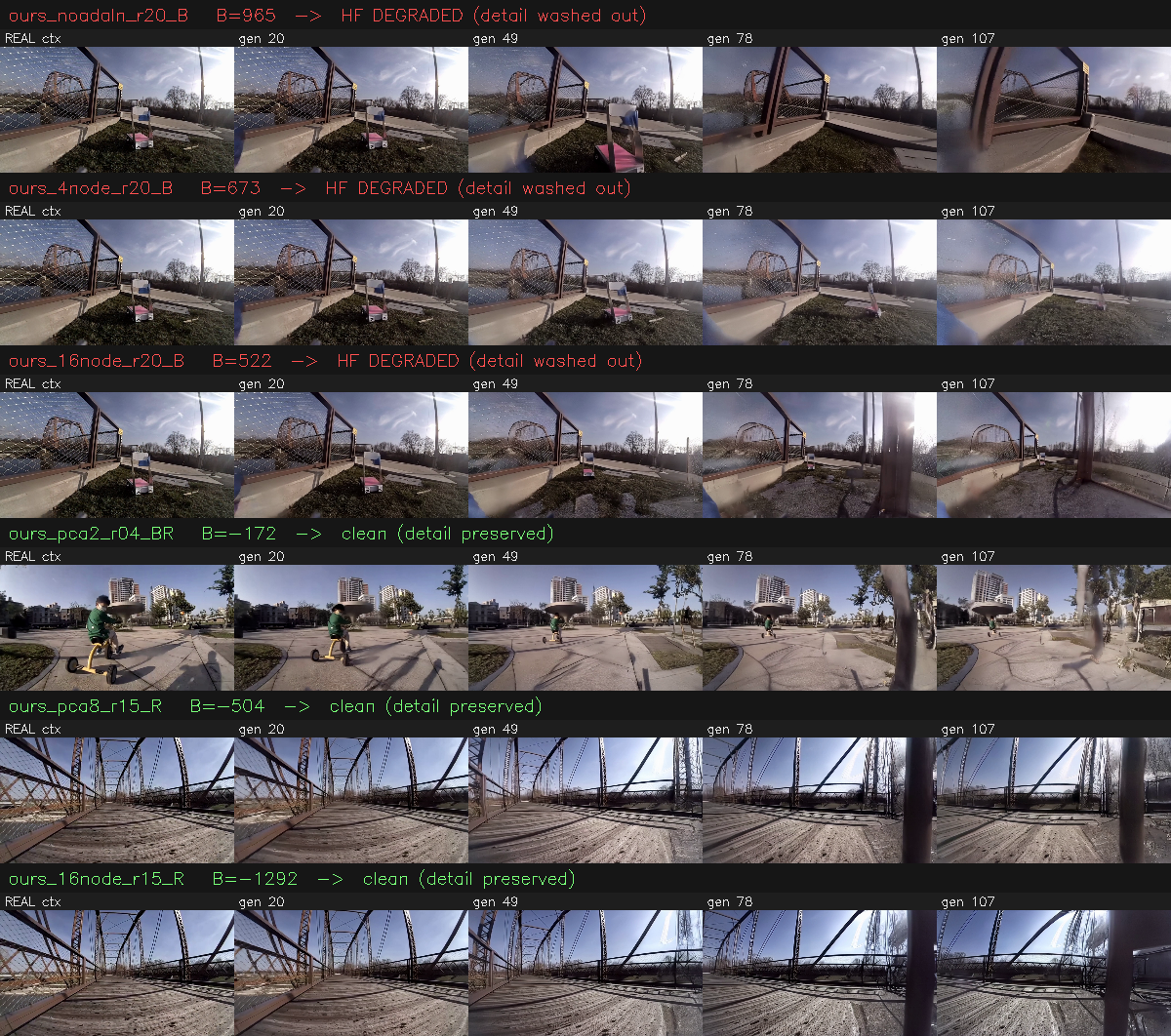}
    \caption*{
        The high-frequency-degradation instrument on our ablation variants. High-$B$
        rollouts (top) visibly wash out over the generated span---texture and contrast are
        lost by the end---while low-$B$ rollouts (bottom) preserve detail. The same source
        scenes appear in both classes, so the effect is model-dependent, not scene-dependent.}

    \reviewoldlabel{fig:hf_internal}{29}
\end{figure*}

\section{Full Stationary Evaluation}
\label{app:stationary_results}

\subsection{Controllability}
\label{app:stationary_controllability}

We apply a stationary command to every model using the protocol in
Appendix~\ref{app:baseline_controls}. We use raw PCA projections of the mean per-frame CoTracker grid flow
(pixels/frame), before per-component tanh squashing. Let $p_s\in\mathbb R^2$
contain the first two mean raw projections for scene $s$, and let
$p_s^{\mathrm{real}}$ be the corresponding real-continuation readout. We report
\[
\mathrm{Movement}=\frac{1}{32}\sum_{s=1}^{32}
\left\|p_s-p_s^{\mathrm{real}}\right\|_2.
\]
Movement is thus measured in raw PCA-projection units, not pixel displacement
or squashed action units; there is no additional scale factor or temporal
accumulation. Lower is better. REAL is $0.0$ by construction because its
readout is compared with itself, not because its measured motion is exactly
zero. This differs from the squashed readout used for the no-op control
threshold in Section~\ref{sec:stationarity}.

\begin{figure*}[t]
    \centering
    \includegraphics[width=\textwidth]{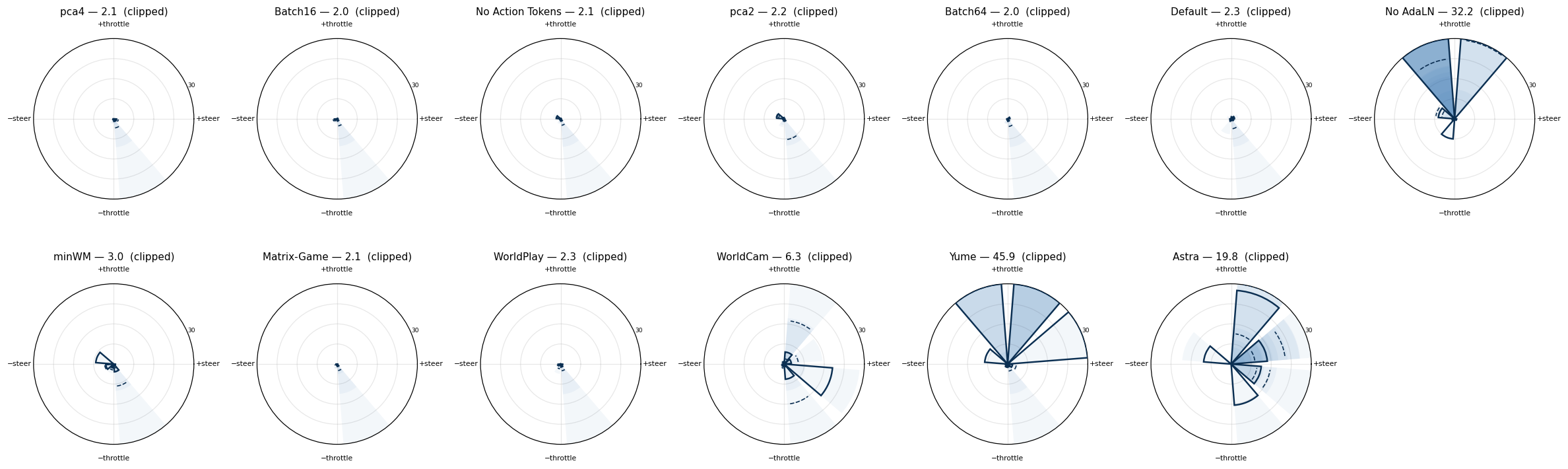}
    \caption{
    Residual egomotion under the null command. Each translucent wedge
    represents one of $32$ contexts. Its angle and radius describe the
    two-component raw PCA departure from the real continuation; radius uses
    the same raw PCA-projection units as Movement, before tanh squashing.
    Blue denotes forward drift and red backward drift. Solid outlines show
    medians and dashed arcs show interquartile ranges.
}
    \label{fig:stat_cont}
\end{figure*}

Figure~\ref{fig:stat_cont} shows the
corresponding departures. 

\paragraph{Animation}
Low Movement alone can reward a frozen scene. Animation therefore measures
residual CoTracker motion after removing the top three PCA modes. Tracks
are binned into an $8\times8$ grid; a cell is alive if it has at least three
tracks, residual magnitude above $2$ pixels and coherence ratio
$\|\mathrm{mean}\|/\mathrm{mean}(\mathrm{std})>1.0$. Animation is the mean count
of alive cells (0--64). It is a proxy for independent scene activity, not an
object count: one mover can activate several cells. 

Two confounds remain. Depth-dependent parallax can inflate Animation in
strongly drifting models. Native bitrate also differs: our variants encode
at about $7.5$ Mbps and external outputs at about $1.6$ Mbps; Animation has a
weak positive association with bitrate (Spearman $0.185$, $p=0.003$, $n=256$).
Matrix-Game's and WorldPlay's near-zero values therefore suggest suppressed
scene activity but are not an unconfounded measure of freezing.
Table~\ref{tab:anim} reports both axes.
\begin{table}[t]
\centering
{
\small
\begin{tabular}{lcc}
\toprule
Model                     & Movement       & Animation \\
\midrule
REAL (reference)          & $0.0$          & $1.1$ \\
\midrule
Ours (Default)            & $2.3$          & $1.4$ \\
Ours (pca4)               & $2.1$          & $0.0$ \\
Ours (pca2)               & $2.2$          & $0.0$ \\
Ours (Batch64)            & $2.0$          & $0.6$ \\
Ours (Batch16)            & $2.0$          & $0.0$ \\
Ours (No Action Tokens)   & $2.1$          & $0.1$ \\
Ours (No AdaLN)           & $32.2$         & $6.2$ \\
Ours (No Critic) & $1.8$ & $0.2$ \\
\midrule
minWM                     & $3.0$          & $0.2$ \\
Matrix-Game               & $2.1$          & $0.0$ \\
WorldPlay                 & $2.3$          & $0.0$ \\
WorldCam                  & $6.3$          & $2.8$ \\
Astra                     & $19.8$         & $11.2$ \\
Yume                      & $45.9$         & $0.8$ \\
\bottomrule
\end{tabular}
}
\caption{
    Stationarity and animation over $32$ null-command contexts.
    \emph{Movement} is the mean distance between generated and real-continuation
    readouts in raw PCA-projection units, before tanh; lower is better and
    REAL $=0.0$ by construction. \emph{Animation} is the mean count of alive
    cells in an $8\times8$ grid after removing the top three PCA motion modes,
    not a count of objects. A faithful no-op should hold the viewpoint without
    freezing the world. Animation is most interpretable for low-Movement
    models and is partly confounded by native bitrate.
}
\label{tab:anim}
\end{table}

\subsection{Quality}
\label{app:stationary_quality}

Table~\ref{tab:stationary_quality} reports stationary quality failure rates across the model variants and external baselines. 
Our full-capacity variants have $0\%$ style shift, $0\%$ geometric corruption
and $3\%$ relocation. Matrix-Game and Astra show geometric corruption in
$94\%$ and $44\%$ of stationary rollouts, respectively. Peer-referenced
relocation is $41\%$ for Astra, $25\%$ for Yume and $16\%$ for WorldCam.
These failures are distinct from the no-op motion test: low measured camera
motion need not imply an intact scene.
\begin{table}[t]
\centering
\small
\setlength{\tabcolsep}{1mm}
\begin{tabular}{@{}lccccc@{}}
\toprule
Model            & Style\,\% & Geom.\,\% & Reloc.\,\% & Conj.\,\% & HF\,\% \\
\midrule
Default          & 0   & 0   & 3   & 0   & 3   \\
pca4             & 0   & 0   & 3   & 0   & 0   \\
pca2             & 0   & 0   & 3   & 0   & 9   \\
Batch64          & 0   & 0   & 3   & 0   & 0   \\
Batch16          & 0   & 0   & 3   & 0   & 12  \\
No Act. Tok.     & 0   & 0   & 3   & 0   & 3   \\
No AdaLN         & 0   & 3   & 12  & 0   & 66  \\
No Critic        & 0   & 0   & 3   & 0   & 6   \\
\midrule
minWM            & 0   & 0   & 6   & 31  & 9   \\
Matrix-Game      & 0   & 94  & 9   & 0   & 3   \\
WorldPlay        & 3   & 16  & 12  & 0   & 12  \\
WorldCam         & 3   & 19  & 16  & 0   & 38  \\
Astra            & 6   & 44  & 41  & 0   & 6   \\
Yume             & 0   & 6   & 25  & 0   & 16  \\
\bottomrule
\end{tabular}
\caption{
    Stationary-rollout failure rates (\%); lower is better.
    Style, Geom., Reloc., Conj., and HF denote style shift, geometric
    corruption, scene relocation, conjuration, and high-frequency
    degradation. HF and relocation are peer-referenced against the
    family-balanced \texttt{v2} reference (one seat per family;
    a variant of ours occupies our seat itself when scored);
    the remaining quality axes are panel-independent.
}
\label{tab:stationary_quality}
\end{table}

The one ablation that stands out is \texttt{noadaln}, which washes out in $66\%$
of stationary rollouts (median $\mathcal B=+228$ under \texttt{HF-Cal-v2}) and relocates in $12\%$.

\section{Training Infrastructure and Statistical Protocol}
\label{app:compute}

\subsection{Hardware and software}

All models were trained on an HPE Cray EX system with NVIDIA GH200
Grace--Hopper nodes: four GPUs per node, 95\,GiB of usable HBM per GPU, compute
capability 9.0, 128 CPU cores per task, and a Slingshot-11 interconnect
(NCCL over the \texttt{hsn} interfaces). Training used PyTorch 2.10.0 with CUDA
12.8, distributed data-parallel across nodes, and LoRA adapters over a frozen
Wan2.1-T2V-1.3B backbone.

The Default model and every conditioning ablation (PCA-2, PCA-4, No-Act-Tokens,
No-AdaLN, No-Critic) were trained on 8 nodes = 32 GPUs at global batch 32 and
learning rate $4.95\times10^{-5}$. The batch-size ablations used 4 nodes = 16
GPUs (global batch 16, lr $3.5\times10^{-5}$) and 16 nodes = 64 GPUs (global
batch 64, lr $7.0\times10^{-5}$). Every variant trains from step 0 to step 5000.
The scheduler enforces a four-hour limit per job, so each run proceeds as a
short sequence of checkpoint-resumed segments; Table~\ref{tab:compute} reports
the total.

\begin{table}[t]
\centering
\small
\begin{tabular}{lrrrr}
\toprule
Variant         & Nodes & GPUs & Wall-clock & GPU-h \\
\midrule
Default (PCA-8) & 8  & 32 & 26.9\,h & 862 \\
Batch64         & 16 & 64 & 28.4\,h & 1819 \\
Batch16         & 4  & 16 & 27.4\,h & 439 \\
PCA-4           & 8  & 32 & 27.5\,h & 879 \\
PCA-2           & 8  & 32 & 26.9\,h & 860 \\
No-Act-Tokens   & 8  & 32 & 21.0\,h & 671 \\
No-AdaLN        & 8  & 32 & 26.7\,h & 854 \\
No-Critic       & 8  & 32 & 26.8\,h & 857 \\
\midrule
\multicolumn{4}{l}{Total} & $\approx$7{,}250 \\
\bottomrule
\end{tabular}
\caption{Measured training cost per reported model.}
\label{tab:compute}
\end{table}

\subsection{One run per variant, many rollouts per run}

Each reported model is a \textbf{single training run} at \texttt{seed:\ 0},
offset per rank so that data order and initialisation are deterministic. We
report no seed replicates. At roughly 850 GPU-hours per variant, a three-seed
replication of the eight-model table would cost on the order of 22{,}000
GPU-hours, which was not available to us; we therefore spent the budget on
breadth of evaluation instead of depth of replication.

Training is not bit-reproducible past the first step, because the backward pass
uses non-deterministic kernels. Measured run-to-run loss agreement is
$\sim$$10^{-3}$, against a loss that moves by $\sim$0.09 between adjacent steps
of a single run, so the residual is well inside one run's own variation.

Generation itself is also hardware-sensitive along the autoregressive chain. With identical weights and seeds on different GPU hardware, first chunks are near-identical (mean absolute difference in frame mean $0.43$) but diverge by a factor of roughly 30 by chunk 8 ($13.08$). The stationary Movement statistic is robust in the measured rerun (per-scene $r=0.995$, maximum $|\Delta|=2.2$), whereas Animation is not ($r=0.50$; for example r07 $9\rightarrow0$ and r17 $19\rightarrow1$). Stationary-table reruns should therefore use generations from the same hardware. Hardware robustness was not measured for HF, relocation, geometry, conjuration, or style, and we make no stability claim for those five metrics.

\subsection{Evaluation Uncertainty}
\label{app:uncertainty}

Each of our variants is trained once. Each evaluated model
produces $256$ directional and $32$ no-op rollouts. Legitimacy is
computed on the common feature-valid subset of $240$ directional
rollouts: $30$ source contexts with eight commands each. These
rollouts measure evaluation variation, not independent training runs.

For the Default--minWM legitimacy comparison, we account for
shared source contexts by keeping all eight commands and cross-model
pairing together. We bootstrap the $30$ contexts with replacement
$20{,}000$ times and take the $2.5$th and $97.5$th percentiles of the
mean paired difference as a $95\%$ confidence interval. A two-sided
paired $t$-test compares the models' per-context legitimacy rates.

Bootstrap resampling uses
\texttt{numpy.random.default\_rng(0)}. The analysis is reproduced
by running \path{grids/eval/legit_cluster_inference.py}
end-to-end.

Default achieves $55.0\%$ legitimacy ($132/240$), versus
$44.2\%$ for minWM ($106/240$). The observed difference is $10.8$
percentage points, with a context-bootstrap $95\%$ interval of
$[-4.6,+26.2]$ points and paired-test $p=0.19$. We therefore report
an empirical benchmark advantage without claiming statistical
significance. Other model comparisons are descriptive unless
explicitly tested. We wish to highlight that we train on unsupervised video with a recovered affine action basis whereas minwm need well labelled data.  

This uncertainty analysis conditions on the trained checkpoints
and fixed peer reference; it does not quantify training-seed or
reference-panel variation. Response-curve SEM bands are unadjusted
rollout-level summaries, while wedge arcs show interquartile ranges
across rollouts. Neither represents variation across training runs.
\end{document}